\RequirePackage{fix-cm}
\documentclass[]{xiaomiev}
\microtypesetup{expansion=false}
\DeclareFontShape{T1}{xiaomiev}{b}{sc}{<-> s * [1] xiaomiev}{}
\usepackage{xspace}
\usepackage{mathtools,amsmath,amssymb}
\usepackage[ruled,linesnumbered]{algorithm2e}
\usepackage{enumitem}
\usepackage{float}
\usepackage{pifont}
\usepackage{needspace}
\usepackage{tikz}
\usepackage{pgfplots}
\usetikzlibrary{arrows.meta,positioning,calc,shapes.geometric,fit,backgrounds}
\pgfplotsset{compat=1.18}
\setcitestyle{authoryear,round}
\usepackage{tabularx}
\usepackage{colortbl}
\usepackage{etoolbox}

\definecolor{hxOrange}{HTML}{ED722E}
\definecolor{hxNavy}{HTML}{1B262C}
\definecolor{hxBlue}{HTML}{0F4C75}
\definecolor{hxTeal}{HTML}{3282B8}
\definecolor{hxSky}{HTML}{BBE1FA}
\definecolor{miBest}{HTML}{FF7E00}
\definecolor{miSecond}{HTML}{FFBE80}
\definecolor{cPR}{HTML}{718FBF}
\definecolor{cP}{HTML}{D891A2}
\definecolor{cE}{HTML}{8DB879}
\definecolor{cF}{HTML}{D8B85E}
\definecolor{cN}{HTML}{A1A5AA}
\definecolor{cMZ}{HTML}{7E8490}
\colorlet{cBox}{miBest!10}
\definecolor{cUp}{HTML}{527D40}
\definecolor{cDn}{HTML}{A45E68}
\definecolor{paperInk}{HTML}{252A31}
\colorlet{paperRule}{black}
\definecolor{paperGrid}{HTML}{E4E7EB}

\newcommand{\papertablefont}{\footnotesize}
\newcommand{\paperdensefont}{\fontsize{8}{9.5}\selectfont}
\newcolumntype{R}{>{\raggedleft\arraybackslash}X}
\AtBeginEnvironment{table}{%
  \papertablefont\setlength{\tabcolsep}{5pt}%
  \arrayrulecolor{black}}

\setlist[itemize]{leftmargin=15pt}
\DeclareTextFontCommand{\textbf}{\bfseries\color{black}}
\titleformat*{\paragraph}{\normalfont\normalsize\bfseries\color{black}}

\definecolor{caseBefore}{HTML}{ED722E}
\definecolor{caseAfter}{HTML}{0F4C75}
\definecolor{caseBlue}{HTML}{0F4C75}
\newtcolorbox{casecard}[2]{
  enhanced,colback=white,colframe=#1!70,colbacktitle=#1!12,
  coltitle=#1,fonttitle=\sffamily\bfseries\small,title={#2},
  arc=1mm,boxrule=.5pt,left=3mm,right=3mm,top=2mm,bottom=3mm,
  before skip=0pt,after skip=0pt,fontupper=\small,fontlower=\small,
  equal height group=casecard-\arabic{figure},space to upper,
  lower separated=false,middle=1mm
}
\newlength{\caseslotheight}
\newlength{\casepartnerheight}
\newcommand{\caseimage}[2][]{%
  \settoheight{\caseslotheight}{\includegraphics[width=1.0\linewidth]{#2}}%
  \if\relax\detokenize{#1}\relax\else
    \settoheight{\casepartnerheight}{\includegraphics[width=1.0\linewidth]{#1}}%
    \ifdim\casepartnerheight>\caseslotheight\setlength{\caseslotheight}{\casepartnerheight}\fi
  \fi
  \noindent\parbox[b][\caseslotheight][c]{\linewidth}{\centering\includegraphics[width=1.0\linewidth]{#2}}\par\smallskip
}
\newcommand{\casemodelstage}[1]{%
  \par{\footnotesize
  \textbf{Model:} #1\par
  \textbf{Before:} $S_0$ (public starting model)\par
  \textbf{After:} $S_{32}$ (Pool 3, window 2)\par
  \textbf{Training progress:} continuation window 8\par
  }\smallskip
}
\newcommand{\casetokens}[2]{%
  \tcblower
  \begingroup
  \pgfmathsetlengthmacro{\casebarwidth}{.93\linewidth}
  \begin{tikzpicture}[x=\casebarwidth,y=.34cm,font=\scriptsize]
    \node[anchor=west,inner sep=0pt] at (0,2.35) {Generated tokens (decode seed 0)};
    \fill[caseBefore!24,rounded corners=1pt] (0,1) rectangle (1,1.55);
    \node[anchor=west,inner sep=2pt] at (0,1.28) {Before $S_0$: #1};
    \fill[caseAfter!24,rounded corners=1pt] (0,0) rectangle ({#2/#1},.55);
    \node[anchor=west,inner sep=2pt] at (0,.28) {After $S_{32}$: #2};
  \end{tikzpicture}
  \endgroup\par
}

\newcommand{\macro}{\textsc{Macro12}\xspace}
\newcommand{\lj}{\textsc{LJ}\xspace}
\newcommand{\St}{S_t}
\DeclareRobustCommand{\methodwordmark}{\mbox{{\fontfamily{xiaomievmedium}\fontseries{m}\fontshape{n}\selectfont\mathversion{normal}\ensuremath{\mathcal{A}}\kern-.035em-RSI}}}
\DeclareRobustCommand{\methodwordmarkbold}{\mbox{{\fontfamily{xiaomiev}\fontseries{b}\fontshape{n}\selectfont\ensuremath{\boldsymbol{\mathcal{A}}}\kern-.035em-RSI}}}

\DeclareRobustCommand{\methodnamebold}{\textcolor{hxOrange}{\methodwordmarkbold}}
\newcommand{\arsi}{\methodname\xspace}
\newcommand{\arsifirst}{\methodnamebold\xspace}
\newcommand{\abstractkey}[1]{\textbf{#1}}
\newcommand{\armFront}{\textsc{Frontier}\xspace}
\newcommand{\armPeriodic}{\textsc{Periodic}\xspace}
\newcommand{\armCont}{\textsc{Continuous}\xspace}
\newcommand{\armNR}{\textsc{No-Refresh}\xspace}

\title{{\fontsize{14}{18}\selectfont \methodnamebold: Adaptive Task Refresh\\for Zero-Data Multimodal Self-Improvement}}
\DeclareRobustCommand{\centeredlogo}[2]{\raisebox{-.5\height}{\includegraphics[#1]{assets/logos/#2}}}

\DeclareRobustCommand{\numberedlogo}[3]{\textsuperscript{\normalfont\scriptsize #1}\kern1.5pt\centeredlogo{#2}{#3}}
\newcommand{\institutionlogos}{%
  \noindent\makebox[\linewidth]{\hfil\numberedlogo{1}{width=90pt,height=28pt,keepaspectratio}{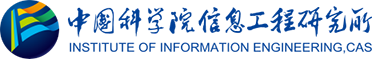}\hfil\numberedlogo{2}{width=100pt,height=28pt,keepaspectratio}{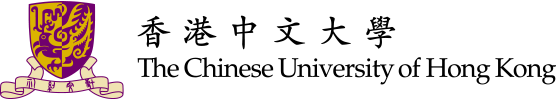}\hfil\numberedlogo{3}{width=65pt,height=28pt,keepaspectratio}{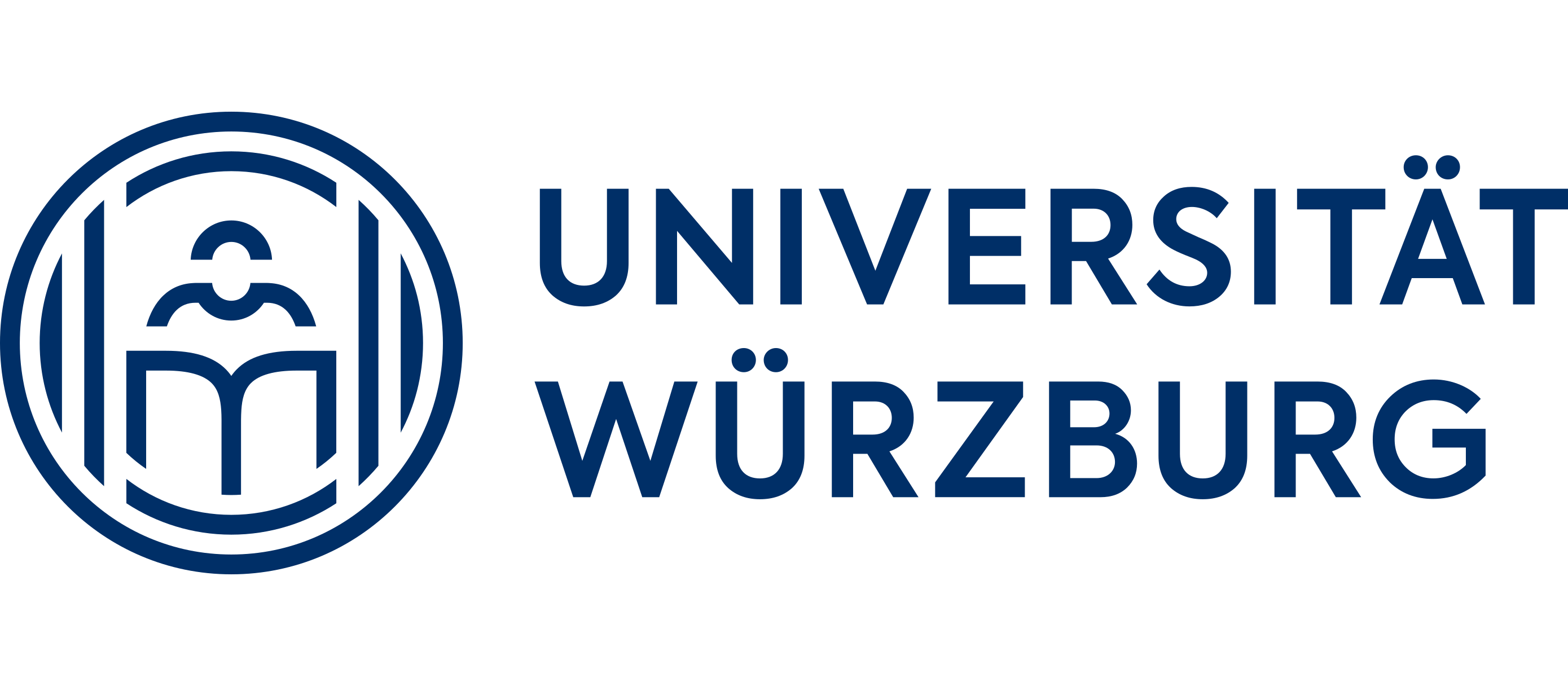}\hfil\numberedlogo{4}{height=22pt,keepaspectratio}{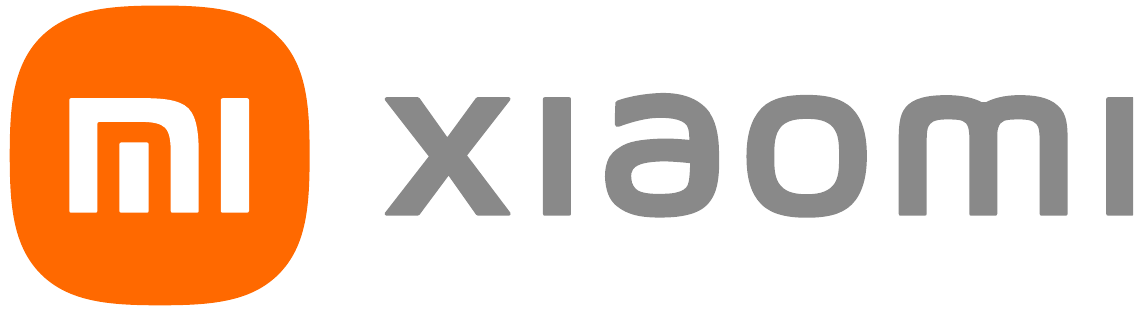}\hfil}\par\vspace{3pt}%
  \noindent\makebox[\linewidth]{\hfil\numberedlogo{5}{width=105pt,height=28pt,keepaspectratio}{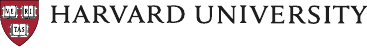}\hfil\numberedlogo{6}{width=37pt,height=28pt,keepaspectratio}{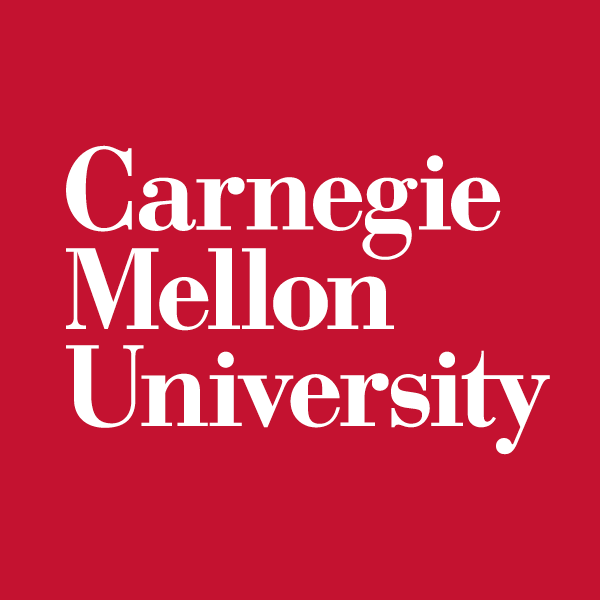}\hfil\numberedlogo{7}{width=78pt,height=28pt,keepaspectratio}{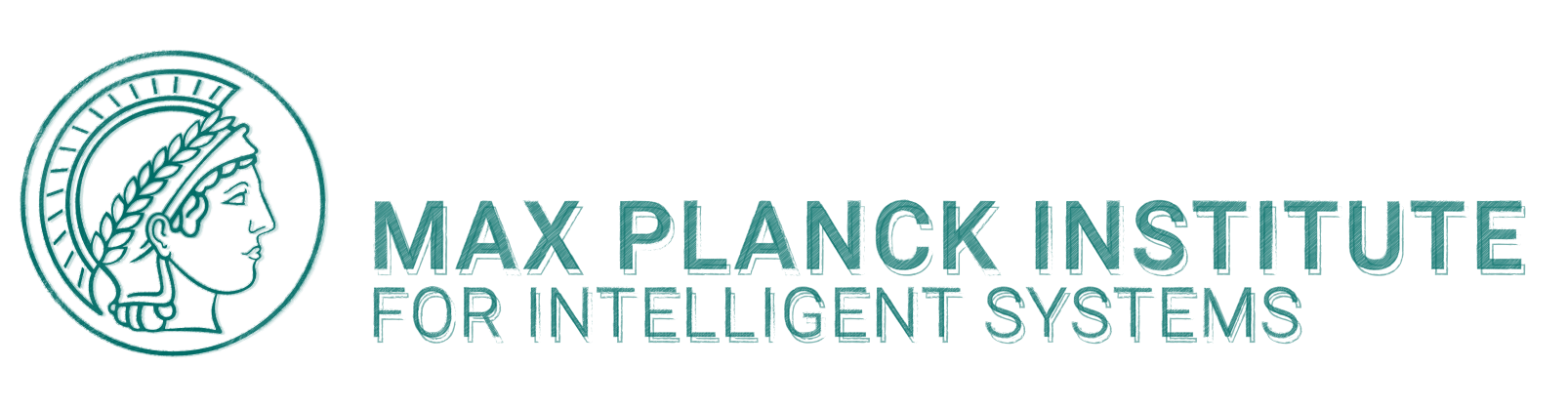}\hfil\numberedlogo{8}{width=75pt,height=28pt,keepaspectratio}{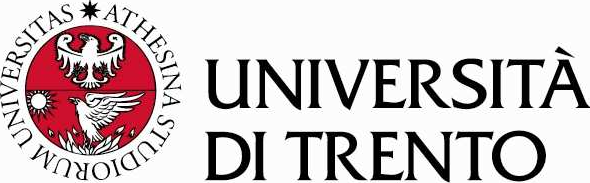}\hfil\numberedlogo{9}{width=90pt,height=28pt,keepaspectratio}{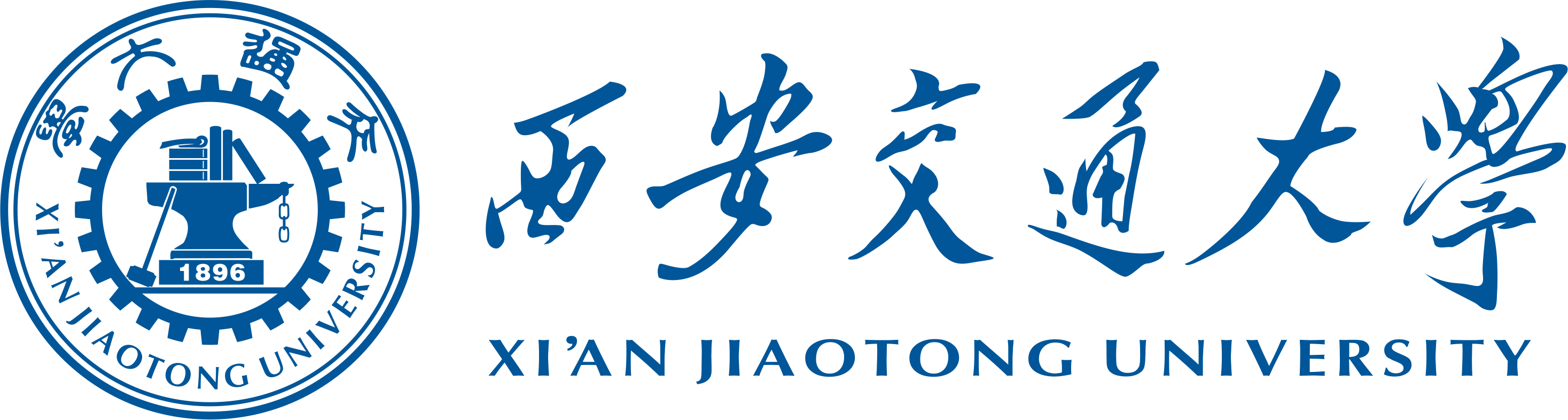}\hfil}%
}
\renewcommand{\authorformat}[2][]{{\fontsize{9.5}{11.5}\selectfont\rmfamily\bfseries #2}}
\author{%
\mbox{Yu Liu\textsuperscript{1,4,*}}\quad
\mbox{Youxing Li\textsuperscript{2,*,\textsection}}\quad
\mbox{Yuedong Tan\textsuperscript{3,*}}\quad
\mbox{Zhiwei Yang\textsuperscript{1,4}}\quad
\mbox{Zihan Su\textsuperscript{5}}\\[1pt]
\mbox{Hanqi Feng\textsuperscript{6}}\quad
\mbox{Shaojie Zhang\textsuperscript{4}}\quad
\mbox{Libin Liao\textsuperscript{9}}\quad
\mbox{Zihao Wan\textsuperscript{9}}\quad
\mbox{Chenxu Wu\textsuperscript{9}}\\[1pt]
\mbox{Cong Cao\textsuperscript{1,\textdagger}}\quad
\mbox{Fangfang Yuan\textsuperscript{1}}\quad
\mbox{Yanbing Liu\textsuperscript{1}}\quad
\mbox{Ziheng Chen\textsuperscript{7,8}}\quad
\mbox{Pei Fu\textsuperscript{4,\textdagger}}\\[1pt]
\mbox{Radu Timofte\textsuperscript{3}}\quad
\mbox{Zongwei Wu\textsuperscript{3}}\quad
\mbox{Nicu Sebe\textsuperscript{8}}\quad
\mbox{Barnabas Poczos\textsuperscript{6}}\quad
\mbox{Jian Luan\textsuperscript{4}}\\[3pt]
{\normalfont\fontsize{8}{9.5}\selectfont
\textsuperscript{*}Equal contribution\quad
\textsuperscript{\textsection}Project lead\quad
\textsuperscript{\textdagger}Corresponding authors}\\[4pt]
\institutionlogos}

\makeatletter
\patchcmd{\@toptitlebar}{\vskip 8mm}{\vskip 2mm}{}{}
\patchcmd{\@bottomtitlebar}{\vskip 6mm}{\vskip 1.5mm}{}{}
\patchcmd{\mymaketitle}{\vskip 1.5mm}{\vskip 0.5mm}{}{}
\patchcmd{\mymaketitle}{\vskip 8mm}{\vskip 1.5mm}{}{}
\patchcmd{\mymaketitle}{\vskip 6mm}{\vskip 0.5mm}{}{}
\pretocmd{\mymaketitle}{\vspace*{-15mm}}{}{}
\makeatother
\renewcommand{\abstractinfont}{\fontsize{9}{10.5}\selectfont}
\abstract{
\abstractkey{Recursive self-improvement} requires models not only to acquire useful training experience, but also to decide when their current experience has ceased to be useful. Existing methods often focus on what experience to generate next, while paying less attention to when fresh experience should be introduced. Many also construct training targets through self-consistency or majority voting, making supervision depend on the model's own uncertain predictions. We address these two questions separately: \abstractkey{training dynamics} determine when to acquire new experience, while the answers to new tasks remain \abstractkey{independently checkable}. \arsifirst{} monitors reward variation and accuracy in training logs to refresh a task pool when its learning signal declines, without consulting validation or test scores. A static library of \abstractkey{model-authored programs} produces multimodal questions and derives their reference answers; we specify the task-generation contract, automatically verify chosen properties of the reward and refresh rules, and separately audit the generated answers. The training loop uses no external training examples or human-written answer labels. On the primary eleven-benchmark suite, \arsi{} raises the performance average from 57.77 to 61.05 with 800 fresh tasks.\par
}

\hypersetup{
  pdftitle={A-RSI: Adaptive Task Refresh for Zero-Data Multimodal Self-Improvement},
  pdfauthor={Yu Liu; Youxing Li; Yuedong Tan; Zhiwei Yang; Zihan Su; Hanqi Feng; Shaojie Zhang; Libin Liao; Zihao Wan; Chenxu Wu; Cong Cao; Fangfang Yuan; Yanbing Liu; Ziheng Chen; Pei Fu; Radu Timofte; Zongwei Wu; Nicu Sebe; Barnabas Poczos; Jian Luan},
  pdfsubject={Adaptive Recursive Self-Improvement}
}

\begin{document}
\maketitle\par

\begingroup
\setlength{\intextsep}{5pt}
\begin{figure}[H]
\centering
\includegraphics[width=1.0\linewidth]{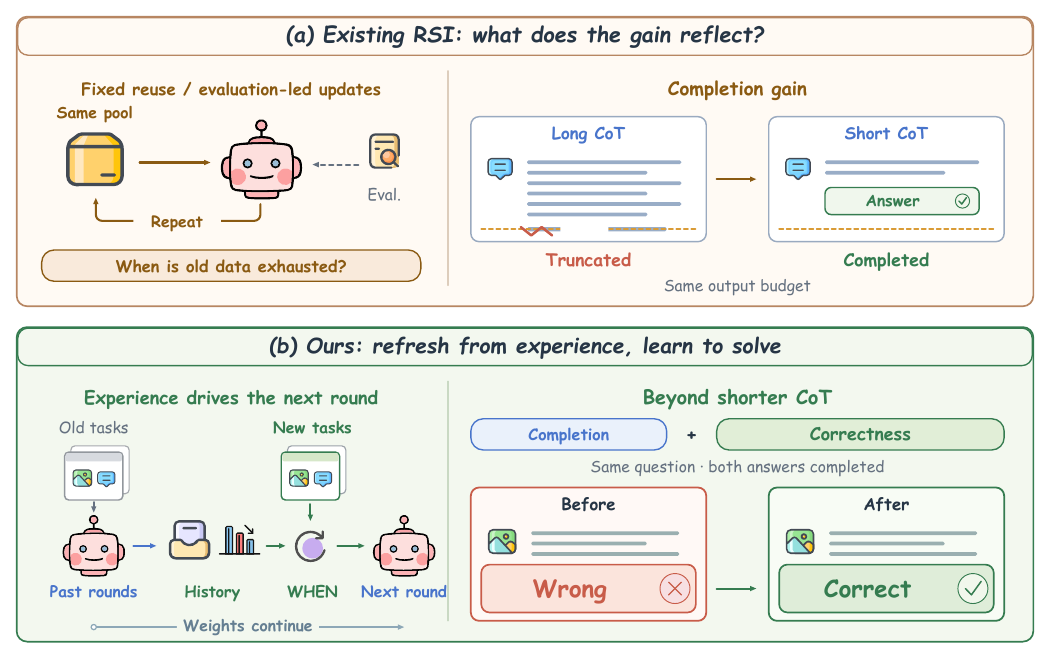}
\caption{Motivation for \arsi{}. (a) Pool reuse can stall learning, while fewer truncations can raise accuracy. (b) Training logs guide refresh, and response audits distinguish completion from error correction.}
\label{fig:aspire-motivation}
\end{figure}
\endgroup

\clearpage
\setcounter{tocdepth}{2}
\tableofcontents
\clearpage

\section{Introduction}

Recursive self-improvement depends on both acquiring useful experience and recognizing when the current experience is no longer useful. Recent methods have emphasized what to generate next, such as harder problems or tasks near the model's estimated competence. Yet a system can continue spending updates on a pool whose learning signal has declined, while refreshing too often consumes new tasks that may not be needed. This makes the timing of replacement a distinct decision from the choice of replacement tasks.

The answers attached to self-generated tasks raise a separate reliability question. When training targets come from self-consistency or majority voting, errors in the model's predictions can become labels for its next update. We therefore separate the source of a refresh decision from the source of answer correctness: training dynamics can indicate when to draw new experience, while the new tasks should have independently checkable answers.

We implement this separation in \arsifirst{} (Adaptive Recursive Self-Improvement). In group-based reinforcement learning, prompts whose sampled responses receive different rewards provide nonzero relative-advantage signal. \arsi{} monitors how many distinct prompts retain this signal, together with logged answer accuracy, and refreshes the active pool when the specified training-log condition fires. This is an operational indicator of declining learning signal, not a proof that the pool has been mastered. The controller does not consult validation or test scores.

A fixed library of model-written programs supplies the new multimodal tasks. Each program constructs an image and question from sampled parameters and computes a reference answer from the same underlying configuration. We specify this shared-configuration contract, machine-check selected reward and refresh properties, and independently audit generated answers. These checks support the use of program-computed supervision without asserting that every rendered question has been formally proved correct.

The loop trains the Solver on the current pool, reads its training logs, and either retains the pool or activates a prepared, unused replacement. A refresh keeps the learned model weights; the task templates and sampling quotas remain fixed. The training stage uses no external training examples, human-written answer labels, or evaluation scores. In the tested Qwen comparisons, difficulty-aware and competence-aware replacement rules do not outperform random sampling under the fixed task-family shares.

On the primary eleven-benchmark suite, Qwen improves from 57.77 at the post-initialization checkpoint $\St$ to 61.05 after eight continuation windows and 800 fresh tasks. MiMo improves from 59.08 at its public base model $S_0$ to 60.71. On the same eleven-benchmark protocol, \arsi scores 61.05 with 800 fresh tasks, while every-window refresh scores 60.64 with 3,200. At the matched sixth continuation window, \arsi scores 60.40 against 58.35 for continued training on the old pool. Figure~\ref{fig:aspire-motivation} also motivates our response-level audit: a score can rise because more answers finish within the output budget or because already completed answers are corrected. We report these outcomes separately.

Our contributions are threefold:
\begin{itemize}
    \item \textbf{A training-log refresh rule.}
    We formulate a pool-replacement rule using the decline in prompts with within-group reward variation, without consulting evaluation scores.

    \item \textbf{Programmatically verified experience.}
    Model-written programs instantiate multimodal tasks and compute their reference answers, allowing the training loop to operate without external training examples or human-written answer labels.

    \item \textbf{Matched comparisons and output-level analysis.}
    We compare refresh policies at matched training progress across full benchmarks and distinguish gains linked to response completion from corrections among completed answers. The Qwen results identify the benefit of the implemented replacement policy and the limits of the timing claim.
\end{itemize}

\section{Related Work}\label{sec:related}

Recursive self-improvement research studies how a system can produce useful experience and use it to improve. Prior work has changed the system around a model, as in STOP and the Darwin G\"odel Machine, or updated model weights from generated rationales, filtered samples, self-play, or model-based judgments \citep{zelikman2024stop,zhang2025dgm,zelikman2022star,singh2024restem,chen2024spin,yuan2024selfrewarding,wang2023selfinstruct}. We focus on a control question within weight-based improvement: how can a system tell, without evaluation feedback, when its current training pool has stopped supplying useful signal? Our refresh decision comes from training logs; it does not adapt the task-generation rules during training. Table~\ref{tab:related} compares how related systems obtain supervision and change their training data.

\paragraph{Zero-data self-improvement and verifiable rewards.} Absolute Zero uses an executor to validate code-generated tasks and answers, while R-Zero trains a Challenger and Solver to co-evolve around the Solver's current ability \citep{zhao2025absolutezero,huang2026rzero}. Multimodal work extends self-play through generated visual games or multi-role systems, including Vision-Zero and MM-Zero \citep{wang2026visionzero,li2026mmzero}. These methods primarily develop how tasks are proposed, solved, or adapted. \arsi{} instead fixes a library of task-generating programs and studies when to draw a fresh pool. Each sampled instance has a program-computed answer, so answer correctness does not depend on a model vote. This distinction complements reinforcement learning with verifiable rewards (RLVR), where methods such as GRPO and DeepSeek-R1 optimize against outcome checks rather than human-written reasoning traces \citep{shao2024deepseekmath,guo2025deepseekr1}. DAPO's dynamic sampling also removes prompts whose sampled responses all receive the same outcome, since they yield no within-group policy gradient \citep{yu2025dapo}. We use the related signal for a different decision: $U_w$ aggregates informative prompts across a window and helps determine whether the entire pool should be refreshed. It is a pool-level diagnostic, not a replacement for batch-level sampling.

\paragraph{Procedural tasks and task selection.} Programmatic generation provides an open-ended source of varied tasks, as demonstrated by Procgen for reinforcement-learning environments and by language-model resources such as Reasoning Gym and SynLogic for verifiable reasoning \citep{cobbe2019procgen,stojanovski2025reasoninggym,liu2025synlogic}. Curriculum and adaptive-sampling methods address a related but distinct problem: they prioritize examples or environments according to estimated difficulty or learning value. Examples include Prioritized Level Replay and PAIRED in reinforcement learning, and online difficulty filtering, bandit-based curricula, and distribution-level sampling for language-model training \citep{jiang2020plr,dennis2020paired,bae2026odf,chen2025sec,wang2025dump}. ThinkLite-VL is especially close to our setting: it uses search-based difficulty estimates to select a smaller, challenging subset from a fixed collection of visual reasoning examples \citep{wang2025thinklite}. Those studies motivate testing selection, but they do not settle how to choose a replacement pool after a training plateau. We compare difficulty-frontier, difficulty-ladder, and rollout-based selection with random draws under fixed family shares. In our experiments, the selection strategies do not outperform random replacement; this result is limited to our generators, model, and tested selection rules.

\paragraph{Training duration and evaluation budgets.} The apparent benefit of continued reasoning can depend on how generations are terminated and scored. Dr.~GRPO analyzes response-length bias, DAPO includes overlong-reward shaping to address truncation, and L1 explicitly trains length-budget control \citep{liu2025drgrpo,yu2025dapo,aggarwal2025l1}. This work motivates a separate audit in our setting: we check whether a measured gain persists under a longer generation budget and format-corrected scoring, including when evaluating released MM-Zero weights (Section~\ref{sec:budget}). The audit is not a new reward or refresh rule; it tests whether an apparent improvement reflects better task performance or only a greater chance of completing within the evaluation limit.

\begin{table}[!htbp]
\caption{Comparison with related self-improvement and data-selection methods. Most change their training data continuously or at every step, whereas \arsi replaces the whole pool only when its own training logs show saturation and refills it with random programmatic tasks.}
\label{tab:related}
\centering
\papertablefont
\begin{tabularx}{\linewidth}{@{}>{\raggedright\arraybackslash}p{2.9cm}*{4}{>{\raggedright\arraybackslash}X}@{}}
\toprule
\textbf{Method} & \textbf{Task source} & \textbf{Supervision} & \textbf{Data updated} & \textbf{Selection rule} \\
\midrule
\multicolumn{5}{@{}l}{\textit{Generate tasks during training}} \\
Absolute Zero & Model-proposed code & Code execution & Continuously & Learnability reward \\
R-Zero & Challenger model & Model labels & Co-evolving & Near solver ability \\
MM-Zero & Multi-role model & Majority vote & Continuously & Difficulty reward \\
Vision-Zero & Image-pair games & Game rewards & Alternating & Self-play $\leftrightarrow$ RLVR \\
\midrule
\multicolumn{5}{@{}l}{\textit{Select from existing tasks or distributions}} \\
ThinkLite-VL & 11k public tasks & Dataset labels & One-time & MCTS / pass rate \\
Online Diff.\ Filtering & Fixed prompt set & Dataset labels & Every batch & Mid-range difficulty \\
SEC / DUMP & Fixed distributions & Dataset labels & Every step & Bandit on advantage \\
\midrule
\rowcolor{cBox}
\textbf{\arsi{} (ours)} & Model-written templates & Program-computed & \textbf{Log-triggered} & Random, fixed shares \\
\bottomrule
\end{tabularx}

\end{table}
\FloatBarrier
\section{Method}\label{sec:method}

\begin{figure}[!htbp]
\centering
\includegraphics[width=1.0\linewidth]{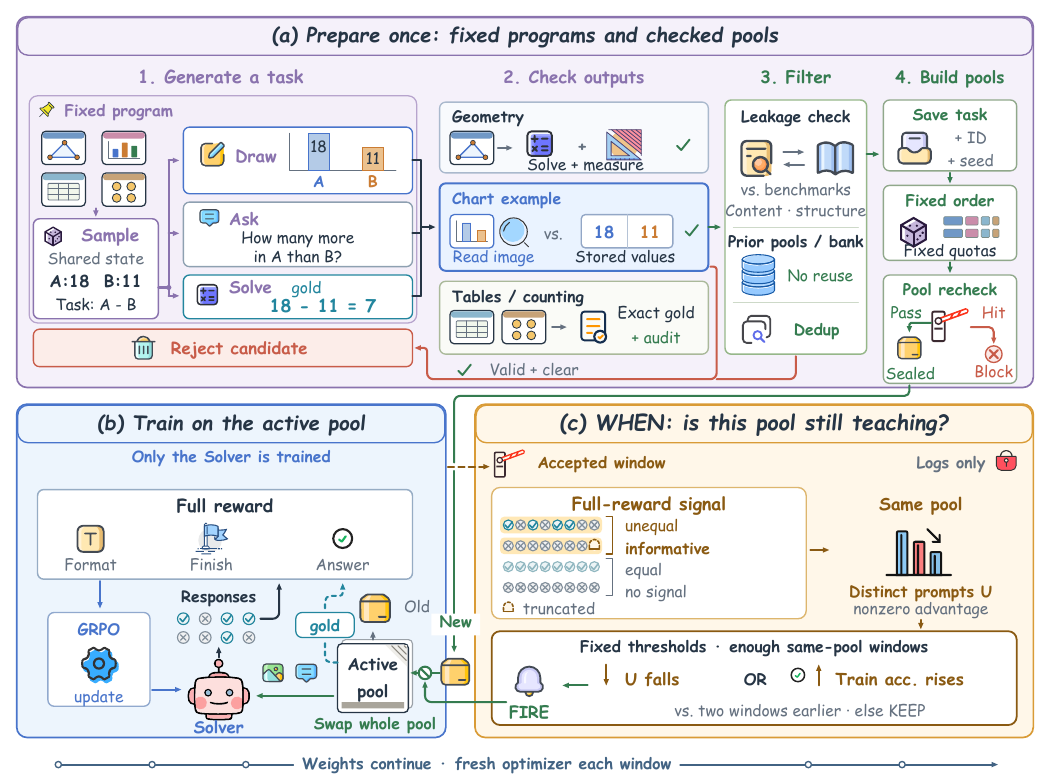}
\caption{Overview of \arsi{}. Task templates, the candidate reservoir, and replacement pools are fixed before training. The controller reads training logs to decide when to activate a fresh pool; only the Solver policy is optimized.}
\label{fig:aspire-rsi-overview}
\end{figure}

\arsi{} supports continued multimodal reinforcement learning through a programmatic task pipeline and a training-log controller. Figure~\ref{fig:aspire-rsi-overview} organizes the workflow into three parts. Before training, fixed, model-written templates generate images, questions, and program-computed answers; the pipeline checks candidates and prepares disjoint replacement pools under fixed family quotas (Figure~\ref{fig:aspire-rsi-overview}(a); Section~\ref{sec:gen}). During training, the Solver learns on the active pool with GRPO using correctness, format, and termination rewards (Figure~\ref{fig:aspire-rsi-overview}(b); Section~\ref{sec:signal}). After each accepted window, the controller reads training-side accuracy and group-relative advantage signals to decide whether to retain the current pool or activate the next unused one (Figure~\ref{fig:aspire-rsi-overview}(c); Section~\ref{sec:det}). A refresh replaces the whole active pool while retaining the latest accepted model weights (Section~\ref{sec:refresh}). Training experience therefore changes the timing of data replacement; it does not update the generator or the task-sampling distribution.

Let $\theta_{w-1}$ be the model parameters and $\mathcal{P}_w$ the active pool at the start of window $w$. A window consists of $H$ GRPO updates \citep{shao2024deepseekmath}, producing parameters $\theta_w$ and a rollout log $\mathcal{L}_w$. In the main Qwen experiment, $\theta_0=\St$ is obtained by training the public Qwen3-VL-8B-Instruct checkpoint on the initial task pool. The index $w$ counts subsequent continuation windows, whereas $S_0$ denotes the public model before this initialization. We report the resulting model states with pool-indexed notation $S_{ij}$, defined in Section~\ref{sec:setup}. Each replacement introduces $B$ unused tasks; concrete pool sizes and optimization settings are given in Appendix~\ref{app:initial-pool}. Algorithm~\ref{alg:refresh} specifies this main training loop; Section~\ref{sec:setup} describes the comparison arms and {MiMo experiments}.

\subsection{Programmatic task construction}\label{sec:gen}

Template construction takes place once, before training. Starting from seed task types in geometry, charts, tables, and counting, the base model generates reusable problem structures and writes their parameterized programs. For each structure, the model specifies the admissible parameters, the conditions and quantity to ask for, and the procedures for sampling values, drawing an image, composing a question, and computing its answer. The resulting library $\mathcal{T}$ stores these executable templates and stays fixed throughout training.

\Needspace{6\baselineskip}
Task construction then executes the frozen templates. A generator selects $\tau\in\mathcal{T}$ and samples parameters $z$ from its distribution $p_\tau$. The resulting instance produces an image $I$, a question $q$, and a reference answer $a$:
\begin{equation}
 z\sim p_\tau,\qquad (I,q,a)=G_\tau(z).
 \label{eq:task-generation}
\end{equation}
The program first constructs the underlying data or geometric configuration, then uses it to produce all three outputs. For example, a chart template can sample category values $v_A=18$ and $v_B=11$, render their bars, ask how much larger $A$ is than $B$, and compute $a=v_A-v_B=7$ from the stored values. The learner receives only $(I,q)$, rather than the latent values or answer, and its response is scored against $a$. Replacement tasks require no further model call to generate their questions or reference answers.

Admission checks are specific to each generator: they test the sampled configuration, the answer computation, and applicable rendering and ambiguity constraints. Freshness and contamination screens then reject flagged candidates, and held-out diagnostic items are excluded from training. Retained records associate the image, question, and answer with a generator identity and seed for reconstruction. These checks and independent answer audits are detailed in Appendix~\ref{app:task-details}; program-computed supervision remains subject to the correctness of the template and its implementation.

The generators build a reservoir before continuation training. Candidates that pass the construction, freshness, and contamination checks and lie within the configured training ladders form the eligible set $\mathcal{C}$, after excluding diagnostic holdouts. This set supplies the replacement pools. The initial pool comes from earlier geometry and chart batches with model-assisted question rewriting and checks. Generator allocations and reservoir sizes are given in Appendix~\ref{app:task-eligibility}.

\subsection{Rewards and policy optimization}\label{sec:signal}

For each sampled task, the model produces $m$ responses. We score each response $y$ against the program-computed answer $a$ and check its format and stopping condition:
\begin{equation}
 r(y,a)=r_{\mathrm{ans}}(y,a)+\lambda_{\mathrm{fmt}}\,r_{\mathrm{fmt}}(y)+\lambda_{\mathrm{stop}}\,r_{\mathrm{stop}}(y).
 \label{eq:reward}
\end{equation}
The answer reward $r_{\mathrm{ans}}$ is 1 when the extracted answer matches $a$ and 0 otherwise. The format reward $r_{\mathrm{fmt}}$ is 1 when the response follows the required boxed-answer format and 0 otherwise. The stopping term $r_{\mathrm{stop}}$ is 0 when decoding ends normally with a final boxed answer and $-1$ otherwise. Reaching the response limit incurs the stopping penalty even if an earlier boxed expression is present. All normally completed responses with a final boxed answer receive the same stopping reward, irrespective of length; there is no per-token length penalty or separate reward for hop count.

We update only the Solver policy with GRPO, using relative advantages computed from these rewards within each response group. The templates and controller remain fixed. The coefficients $\lambda_{\mathrm{fmt}}$ and $\lambda_{\mathrm{stop}}$ are fixed throughout training. Their values, the response limit, and optimizer and sampling settings appear in Appendix~\ref{app:initial-pool}.

\subsection{Detecting pool saturation}\label{sec:det}

A response group with identical total rewards has zero GRPO advantages and provides no reward-derived policy-gradient signal. As training changes the response distribution, previously informative prompts may cease to provide within-group reward differences. We monitor the number of prompts with nonzero advantages alongside response accuracy to obtain a training-side heuristic for deciding when to refresh the pool. Here, ``saturation'' refers to this operational rule, rather than a measured plateau in held-out performance.

Let $\mathcal{G}_w$ contain the response groups logged after dynamic resampling, with each generation batch counted once. For group $g$, let $x_g$ be the full prompt, $A_{g,j}$ the logged advantage of response $j$, and $c_{g,j}$ its answer reward. We compute
\begin{equation}
 \begin{aligned}
 U_w&=\left|\left\{x_g:\ g\in\mathcal{G}_w,\ \max_{1\leq j\leq m}|A_{g,j}|>\varepsilon\right\}\right|,\\
 a_w&=\frac{1}{m|\mathcal{G}_w|}\sum_{g\in\mathcal{G}_w}\sum_{j=1}^{m}c_{g,j},
 \end{aligned}
 \label{eq:pool-signals}
\end{equation}
where $\varepsilon>0$ is a numerical tolerance. $U_w$ counts distinct observed prompts with nonzero advantages, and $a_w$ measures how often the logged responses are correct. Advantages use the full reward in Eq.~\eqref{eq:reward}, so differences in formatting or stopping can also contribute to $U_w$.

At the end of each window, we compare these signals with their values $\ell$ windows earlier. Let $n_w$ be the number of accepted consecutive windows on this pool, $n_{\min}>\ell$ the minimum observation period, $\rho\in(0,1)$ the signal-retention threshold, and $\eta>0$ the required accuracy increase. The decision to refresh is
\begin{equation}
 \delta_w=
 \begin{cases}
 1,& n_w\geq n_{\min}\ \text{and}\ \bigl(U_w\leq\rho U_{w-\ell}\ \text{or}\ a_w-a_{w-\ell}\geq\eta\bigr),\\
 0,&\text{otherwise}.
 \end{cases}
 \label{eq:trigger}
\end{equation}
{
Either a relative decline in $U_w$ below the retention threshold or a sufficient increase in $a_w$ triggers replacement. The first condition detects fewer observed prompts supplying relative-advantage signal. The second uses increased accuracy as a proxy for growing familiarity with the pool. The minimum observation period $n_{\min}$ ensures that both measurements come from the same pool. It does not prescribe a refresh interval: when neither condition holds, the controller retains the pool and checks again after the next accepted window. The lag and thresholds remain fixed across stages and models, while the decision depends on each model's current training feedback.
\par}

Neither condition establishes that the model has mastered the pool or stopped improving on external tasks. Uniformly incorrect responses and a concentrated response distribution can also remove within-group variation, while dynamic resampling affects which prompts are observed. Accordingly, $U_w$ describes the logged training signal, rather than an exhaustive estimate of the pool's remaining learnability. The usefulness of this refresh rule is assessed by the continuation experiments, separately from these signal definitions. {The shared controller settings and reward-component replay are reported together in Appendices~\ref{app:controller-settings} and~\ref{app:controller-replay}.}

\subsection{Fresh replacement and continued training}\label{sec:refresh}

When $\delta_w=1$, we replace the entire pool with unused tasks from $\mathcal{C}$. Let $\mathcal{H}_w$ contain all tasks introduced so far. We draw without replacement under fixed family quotas $\mathbf{k}=(k_{\mathrm{geo}},k_{\mathrm{chart}},k_{\mathrm{table}},k_{\mathrm{count}})$, with $\sum_f k_f=B$:
\begin{equation}
 \mathcal{P}_{w+1}=
 \begin{cases}
 \operatorname{Draw}_{\mathbf{k}}(\mathcal{C}\setminus\mathcal{H}_w),&\delta_w=1,\\
 \mathcal{P}_w,&\delta_w=0.
 \end{cases}
 \label{eq:pool-update}
\end{equation}
Here $\operatorname{Draw}_{\mathbf{k}}$ takes tasks in a fixed pseudorandom order within each family. The ordering does not depend on model answers or difficulty estimates. We execute these draws before training to prepare an ordered list of disjoint replacement pools, and the detector decides when to activate the next one. Replacing the entire active pool does not reset the model: $\theta_w$ carries forward experience from all preceding windows. The within-pool window count returns to zero after a refresh. Each new window starts a fresh optimizer and scheduler while inheriting the accepted model weights, whether or not the pool changes.

The method follows a training-feedback loop: train on an active pool, observe training-side signals, retain or replace the pool, and continue from accepted model weights. Only accepted windows contribute to the detector's history. Training-time health checks identify interrupted attempts, whose weights are discarded. Algorithm~\ref{alg:refresh} specifies the training and refresh procedure used to define the method.

In the main training loop, a failed window is retried once from the preceding accepted checkpoint on the same pool and seed; a second failure stops the run. If a refresh is requested but the declared replacement pools are exhausted, the controller refuses to start the next window. It does not silently reuse a pool or change the sampling rule. No replacement is needed after the final budgeted window.

\begin{algorithm}[!htbp]
\caption{\arsi{}: pool refresh from training logs}
\label{alg:refresh}
\KwIn{Starting checkpoint $\theta_0=\St$, initial pool $\mathcal{P}_1$, eligible library $\mathcal{C}$}
\KwIn{Ordered, disjoint replacement pools prepared from $\mathcal{C}$ under quotas $\mathbf{k}$}
\KwIn{Training budget $W$ windows}
Set within-pool count $n\gets0$ and task history $\mathcal{H}\gets\mathcal{P}_1$\;
\For{$w=1,\ldots,W$}{
  Initialize a fresh optimizer and scheduler; train $H$ GRPO steps from $\theta_{w-1}$ on $\mathcal{P}_w$ to obtain $\theta_w,\mathcal{L}_w$\;
  \If{the health gate fails}{
    Retry from $\theta_{w-1}$ with fresh optimizer state, the same pool and seed; replace $\theta_w,\mathcal{L}_w$ with the retry outputs\;
    Halt if the retry fails the health gate\;
  }
  Increment $n$ and compute $U_w,a_w$ from $\mathcal{L}_w$ by Eq.~\eqref{eq:pool-signals}\;
  Evaluate $\delta_w$ by Eq.~\eqref{eq:trigger}, using $n_w=n$\;
  \If{$w=W$}{
    \Return $\theta_w$\;
  }
  \If{$\delta_w=1$ and no declared replacement pool remains}{
    Halt and return the last accepted checkpoint $\theta_w$\;
  }
  Set $\mathcal{P}_{w+1}$ by Eq.~\eqref{eq:pool-update}, using $\mathcal{H}_w=\mathcal{H}$\;
  \If{$\delta_w=1$}{
    Add $\mathcal{P}_{w+1}$ to $\mathcal{H}$ and reset $n\gets0$\;
  }
}
\Return last accepted checkpoint\;
\end{algorithm}

Appendices~\ref{app:task-details} and~\ref{app:health} provide the task construction details, optimization settings, and training checks.

\section{Experimental Setup}\label{sec:setup}

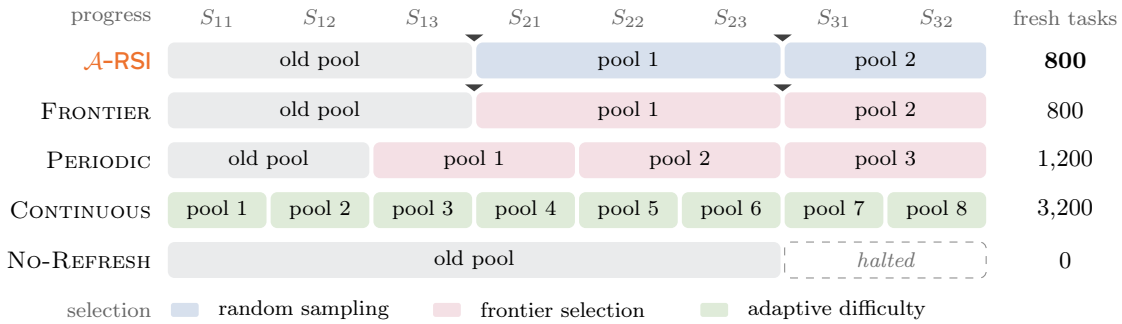
\begin{figure}[!htbp]
\centering
\begin{tikzpicture}[x=1.36cm, y=0.66cm, font=\small]
\newcommand{\armpool}[5]{
  \fill[#4, rounded corners=2.5pt] (#2-0.48,#1-0.36) rectangle (#3+0.48,#1+0.36);
  \node[font=\footnotesize] at ({(#2+#3)/2},#1) {#5};}
\foreach \w/\stage in {1/11,2/12,3/13,4/21,5/22,6/23,7/31,8/32} \node[font=\footnotesize, text=black!60] at (\w,0.85) {$S_{\stage}$};
\node[font=\footnotesize, text=black!60, anchor=east] at (0.45,0.85) {progress};
\node[font=\footnotesize, text=black!60] at (9.25,0.85) {fresh tasks};
\node[anchor=east] at (0.45,0) {\textbf{\arsi}};
\node[anchor=east] at (0.45,-1) {\armFront};
\node[anchor=east] at (0.45,-2) {\armPeriodic};
\node[anchor=east] at (0.45,-3) {\armCont};
\node[anchor=east] at (0.45,-4) {\armNR};
\armpool{0}{1}{3}{cN!22}{old pool}
\armpool{0}{4}{6}{cPR!28}{pool 1}
\armpool{0}{7}{8}{cPR!28}{pool 2}
\armpool{-1}{1}{3}{cN!22}{old pool}
\armpool{-1}{4}{6}{cP!28}{pool 1}
\armpool{-1}{7}{8}{cP!28}{pool 2}
\armpool{-2}{1}{2}{cN!22}{old pool}
\armpool{-2}{3}{4}{cP!28}{pool 1}
\armpool{-2}{5}{6}{cP!28}{pool 2}
\armpool{-2}{7}{8}{cP!28}{pool 3}
\foreach \w in {1,...,8} {\armpool{-3}{\w}{\w}{cE!28}{pool \w}}
\armpool{-4}{1}{6}{cN!22}{old pool}
\draw[black!45, dashed, rounded corners=2.5pt] (7-0.48,-4-0.36) rectangle (8+0.48,-4+0.36);
\node[font=\footnotesize\itshape, text=black!55] at (7.5,-4) {halted};
\foreach \y in {0,-1} \foreach \x in {3.5,6.5}
  \fill[black!75] (\x-0.09,\y+0.52) -- (\x+0.09,\y+0.52) -- (\x,\y+0.38) -- cycle;
\foreach \y/\n in {0/{\textbf{800}},-1/800,-2/{1,200},-3/{3,200},-4/0}
  \node at (9.25,\y) {\n};
\node[anchor=east, font=\footnotesize, text=black!60] at (0.45,-5) {selection};
\foreach \x/\poolcolor/\selection in {1/cPR/{random sampling},3.55/cP/{frontier selection},6.15/cE/{adaptive difficulty}} {
  \fill[\poolcolor!28, rounded corners=1pt] (\x-0.45,-5.14) rectangle (\x-0.18,-4.86);
  \node[anchor=west, font=\footnotesize] at (\x-0.08,-5) {\selection};}
\end{tikzpicture}
\caption{Training pools through $S_{32}$, with columns aligned by continuation updates. Each numbered block represents 400 fresh tasks, and its width shows how long that pool is used. Pools are numbered in replacement order separately for each arm. Colors identify selection strategies, and triangles mark detector-triggered replacements.}
\label{fig:arms}
\end{figure}

\paragraph{Arms.} Figure~\ref{fig:arms} compares five strategies for refreshing the training pool. The \emph{old pool} contains 500 tasks. Each window consists of 100 GRPO updates, and each refresh replaces the entire pool with 400 unused tasks. \emph{Random sampling} draws unused tasks under fixed task-family quotas. \emph{Frontier selection} uses accuracy on held-out synthetic tasks to find the easiest difficulty level the model does not yet solve reliably, then draws tasks from that level and its neighbors.
\begin{itemize}[leftmargin=*, topsep=3pt, itemsep=3pt, parsep=0pt]
  \item \textbf{\arsi} refreshes when the saturation detector fires, using random sampling.
  \item \textbf{\armFront} uses the same detector, but selects new tasks by frontier selection.
  \item \textbf{\armPeriodic} uses frontier selection and refreshes every two windows.
  \item \textbf{\armCont} refreshes every window and adjusts the selected difficulty up or down according to held-out accuracy.
  \item \textbf{\armNR} keeps the old pool throughout training.
\end{itemize}
Appendix~\ref{sec:tiers} specifies the difficulty thresholds and pool construction rules.

\paragraph{Training.} The main experiments use Qwen3-VL-8B-Instruct. All five arms start from $\St$, obtained after 100 GRPO updates on the old pool, and train for up to eight additional windows with the same optimization settings. We also evaluate the method on MiMo-VL-7B-SFT.

\paragraph{Checkpoint notation.}
$S_0$ denotes the public starting model. In our main trajectory, $S_{ij}$ is the accepted model after the $j$th continuation window on the $i$th data pool. Each continuation window contains 100 updates. The Qwen and MiMo main trajectories use three windows on each of their first three pools, so continuation indices $w=1,2,3$ correspond to $S_{11},S_{12},S_{13}$, indices $w=4,5,6$ to $S_{21},S_{22},S_{23}$, and indices $w=7,8,9$ to $S_{31},S_{32},S_{33}$. This correspondence describes the observed refresh decisions, rather than imposing a fixed three-window rule. Qwen additionally has the 100-update initialization from $S_0$ to $\St$ before these continuation windows. In comparisons between refresh strategies, column and horizontal-axis labels refer to the training progress of the main trajectory; control arms are evaluated after the same number of continuation updates and need not use the same pool or within-pool count. Earlier independent experiments retain their original window indices because their refresh schedules differ.

\paragraph{Evaluation protocols.} The Qwen and MiMo main trajectories, refresh-policy and replacement-selection comparisons, and earlier-chain benchmark results use the same eleven full test sets. Macro11 equally averages their accuracies. Following MM-Zero \citep{li2026mmzero}, we additionally report LLM-judged accuracy (\lj), using Qwen2.5-14B-Instruct to compare extracted final answers with reference answers. Policy comparisons align training updates and output limits. The length-budget comparisons report their matched fixed-weight evaluation on the same eleven-benchmark suite, comparing each trained checkpoint, including released MM-Zero v5 \citep{li2026mmzero}, with the base model at the same output limit. Scoring details are given together in Appendix~\ref{app:eval}.

\section{Results}

\subsection{Performance across successive training windows}\label{sec:main-eleven}

{Tables~\ref{tab:main-eleven-qwen} and~\ref{tab:main-eleven-mimo}} report full-test-set results on the same eleven benchmarks, selected from the Qwen trajectory, for each backbone. $S_0$ denotes the public starting model. For Qwen and MiMo, in $S_{ij}$, $i$ indexes the active data pool and $j$ indexes its recorded continuation window. Scores are evaluated at an 8,192-token response limit. The Macro11 row is an equal-weight average over these eleven benchmarks; Pool mean additionally averages the three displayed columns within each pool. $S_{32}$ is the eight-window endpoint; $S_{33}$ is a subsequent one-window continuation and is included in the displayed Pool~3 mean. A dash denotes an unavailable result.

\begin{table}[!htbp]
\caption{Qwen3-VL-8B-Instruct on the eleven full benchmark sets (8192 tokens). Columns follow the $S_{ij}$ notation defined in the text. Pool mean averages the three displayed columns within each pool before rounding.}
\label{tab:main-eleven-qwen}
\centering
\paperdensefont
\begin{tabularx}{\linewidth}{@{}l*{11}{R}@{}}
\toprule
Benchmark & $n$ & $S_0$ & \multicolumn{3}{c}{Pool 1} & \multicolumn{3}{c}{Pool 2} & \multicolumn{3}{c}{Pool 3} \\
\cmidrule(lr){4-6}\cmidrule(lr){7-9}\cmidrule(l){10-12}
 & & & $S_{11}$ & $S_{12}$ & $S_{13}$ & $S_{21}$ & $S_{22}$ & $S_{23}$ & $S_{31}$ & $S_{32}$ & $S_{33}$ \\
\midrule
MMSI & 1000 & 27.10 & 27.90 & 26.70 & 28.10 & 28.50 & 28.70 & 30.70 & 28.80 & 29.30 & 31.00 \\
MathVerse & 3940 & 62.03 & 62.01 & 62.39 & 62.46 & 63.40 & 64.06 & 63.38 & 63.25 & 63.48 & 63.81 \\
MathVision & 3040 & 45.92 & 47.27 & 47.20 & 47.66 & 48.98 & 50.07 & 50.62 & 51.45 & 51.45 & 52.53 \\
MMMU-Pro (4) & 1592 & 58.42 & 59.23 & 58.79 & 59.74 & 60.24 & 62.63 & 62.06 & 61.31 & 63.19 & 63.00 \\
VisNumBench & 1913 & 49.03 & 52.74 & 53.84 & 54.47 & 55.36 & 54.68 & 55.62 & 54.05 & 55.99 & 56.19 \\
MMMU-Pro (10) & 1592 & 50.75 & 51.19 & 52.64 & 52.32 & 52.70 & 53.45 & 52.32 & 54.27 & 53.96 & 54.52 \\
MMMU-Pro (vision) & 1730 & 50.06 & 49.36 & 50.46 & 50.58 & 51.62 & 51.21 & 51.04 & 51.39 & 53.76 & 52.72 \\
MMMU & 895 & 60.56 & 62.68 & 62.57 & 63.24 & 65.25 & 65.14 & 64.25 & 65.36 & 64.36 & 65.92 \\
ChartQA & 2500 & 81.24 & 82.48 & 82.20 & 81.88 & 84.40 & 83.32 & 84.52 & 84.28 & 84.96 & 85.00 \\
PhyX & 3000 & 63.50 & 66.93 & 67.07 & 68.30 & 70.00 & 70.97 & 72.03 & 71.73 & 74.17 & 73.33 \\
SCEMQA-MC & 840 & 74.52 & 77.02 & 76.90 & 77.50 & 77.74 & 77.14 & 77.86 & 76.79 & 76.90 & 77.86 \\
\midrule
\rowcolor{cBox}
Macro11 & -- & 56.65 & 58.08 & 58.25 & 58.75 & 59.84 & 60.12 & 60.40 & 60.24 & 61.05 & 61.44 \\

\midrule
Pool mean & -- & 56.65 & \multicolumn{3}{c}{\textbf{58.36}} & \multicolumn{3}{c}{\textbf{60.12}} & \multicolumn{3}{c}{\textbf{60.91}} \\
\bottomrule
\end{tabularx}%

\end{table}

\begin{table}[!htbp]
\caption{MiMo-VL-7B-SFT on the eleven full benchmark sets (8192 tokens). Columns follow the $S_{ij}$ notation defined in the text. Pool mean averages the three displayed columns within each pool before rounding.}
\label{tab:main-eleven-mimo}
\centering
\paperdensefont
\begin{tabularx}{\linewidth}{@{}l*{11}{R}@{}}
\toprule
Benchmark & $n$ & $S_0$ & \multicolumn{3}{c}{Pool 1} & \multicolumn{3}{c}{Pool 2} & \multicolumn{3}{c}{Pool 3} \\
\cmidrule(lr){4-6}\cmidrule(lr){7-9}\cmidrule(l){10-12}
 & & & $S_{11}$ & $S_{12}$ & $S_{13}$ & $S_{21}$ & $S_{22}$ & $S_{23}$ & $S_{31}$ & $S_{32}$ & $S_{33}$ \\
\midrule
MMSI & 1000 & 31.50 & 30.20 & 31.30 & 30.10 & 30.00 & 29.00 & 29.30 & 31.10 & 30.70 & 30.00 \\
MathVerse & 3940 & 70.10 & 72.34 & 72.23 & 72.11 & 71.88 & 71.80 & 71.42 & 71.83 & 72.39 & 71.93 \\
MathVision & 3040 & 53.95 & 54.51 & 55.69 & 55.46 & 55.62 & 55.82 & 56.35 & 56.78 & 56.41 & 55.30 \\
MMMU-Pro (4) & 1592 & 57.10 & 59.23 & 59.92 & 59.80 & 59.67 & 60.24 & 60.80 & 60.05 & 59.42 & 59.74 \\
VisNumBench & 1913 & 48.93 & 48.77 & 48.30 & 48.09 & 49.71 & 49.76 & 50.91 & 50.03 & 49.24 & 49.71 \\
MMMU-Pro (10) & 1592 & 49.94 & 50.06 & 51.01 & 52.01 & 51.19 & 52.58 & 52.32 & 52.95 & 52.01 & 51.44 \\
MMMU-Pro (vision) & 1730 & 48.21 & 48.73 & 49.60 & 49.65 & 48.96 & 49.36 & 50.00 & 48.15 & 49.19 & 48.73 \\
MMMU & 895 & 60.56 & 62.91 & 63.24 & 62.23 & 62.23 & 63.46 & 64.47 & 61.68 & 62.23 & 61.79 \\
ChartQA & 2500 & 88.12 & 87.88 & 88.52 & 88.16 & 88.28 & 88.44 & 88.08 & 88.20 & 88.32 & 88.72 \\
PhyX & 3000 & 65.67 & 67.27 & 68.27 & 69.13 & 68.97 & 69.80 & 70.57 & 69.57 & 69.13 & 68.87 \\
SCEMQA-MC & 840 & 75.83 & 77.62 & 76.90 & 77.14 & 77.86 & 77.50 & 77.26 & 76.90 & 78.81 & 77.62 \\
\midrule
\rowcolor{cBox}
Macro11 & -- & 59.08 & 59.96 & 60.45 & 60.35 & 60.40 & 60.71 & 61.04 & 60.66 & 60.71 & 60.35 \\

\midrule
Pool mean & -- & 59.08 & \multicolumn{3}{c}{\textbf{60.25}} & \multicolumn{3}{c}{\textbf{60.72}} & \multicolumn{3}{c}{\textbf{60.57}} \\
\bottomrule
\end{tabularx}%

\end{table}

\subsection{Continued gains with fewer fresh tasks}\label{sec:main}

\begin{figure}[!htbp]
\centering
\includegraphics[width=1.0\linewidth]{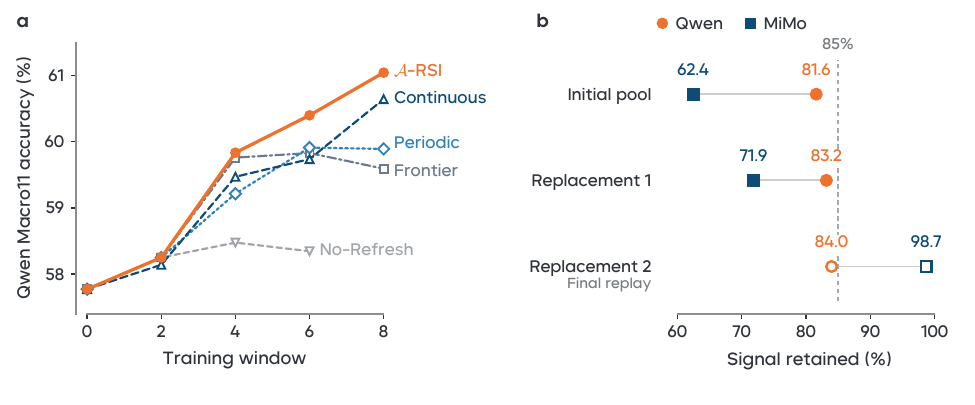}
\caption{{Refresh policies and training feedback. (a) Qwen Macro11 on the eleven-benchmark suite (8192 tokens), with window 0 at the initialized checkpoint $\St$. (b) Within-pool signal retention. Paired points compare models against the 85\% signal threshold. Open markers denote final-round replay. The controller also checks accuracy gains.}}
\label{fig:learning}
\end{figure}

Figure~\ref{fig:learning}(a) compares the five refresh policies on the eleven-benchmark suite. Starting from $\St$ at 57.77, \arsi reaches 59.84, 60.40, and 61.05 at continuation windows 4, 6, and 8. Original-pool continuation scores 58.48 and 58.35 at windows 4 and 6, then halts before producing a valid final checkpoint. The refreshing arms continue learning beyond this plateau.

\paragraph{Score and data use.}\label{sec:i2}
At window 8, \arsi scores 61.05 after introducing 800 new tasks. \armCont scores 60.64 with 3,200, \armPeriodic 59.89 with 1,200, and \armFront 59.59 with 800. \arsi is $+0.40$ points above every-window refresh while using one quarter as many new tasks. The shared initial pool is excluded from these counts. Table~\ref{tab:ablation-refresh} gives the matched-update comparisons under the same eleven-benchmark protocol.

\subsection{Training signals identify repeated pool saturation}\label{sec:i1}

{We first examine the executed refresh decisions in the Qwen run.} On the old pool, the number of tasks with nonzero group advantages falls from 239 to 195 over three windows, giving $U_w/U_{w-2}=.816$ and triggering replacement. On the first new pools, the detector fires again after three windows, with ratios .832 for \arsi and .826 for \armFront. The accuracy condition never fires alone. The first new random pool starts at training accuracy .35, compared with .57 at the end of the old pool, exposing tasks the model has not yet mastered.

Keeping the old pool eventually exhausts its useful signal. At $S_{31}$, \armNR records 35 zero-signal warnings and only 94 of 100 normal gradient steps. Its health gate fails twice and training halts. Before that halt, at the matched window-6 budget, \arsi scores 60.40 against 58.35 for continued reuse under the eleven-benchmark protocol. The observed refreshes coincide with a three-window schedule. Section~\ref{sec:ablation-refresh} separates what the policy comparison establishes from the timing question.

{Section~\ref{sec:cross-model-refresh-replay} connects these executed refreshes with the final-round replay to examine how the same rule responds across stages and models.}

The endpoint, \lj, truncation, and visual-type readouts use the same eleven-benchmark scope and are collected in Appendix~\ref{sec:ablation-twelve}.

\section{{Refresh Ablations and Adaptive Decisions}}\label{sec:ablation}

Refresh-policy and replacement-selection ablations use Qwen3-VL-8B-Instruct, the same eleven full benchmark sets as the main trajectories, and an 8,192-token output limit. Macro11 is the equal-weight average of the eleven benchmark accuracies. Comparisons align policies by the number of training updates corresponding to the indicated \arsi state, rather than by each policy's pool index. Runs without a valid final evaluation are marked with a dash. We then examine how the same controller responds to the training feedback of Qwen and MiMo across successive pools.

\subsection{Fresh data and refresh timing}\label{sec:ablation-refresh}\label{sec:ablation-timing}

Table~\ref{tab:ablation-refresh} reports complete eleven-benchmark evaluations for all available refresh-policy checkpoints. At $S_{23}$, \arsi reaches 60.40 versus 58.35 for original-pool continuation, a $+2.05$-point difference at the same update budget. The no-refresh arm stopped at $S_{31}$ after failing its training health check and has no $S_{32}$ result; that cell is marked with a dash. Each policy is represented by one training run; evaluation items and intermediate checkpoints are not independent training replications.

\begin{table}[!htbp]
\caption{Qwen refresh policies at matched updates. Scores are Macro11 on the eleven-benchmark suite (8192 tokens). New tasks exclude the shared initial pool. Differences are computed before rounding. A dash indicates an unavailable evaluation result.}
\label{tab:ablation-refresh}
\centering
\paperdensefont
\begin{tabular*}{\linewidth}{@{\extracolsep{\fill}}l l lrrrr@{}}
\toprule
Variant & Refresh rule & New-pool selection & $S_{21}$ & $S_{23}$ & $S_{32}$ & New tasks by $S_{32}$ \\
\midrule
\rowcolor{cBox}
\arsi & Training-log trigger & Random & 59.84 & 60.40 & 61.05 & 800 \\
\armNR & Never & None & 58.48 & 58.35 & -- & 0 \\
\armCont & Every window & Adaptive difficulty & 59.47 & 59.74 & 60.64 & 3,200 \\
\armPeriodic & Every two windows & Frontier & 59.22 & 59.91 & 59.89 & 1,200 \\
\armFront & Training-log trigger & Frontier & 59.76 & 59.83 & 59.59 & 800 \\
\bottomrule
\end{tabular*}%

\end{table}

At $S_{32}$, \arsi exceeds \armPeriodic by $+1.15$ points and \armCont by $+0.40$ points while introducing fewer new tasks. These comparisons change both selection and timing, so they do not isolate either mechanism. With frontier selection held fixed, detector-triggered refresh instead trails periodic refresh by $0.30$ points. Moreover, the first two realized random-refresh decisions coincide with a three-window schedule. The results therefore support the implemented fresh-pool policy in this run, without establishing that the detector selects a better time than a fixed schedule.

\subsection{Replacement selection}\label{sec:pick}

The replacement-selection variants use the same eleven full benchmarks and accuracy scoring (Table~\ref{tab:three}). Failure-prioritized, learnable-band, and always-wrong selection all trail their every-window random control. The tested ladder, always-increase, truncation-aware, and all-tier controllers also have lower point estimates than that shared control. These variants share control runs and training histories; they are not independent replications. Table~\ref{tab:three-twelve} reports the corresponding stable-format subset on the same eleven benchmarks.

\begin{table}[!htbp]
\centering
\begin{minipage}{.82\linewidth}
\caption{Replacement-selection ablations. Scores are Macro11 at 8192 tokens on the eleven-benchmark suite. Each difference uses the control in its block and is computed before rounding.}
\label{tab:three}
\papertablefont
\begin{tabularx}{\linewidth}{@{}Xrr@{}}
\toprule
Replacement selection & Macro11 & Difference \\
\midrule
\multicolumn{3}{@{}l}{\textit{Four windows, every-window refresh}} \\
\addlinespace[3pt]
\rowcolor{cBox}
Random control & \textbf{60.29} & -- \\
Failure-prioritized sampling & 57.67 & $-2.63$ \\
Learnable-band sampling & 58.39 & $-1.90$ \\
Always-wrong sampling & 57.89 & $-2.40$ \\
Ladder controller & 59.47 & $-0.82$ \\
Always-increase controller & 59.81 & $-0.48$ \\
Truncation-aware controller & 60.24 & $-0.05$ \\
All-tier sampling & 59.10 & $-1.19$ \\
\midrule
\multicolumn{3}{@{}l}{\textit{Eight windows, detector-triggered refresh ($S_{32}$)}} \\
\addlinespace[3pt]
\rowcolor{cBox}
Random control (\arsi) & \textbf{61.05} & -- \\
Frontier selection & 59.59 & $-1.45$ \\
\bottomrule
\end{tabularx}
\end{minipage}
\end{table}

\subsection{{Adaptive decisions across stages and models}}\label{sec:adaptive-decisions}\label{sec:cross-model-refresh-replay}

{
The controller adapts through its decisions while keeping the observation period and thresholds fixed. Figure~\ref{fig:learning}(b) compares the first and third windows on each pool. For MiMo, the distinct-prompt advantage count falls from 245 to 153 on the initial pool and from 199 to 143 on the first replacement pool. These retain 62.45\% and 71.86\% of the initial counts, below the 85\% threshold, and both stages trigger replacement. Their accuracy increases also exceed the ten-point threshold. On the second replacement pool, the count remains nearly unchanged at 152 to 150, retaining 98.68\%, while accuracy rises by 6.63 points. The final-round replay therefore keeps this pool. Equal periods of reuse produce different decisions as MiMo's training feedback changes.
\par}

{
The same rule also distinguishes the two models at the same stage. On Qwen's second replacement pool, the count falls from 181 to 152, retaining 83.98\%, while accuracy rises by 6.59 points. The replay requests replacement for Qwen and continued reuse for MiMo. A fixed three-window schedule would replace both. Thus, the minimum observation period makes a decision eligible, while the observed training state determines the action. The first two decisions for each model were executed during training. The final-stage actions are offline next-window decisions. Raw counts, accuracy changes, and decision records are collected in Appendix~\ref{app:controller-replay}.
\par}

\FloatBarrier

\section{Analysis}\label{sec:analysis}

We examine reward shortcuts and pipeline validity through controlled reward scoring, independent checks of generated answers, and qualitative correction examples across eleven benchmarks. The generated-task audit $\mathcal{A}$ contains 1,000 actually consumed training tasks for independent answer checks and controlled reward-scoring tests. The external examples illustrate changes in completed answers; aggregate effects are assessed separately using the full-suite and outcome-independent output audits. The generated-task and reward-check protocols are given in Appendix~\ref{app:generated-check-protocol}.

\subsection{Reward hacking and non-truncated corrections}\label{sec:reward-hacking}\label{sec:budget}

\paragraph{A controlled reward-scoring stress test.}
For every generated task in $\mathcal{A}$, we construct seven response variants, including correctly and incorrectly boxed answers, the same answer with different finish reasons, and unboxed answers. We score all 7,000 variants with the actual correctness, format, and termination functions used by training. The task reference and response content are held fixed where appropriate, so format and termination changes can be isolated (Table~\ref{tab:reward-stress-1000}). These are counterfactual scoring probes, not generated model responses or a reward-removal training ablation.

\begin{table}[!htbp]
\caption{Controlled reward scoring on 1,000 independently checked generated tasks. The second and third checks retain an incorrect answer, isolating auxiliary-reward increases without a correctness increase. Full response variants and component rewards are retained.}
\label{tab:reward-stress-1000}
\centering
\papertablefont
\begin{tabular}{@{}p{.74\linewidth}r@{}}
\toprule
Controlled check & Tasks passing \\
\midrule
Correct committed answer ranks above incorrect committed answer & 1,000 / 1,000 \\
Canonical formatting raises reward with correctness and termination fixed & 1,000 / 1,000 \\
Normal commitment raises reward with correctness and formatting fixed & 1,000 / 1,000 \\
\bottomrule
\end{tabular}

\end{table}

\paragraph{Adversarial answer-checker tests.}
We extend the same 1,000 references to 18 response variants, with expected correctness specified for 16 variants. The production checker fails 220 of the 16,000 specified checks: on 110 choice questions, it accepts both a gold-prefixed ambiguous answer and a gold option followed by unrelated text. An isolated candidate that requires a complete terminal option passes all 16,000 checks. On 16,000 actual logged responses from five training windows, 3,760 are choice responses; the candidate changes one score. These results expose a parser vulnerability without establishing widespread exploitation by the trained model. The candidate has not been substituted into the reported training runs. Appendix~\ref{app:mechanism-validation} gives the checks and prospective training comparison.

\paragraph{Auxiliary rewards cannot reverse the correctness ranking.}
Let $r_c,r_w$ receive correctness rewards one and zero, respectively. Since the auxiliary reward lies in $[-\lambda_{\mathrm{stop}},\lambda_{\mathrm{fmt}}]$, Eq.~\eqref{eq:reward} gives
\begin{equation}
 R(r_c)-R(r_w)\geq 1-\lambda_{\mathrm{fmt}}-\lambda_{\mathrm{stop}}>0,
 \label{eq:reward-dominance}
\end{equation}
where the final inequality follows from our fixed coefficients. Thus, auxiliary terms cannot make a response with zero correctness reward outrank one with unit correctness reward, assuming the reference and answer checker are valid. This ranking property does not prevent optimization within the all-wrong class: the stress test also confirms that formatting or committing a wrong answer can increase reward without improving correctness. We therefore examine non-truncated corrections separately rather than equating higher reward with better reasoning.

\paragraph{MiMo: qualitative examples of non-truncated correction.}
The MiMo examples compare $S_0$ with $S_{32}$. Both responses finish normally, while the earlier incorrect answer becomes correct under both original and format-corrected answer checks. Such a transition illustrates a change beyond simply finishing a previously truncated response. It does not by itself establish that the reasoning steps are valid or exclude guessing; those judgments require the original image and full response. These examples are used qualitatively, without estimating how frequently this change occurs or presenting the automated checks as completed human annotations.

Full-suite output decompositions, an outcome-independent random sample, and fixed-weight length-budget comparisons are reported in Appendices~\ref{app:full-output-decomposition}, \ref{app:random1000}, and~\ref{app:reward-budget-audits}, respectively. Those analyses retain both corrections and regressions.

\begin{figure}[!htbp]
\centering
\includegraphics[width=1.0\linewidth]{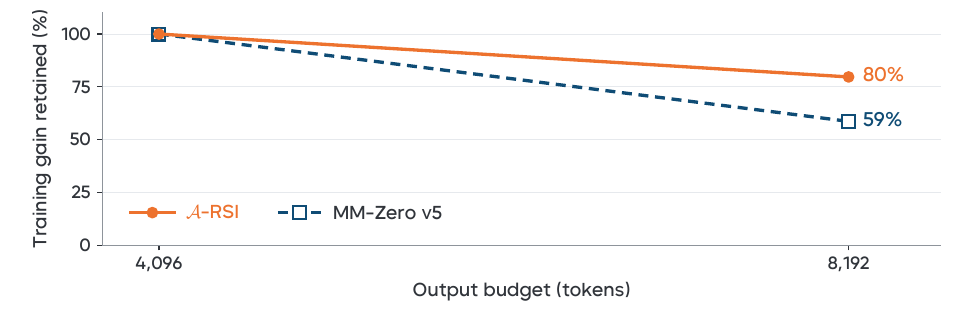}
\caption{\textbf{Training gains retained at twice the output budget.} Gains over base on the eleven-benchmark suite are normalized to each method's 4096-token gain (4.08 points for \arsi and 4.18 for MM-Zero v5). Base and trained checkpoints use the same output budget.}
\label{fig:budget-correction}
\end{figure}

\paragraph{Training gains persist at a larger output budget.}
When the generation limit doubles from 4096 to 8192 tokens, \arsi retains 80\% of its gain over base, compared with 59\% for MM-Zero v5 (Figure~\ref{fig:budget-correction}). Each trained checkpoint is compared with the base model at the same limit on the eleven-benchmark suite. At 8192 tokens, \arsi remains 3.25 points above base, versus 2.45 for MM-Zero v5. Its training gain is thus less sensitive to the output budget.

\FloatBarrier

\subsection{Pipeline validity: verifiable data and experience-driven refresh}\label{sec:pipeline-validity}

\paragraph{An independent check of generated answers.}
For $\mathcal{A}$, we randomly sample 200 tasks without replacement from each of five supported generator groups in actually consumed replacement pools, fixing membership before recomputing any answers. Independent solvers agree with all 1,000 reference answers. Training gold/solution values agree with the selected source records, and image-file hashes agree with the recorded hashes on all 1,000 tasks. The five groups cover geometry, charts, tables, and counting. The group-level counts and sampling protocol are given together in Appendix~\ref{app:generated-check-protocol}, which also states the excluded generator groups.

\paragraph{A conditional specification for task validity.}
For a template $\tau$ and sampled parameters $z$, let $s_\tau(z)$ be the underlying configuration, $E_\tau$ its answer executor, $R_\tau$ its renderer, and $Q_\tau$ its question constructor. The generation contract is
\begin{equation}
 \begin{aligned}
  I&=R_\tau(s_\tau(z)),\qquad q=Q_\tau(s_\tau(z)),\\
  a&=E_\tau(s_\tau(z)).
 \end{aligned}
 \label{eq:validity-contract}
\end{equation}
Thus, image, question, and answer share one underlying configuration. Let $V_\tau(I,q,a,z)$ denote the generator-specific construction checks, $F(x)$ the freshness check, $C(x)$ the contamination screen, and $H$ the diagnostic holdout. The admission rule can be specified as
\begin{equation}
 x\in\mathcal{C}\ \Longrightarrow\
 V_\tau(x)=1\ \land\ F(x)=1\ \land\ C(x)=1\ \land\ x\notin H.
 \label{eq:admission-contract}
\end{equation}
We use an SMT solver to check four properties of the specified model: correctness-ranking protection, within-pool lag indexing, bounded one-step difficulty updates, and freshness across successive admitted pools (Appendix~\ref{app:mechanism-validation}). These are machine-checked properties of the specification, not verification of the full Python implementations or renderer. It certifies only the properties tested by those checks. Interpreting $a$ as the answer to the rendered question additionally assumes that the executor is correct and that rendering and wording preserve the configuration without ambiguity. Independent recomputation tests that assumption; repeating the original answer function would not provide an independent audit.

For the finite audit set $\mathcal{A}$, the checks establish agreement with independently recomputed answers and consistency of source labels and image hashes. Every audited record satisfies those checks. Interpreting the recomputed answer as the answer to the rendered question still relies on rendering and wording preserving the underlying configuration. The broader generator audits and remaining ambiguities are retained in Appendix~\ref{app:pipeline-audit-detail}. All 1,000 image files also decode, meet the minimum-dimension check, and are nonblank; none of the task IDs overlaps the diagnostic holdout. Independent audit checks detect all 4,000 injected reference, image-byte, source-question, and truncated-image defects. These injections test the audit checks rather than all production admission paths.

\paragraph{Rendered-task consistency on the same fixed sample.}
Additional task-specific checks pass for all 1,000 records, including target uniqueness, admissible values, and option consistency where applicable. Replaying the frozen generators reproduces 800 whole images byte for byte under the recorded audit environments. For the remaining 200 referring charts, the plot pixels and the category and numeric-axis labels agree with the source, although whole-image reproduction remains unresolved because of font differences. These checks strengthen the finite quality audit without certifying every visual or linguistic property (Appendix~\ref{app:mechanism-validation}).

\paragraph{Does fresh-pool training improve the learner?}
The complete eleven-benchmark format-corrected scoring comparison in Table~\ref{tab:ablation-refresh} evaluates the downstream consequence of the implemented refresh policy: at matched $S_{23}$ progress, fresh-pool training scores 60.40 against 58.35 for original-pool continuation, a $+2.05$-point difference. Both arms share the trajectory through $S_{13}$ and receive the same number of subsequent updates. Pool replacement also changes the active pool from 500 to 400 tasks, so this is evidence for the combined policy in one run rather than an isolated effect of freshness, validated labels, or detector timing. The output-stratum analysis uses these same eleven-benchmark outputs and decomposes this Macro11 difference into a $-0.60$-point contribution from answers completed by both arms and $+2.66$ from the other completion strata. Appendix~\ref{app:fresh-pool-comparison} gives the definitions and comparison scope.

\paragraph{Qwen: qualitative examples of same-progress fresh-pool correction.}
The Qwen examples compare old-pool continuation with fresh-pool training at the same $S_{23}$ training progress. The arms share the trajectory through $S_{13}$, use the same number of subsequent updates, and are evaluated at 8192 tokens. In these examples, both responses finish normally and the incorrect old-pool answer becomes correct under both scoring rules. They illustrate where fresh-pool training changes a completed answer favorably, without identifying which individual training task caused the correction or estimating an overall advantage. The complete-test-split continuation comparison is reported separately in Appendix~\ref{app:fresh-pool-comparison}.

\paragraph{Difficulty and update rules have distinct roles.}
Dependency depth, required reads, and reading mode describe task construction. The main policy samples unused tasks under fixed family quotas and uses training-side signals to decide when to replace the pool. Equation~\eqref{eq:trigger} requires enough accepted windows on that pool; because $n_{\min}>\ell$, the lagged comparison stays within the pool. Replacement resets the history and continues from accepted model weights. These properties make the realized update decision reproducible, but do not prove that reward-signal decline is answer mastery or optimal timing. The component replay in Appendix~\ref{app:controller-replay} distinguishes total-reward variation from correctness-only variation. Exhaustive integer-threshold tests, bounded difficulty-update tests, and the actual-window replay check conformance to these rules; they do not establish that a trigger implies mastery or that the selected timing is optimal (Appendix~\ref{app:mechanism-validation}).

\paragraph{Connecting cases to mechanisms.}
For the retained external cases, the review procedure checks the reference against the original image, locates the earlier error, and asks whether the later response repairs it with valid steps. Candidate mechanisms include visual reading, omitted conditions, composition, arithmetic, and answer presentation. {These mechanism categories remain unassigned for the qualitative examples.} The generated-task audit supports the checked data properties, the scoring probes establish reward behavior, and the external examples expose favorable answer changes; none identifies which individual training task caused a downstream correction.

\FloatBarrier

\Needspace{10\baselineskip}
\section{Case Study: Finishing and Correcting}\label{sec:cases}

\begin{figure}[p]
\centering
\begin{minipage}[t]{.48\linewidth}
\vspace{0pt}
\begin{casecard}{caseBefore}{A1. An answer, but no ending!}
\textbf{PhyX \#1314: CO bond length}\par\smallskip
\casemodelstage{Qwen3-VL-8B-Instruct}
\caseimage[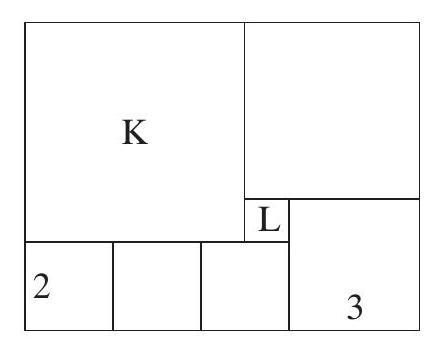]{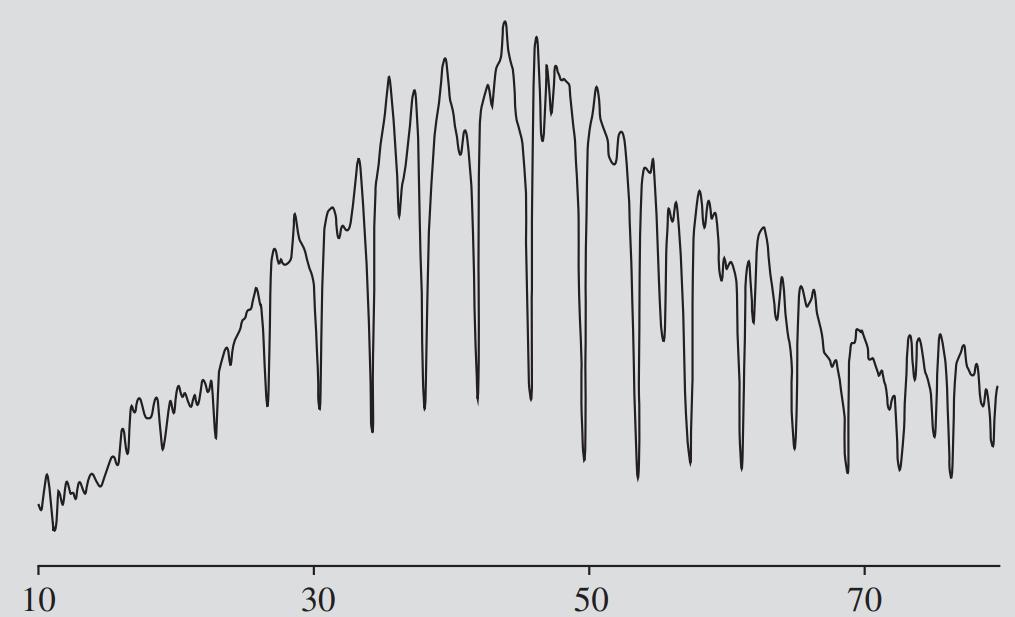}
\textbf{Task.} Use the rotational spectrum to determine the CO bond length. Gold: option C, $1.1\times10^{-10}$ m.

\medskip
{\sffamily\bfseries\color{caseBefore}BEFORE: circles back, then gets cut off}\par
The base response mentions the known CO bond length but repeatedly revisits the calculation. It reaches the limit without a final answer.

\medskip
{\sffamily\bfseries\color{caseAfter}AFTER: commits and stops}\par
The trained response invokes the known bond length, selects C, and ends normally:
\begin{quote}\small
``I recall that the bond length of CO is 1.128 \AA''\par
\hfill $\boxed{\mathrm{C}}$
\end{quote}
\casetokens{8192}{1759}
\medskip
\textbf{What changed?} Less unresolved repetition and an explicit final answer. The spectrum is still misread, so this example does not establish better visual reasoning.
\end{casecard}
\end{minipage}\hfill
\begin{minipage}[t]{.48\linewidth}
\vspace{0pt}
\begin{casecard}{caseBefore}{A2. Stuck, then able to stop!}
\textbf{MATH-Vision \#703: square areas}\par\smallskip
\casemodelstage{Qwen3-VL-8B-Instruct}
\caseimage[assets/cases/sci_phyx_mc_1314_0.png]{assets/cases/mathvision_703_0.png}
\textbf{Task.} A rectangle contains seven squares. How many times is the area of the largest square K that of the smallest L? Gold: 25.

\medskip
{\sffamily\bfseries\color{caseBefore}BEFORE: tries one layout after another}\par
The base response repeatedly reconstructs incompatible layouts. It reaches the limit without a final answer.

\medskip
{\sffamily\bfseries\color{caseAfter}AFTER: gives an answer within budget}\par
The trained response still struggles with the layout, but eventually commits:
\begin{quote}\small
``I think I have to accept that the area of K is 25 times the area of L.''\par
\hfill $\boxed{25}$
\end{quote}
\casetokens{8192}{5140}
\medskip
\textbf{What changed?} Completion and a correct final answer. The derivation remains inconsistent; this is an example of finishing, not a verified reasoning improvement.
\end{casecard}
\end{minipage}
\caption{\textbf{From truncation to completion.} Both base decodes reach the limit for each item. $S_{32}$ finishes correctly in both decodes: 1,759/3,437 tokens for A1 and 5,140/2,515 for A2. Shorter, completed responses can improve benchmark accuracy even when the reasoning remains flawed. These observations alone do not identify which reward component caused the change.}
\label{fig:cases-finishing}
\end{figure}

\begin{figure}[p]
\centering
\begin{minipage}[t]{.48\linewidth}
\vspace{0pt}
\begin{casecard}{caseBlue}{B1. The missing slice matters!}
\textbf{MathVerse \#2042: perimeter}\par\smallskip
\casemodelstage{Qwen3-VL-8B-Instruct}
\caseimage[]{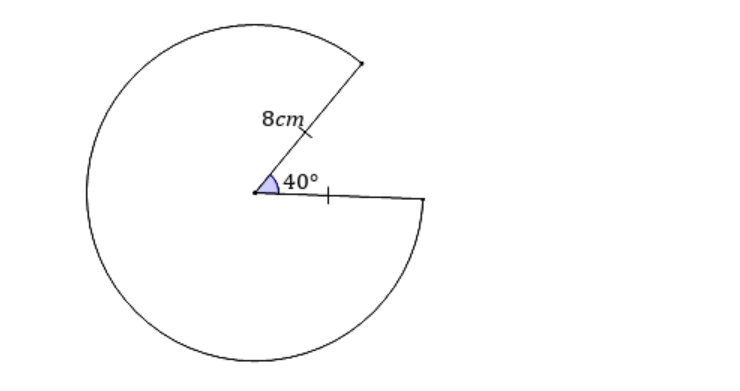}
\textbf{Task.} Find the perimeter to two decimal places. Gold: $60.68$ cm.

\medskip
{\sffamily\bfseries\color{caseBefore}BEFORE: measures the missing piece}\par
The base response treats the labeled $40^\circ$ as the angle of the required sector. It finishes with $\boxed{21.59}$, the perimeter of the small sector.

\medskip
{\sffamily\bfseries\color{caseAfter}AFTER: identifies the remaining shape}\par
``it is a circle with a sector of 40 degrees removed.'' The trained response uses the $320^\circ$ arc and both radii:
\[
 P=2(8)+\frac{320}{360}\,2\pi(8)
   \approx\boxed{60.68}.
\]
\casetokens{1634}{374}
\medskip
\textbf{What changed?} The interpretation of the diagram is corrected, followed by a consistent calculation. Both base and trained responses finish normally, so avoiding truncation cannot explain this item's score change.
\end{casecard}
\end{minipage}\hfill
\begin{minipage}[t]{.48\linewidth}
\vspace{0pt}
\begin{casecard}{caseBlue}{B2. Look at the bottom!}
\textbf{CharXiv \#494: spatial color reading}\par\smallskip
\casemodelstage{Qwen3-VL-8B-Instruct}
\caseimage[assets/cases/mathverse_2042_0.png]{assets/cases/sci_charxiv_494_0.png}
\textbf{Task.} In the parallel column, which W subscript has the brighter sphere at its bottom center? Gold: 1.

\medskip
{\sffamily\bfseries\color{caseBefore}BEFORE: reverses the color gradient}\par
The base response describes the bottom of $W_2$ as bright yellow and finishes with $\boxed{2}$.

\medskip
{\sffamily\bfseries\color{caseAfter}AFTER: reads color at the right position}\par
The trained response correctly locates the dark region in $W_2$:
\begin{quote}\small
``The bottom center of the sphere is dark purple''\par
\hfill $\boxed{1}$
\end{quote}
The orange bottom of $W_1$ is brighter.
\casetokens{1403}{187}
\medskip
\textbf{What changed?} The visual description and answer both become correct. Both versions finish normally; the correction concerns where the colors appear in the image.
\end{casecard}
\end{minipage}
\caption{\textbf{From completed-but-wrong to completed-and-correct.} Both independent base decodes give the same wrong answers (21.59 and 2), while both $S_{32}$ decodes give the correct answers (60.68 and 1). These corrections also persist in the seed-0 evaluations from $S_{21}$ through $S_{33}$. Individual successes do not establish a net improvement over all originally completed items.}
\label{fig:cases-correcting}
\end{figure}

We distinguish two observable changes: a response that previously reached the token limit now finishes (Figure~\ref{fig:cases-finishing}), and a response that previously finished with an incorrect answer now becomes correct (Figure~\ref{fig:cases-correcting}). Each figure illustrates two examples by comparing the Qwen base model with $S_{32}$. All responses use the same 8,192-token evaluation limit. The panels show decode seed 0; an independent decode seed reproduces all four endpoint transitions. These are selected qualitative examples, rather than estimates of how frequently either change occurs.

Appendix~\ref{app:additional-cases} provides ten further completed-answer corrections across eight main-result benchmarks for Qwen and MiMo.

\section{Discussion}

\paragraph{The simple version of the loop wins.} The cheapest policy we found for keeping a zero-data loop alive is to let the model's training log say when the pool is spent and then supply random fresh tasks. Neither of the refinements we tried helped, and one made things worse. We read this as a statement about this setting, in which the tasks are programmatic and the checkpoint is a plateau.

\paragraph{Measured state is for diagnosis.} The ladder is useful for something other than selection. Because difficulty is measured, we can say what the model has learned, which tiers saturate first, and where a mismatch occurs: for example, three of six ladders never reached the promotion threshold, and hard multi-hop tasks are largely length exhaustion. Self-assessed difficulty (AUC near 0.50) could not support these statements.

{
\paragraph{Adaptation through observable training state.}
The controller connects experience reuse to the model's ongoing optimization. MiMo's changing decisions across pools and the divergent Qwen--MiMo final-round replay show why elapsed training alone does not specify the action. A common fixed rule can respond to these differences through model-generated training feedback, without changing its thresholds or consulting evaluation performance.
\par}

Limitations are discussed in Appendix~\ref{sec:limits}.

\section{Conclusion}

An RSI loop stalls first where its data runs out, and its gains can look larger than they are. \arsi combines program-based generators whose answers are computed, a detector that reads only the model's own training log and says when a pool of 400 tasks is spent, a draw of 400 fresh random tasks when it fires, and audits of length-budget sensitivity, answer formatting, and completed responses. On our main model, eight continuation windows raise Macro11 from 57.77 to 61.05 with 800 new tasks. Under that same eleven-benchmark protocol, the tested difficulty-based, controller-based, and rollout-based selectors do not outperform random selection in their matched comparisons. The length-budget comparisons use the eleven-benchmark suite and matched output limits. Part of a zero-data gain at a tight generation limit is learning to finish, as the untrained base and the released MM-Zero weights show. MiMo's random-refresh endpoint is 60.71 versus 60.01 for continued reuse, a $+0.70$-point difference, and exceeds every-window and periodic refresh by $+0.02$ and $+0.18$, respectively; these small single-run margins do not establish a timing advantage. Each arm was trained once. The next steps are replication across seeds, a fixed-schedule random control for detector timing, 16k evaluation of this run, and a refresh policy for low-entropy models.

\FloatBarrier
\bibliography{main}

\clearpage
\phantomsection
\section*{Acknowledgments}
\label{sec:acknowledgments}
\addcontentsline{toc}{section}{Acknowledgments}

This work uses the MM-Vet dataset, licensed under CC BY-NC 4.0, solely for non-commercial academic evaluation.

\clearpage
\beginappendix
\section{Supplementary Eleven-Benchmark Policy Comparisons}\label{sec:ablation-twelve}

The policy, endpoint, stable-format, and per-benchmark comparisons below use Qwen3-VL-8B-Instruct and the same eleven-benchmark protocol as Section~\ref{sec:ablation}, with \lj as an auxiliary scoring view. Visual-type and yes/no subsets retain only their annotated items within those benchmarks. The default output limit is 8192 tokens; other limits are identified explicitly. Each arm is one training run, and policies are compared after matched numbers of updates.

\subsection{Endpoint, data use, and visual types}

Table~\ref{tab:main} reports endpoint accuracy, data use, truncation, and the yes/no sentinel on the eleven-benchmark suite. At the scheduled even-window 8192-token evaluations, \arsi scores 58.25, 59.84, 60.40, and 61.05. At $S_{32}$ it is 0.40 points above \armCont, 1.15 above \armPeriodic, and 1.45 above \armFront. Its gain per thousand new tasks is 4.09, compared with 0.90, 1.77, and 2.28, respectively. These ratios describe the observed policies, which differ in both timing and selection.

\begin{table}[!htbp]
\caption{$S_{32}$ policy results on the eleven-benchmark suite at 8192 tokens. Parentheses give the gain over $\St$; \lj averages judge accuracy over the same eleven benchmarks. Truncation is pooled over 22,042 items. The sentinel uses the 114 retained yes/no items.}
\label{tab:main}
\centering
\papertablefont
\begin{tabularx}{\linewidth}{@{}l>{\raggedleft\arraybackslash}p{2.6cm}*{4}{R}l@{}}
\toprule
Strategy & Accuracy (gain) & Fresh tasks & Gain/1k & \lj & Trunc.\,\% & Sentinel \\
\midrule
$\St$ (start) & 57.77 & 0 & -- & 54.83 & 14.34 & WATCH \\
\armNR & \multicolumn{1}{l}{\textit{Halted at $S_{31}$}} & 0 & -- & -- & -- & -- \\
\armFront & 59.59 (+1.82) & 800 & +2.28 & 58.19 & 8.14 & OK \\
\armPeriodic & 59.89 (+2.12) & 1,200 & +1.77 & 58.40 & 10.33 & OK \\
\armCont & 60.64 (+2.87) & 3,200 & +0.90 & 59.12 & 8.44 & OK \\
\midrule
\rowcolor{cBox}
\textbf{\arsi} & \textbf{61.05 (+3.27)} & 800 & \textbf{+4.09} & \textbf{59.54} & \textbf{4.16} & OK \\
\bottomrule
\end{tabularx}
\end{table}

Table~\ref{tab:vtype} breaks these results down over the retained, annotated visual-type subsets. Geometry and counting show the largest gains over $\St$, alongside large drops in truncation. On MathVision, \arsi scores 51.45 with 8.36\% truncation, compared with 18--22\% truncation for the other refreshing arms. Per-benchmark results are given in Appendix~\ref{app:perset}.

\begin{table}[!htbp]
\caption{Visual-type results at $S_{32}$, 8192 tokens. The fixed annotations are restricted to benchmarks retained in the eleven-set suite; the science MC sets have no visual-type annotations and are excluded from these type subsets. Accuracies and truncation rates are pooled within each type; bold marks the best refreshing arm.}
\label{tab:vtype}
\centering
\papertablefont
\begin{tabular}{@{}lr rr rr rr r@{}}
\toprule
\multirow{2}{*}{Arm} & \multicolumn{2}{c}{Geometry (4,588)} & \multicolumn{2}{c}{Chart (2,960)} & \multicolumn{2}{c}{Table (1,181)} & \multicolumn{2}{c}{Counting (447)} \\
\cmidrule(lr){2-3}\cmidrule(lr){4-5}\cmidrule(lr){6-7}\cmidrule(l){8-9}
 & \textit{Acc.} & \textit{Trunc.} & \textit{Acc.} & \textit{Trunc.} & \textit{Acc.} & \textit{Trunc.} & \textit{Acc.} & \textit{Trunc.} \\
\midrule
$\St$ (start) & 56.19 & 18.64 & 79.63 & 1.69 & 70.96 & 13.80 & 54.81 & 13.20 \\
\midrule
\armFront & 57.65 & 10.48 & 81.32 & 1.05 & 72.14 & 10.16 & 57.94 & 6.71 \\
\armPeriodic & 57.85 & 12.69 & \textbf{82.23} & 1.11 & 73.58 & 11.35 & 59.06 & 10.07 \\
\armCont & 59.70 & 9.94 & 81.15 & \textbf{0.71} & \textbf{74.26} & 9.40 & 57.94 & 5.82 \\
\rowcolor{cBox}
\arsi & \textbf{59.83} & \textbf{4.80} & 81.66 & 0.81 & 72.99 & \textbf{6.60} & \textbf{61.74} & \textbf{2.46} \\
\rowcolor{cBox}
\textit{Change from $\St$} & $+3.64$ & $-13.84$ & $+2.03$ & $-0.88$ & $+2.03$ & $-7.20$ & $+6.94$ & $-10.74$ \\
\bottomrule
\end{tabular}
\end{table}

\subsection{Fresh data versus continued reuse}\label{sec:ablation-refresh-twelve}

Table~\ref{tab:ablation-refresh-twelve} compares the available common evaluation windows under the current eleven-benchmark protocol. At $S_{23}$, detector-triggered random refresh reaches 60.40, against 58.35 when the original pool is retained, a difference of $+2.05$ points. The no-refresh arm has no accepted $S_{32}$ result, so we do not impute one or compare its earlier score with the other arms' later endpoints. This comparison evaluates the complete refresh intervention, including replacing the original pool with a newly constructed pool; it does not isolate freshness from every difference in pool composition.

\begin{table}[!htbp]
\caption{Refresh-policy comparisons for Qwen3-VL-8B-Instruct on the full eleven-benchmark suite. Columns align each arm by the continuation updates needed to reach the indicated \arsi state. Values are Macro11 accuracy at an 8,192-token limit. New tasks count distinct introduced training instances, not total generation or compute. A dash denotes a missing accepted result; the no-refresh arm stopped at the training budget corresponding to $S_{31}$.}
\label{tab:ablation-refresh-twelve}
\centering
\paperdensefont
\begin{tabular*}{\linewidth}{@{\extracolsep{\fill}}l l lrrrr@{}}
\toprule
Variant & Refresh rule & New-pool selection & $S_{21}$ & $S_{23}$ & $S_{32}$ & New tasks by $S_{32}$ \\
\midrule
\rowcolor{cBox}
\arsi & Training-log trigger & Random & 59.84 & 60.40 & 61.05 & 800 \\
\armNR & Never & None & 58.48 & 58.35 & -- & 0 \\
\armCont & Every window & Adaptive difficulty & 59.47 & 59.74 & 60.64 & 3,200 \\
\armPeriodic & Every two windows & Frontier & 59.22 & 59.91 & 59.89 & 1,200 \\
\armFront & Training-log trigger & Frontier & 59.76 & 59.83 & 59.59 & 800 \\
\bottomrule
\end{tabular*}%

\end{table}

\subsection{Refresh timing}\label{sec:ablation-timing-twelve}

With frontier selection in both arms, detector-triggered refresh scores 59.59 at $S_{32}$, versus 59.89 for refreshing every two windows. This comparison provides no evidence that the tested detector timing improves accuracy. Detector-triggered random refresh and every-window adaptive-difficulty refresh score 61.05 and 60.64, while introducing 800 and 3,200 new tasks, respectively. They differ in both timing and selection, so the $+0.40$ score difference cannot be attributed to timing alone and does not establish equivalence.

In the recorded Qwen run, the first two random-pool replacements occur after three windows on the active pool. Those observed actions coincide with a fixed three-window schedule. Consequently, the existing trajectory does not establish an advantage of history-dependent timing over that schedule.

\subsection{Replacement selection}\label{sec:pick-twelve}

The detector gives the same trigger point to \armFront and \arsi because they share $S_{11}$--$S_{13}$ and the old pool; their replacement pools then differ by frontier versus random selection. On the eleven-benchmark format-corrected scoring readout, \armFront $-$ \arsi is $-0.59$, $-0.89$, $-1.25$, $-1.04$, $-2.67$ at 4096 tokens from $S_{21}$ to $S_{32}$. At 8192 tokens the differences at $S_{21}$, $S_{23}$, and $S_{32}$ are $-0.07$, $-0.57$, and $-1.45$, with stable-format $-1.40$ at $S_{32}$. At its second refresh the frontier controller moved six pointer steps up (hr $+3$, mvS $+2$, mvF $+1$), after which its truncation rose to 8.14\% and its score fell.

Table~\ref{tab:three-twelve} also summarizes completed selection comparisons from earlier configurations. Their point estimates favor the random arm, but they are not independent replications: the probe-selection and ladder-control comparisons share the same random control, and multiple variants share training histories. The pass-rate and ladder selectors tested here do not improve over random replacement under these configurations.

\begin{table}[!htbp]
\caption{Replacement-selection differences on the eleven-benchmark suite, 8192 tokens, against the matched random arm. The first two blocks share the every-window random control after four windows; the last compares with \arsi after eight. Stable applies the same scoring to letter-, number-, and expression-gold items and equally averages the eleven subset accuracies. Differences are descriptive single-run estimates.}
\label{tab:three-twelve}
\centering
\paperdensefont
\begin{tabular}{@{}l lrr@{}}
\toprule
Experiment & How the replacement tasks are chosen & Accuracy & Stable \\
\midrule
Probe selection & 1--7 of 8 correct, image-specific failures first & $-2.63$ & $-2.60$ \\
 & 1--7 of 8 correct (ThinkLite-style) & $-1.90$ & $-1.95$ \\
 & 0 of 8 correct & $-2.40$ & $-2.37$ \\
\midrule
Ladder control & pointer up at pass $\geq.80$, down at $\leq.20$ & $-0.82$ & $-0.90$ \\
 & every pointer up one tier per window & $-0.48$ & $-0.58$ \\
 & ladder controller with a truncation rule & $-0.05$ & $-0.01$ \\
 & tasks spread over all tiers & $-1.19$ & $-1.26$ \\
\midrule
Detector refresh & frontier tier of each ladder & $-1.45$ & $-1.40$ \\
\bottomrule
\end{tabular}
\end{table}

\paragraph{Trigger components in the recorded logs.}
The Qwen decisions reported in Section~\ref{sec:i1} are all supported by the decline in $U_w$; the accuracy-increase condition does not fire alone. Removing the accuracy condition would therefore leave these observed decisions unchanged. This is a log-based check of the realized trigger path, not an independently trained component ablation or evidence that the accuracy condition is unnecessary in general.

\FloatBarrier

\section{Supplementary Reward and Length-Budget Audits}\label{app:reward-budget-audits}

The full-suite length-budget comparisons below use the eleven-benchmark suite. Each comparison uses the base and trained scores from its matched-budget readout. Historical reproductions and item-subset analyses retain their explicitly stated scopes. The eleven-benchmark, outcome-independent 1,000-question output audit is reported separately.

\paragraph{The gain shrinks when the budget grows.} At 4096 tokens MM-Zero v5 is $+4.18$ above the base model (57.09 against 52.91). Giving the \emph{untrained} base model 8192 tokens gains $+2.00$ (52.91 to 54.91), so roughly half of the 4096 gain is obtainable without any training. At 8192 the same weights are $+2.45$ above the base model, which retains 59\% (Table~\ref{tab:mmz}, Figure~\ref{fig:budget-correction}). Its truncation rate is 2.84\% at 4096 against 20.1\% for the base model: the trained weights almost always finish, and the base model often does not.

\begin{table}[!htbp]
\caption{Length-budget comparison on the eleven-benchmark suite: the public base model, released MM-Zero v5, and \arsi at $S_{32}$. Each row evaluates fixed weights at both limits; Gain is the 8192-token score minus the 4096-token score. Earlier continuation and curriculum results appear in Table~\ref{tab:budget-supplement}.}
\label{tab:mmz}
\centering
\papertablefont
\begin{tabular}{@{}lr r rr r@{}}
\toprule
\multirow{2}{*}{Model / training state} & \multicolumn{3}{c}{Accuracy} & \multicolumn{2}{c}{Truncation (\%)} \\
\cmidrule(lr){2-4}\cmidrule(l){5-6}
 & \textit{4096} & \textit{8192} & \textit{Gain} & \textit{4096} & \textit{8192} \\
\midrule
Base model & 52.91 & 54.91 & $+2.00$ & 20.10 & 14.42 \\
MM-Zero v5 (released weights) & 57.09 & 57.37 & $+0.28$ & 2.84 & 0.87 \\
\midrule
\rowcolor{cBox}
\arsi, $S_{32}$ & 56.99 & 58.16 & $+1.17$ & 6.82 & 3.44 \\
\bottomrule
\end{tabular}

\end{table}

\paragraph{An untrained model at a long limit matches the trained one at a short limit.} On four datasets, raising only the base model's limit to 16384 without any training gives 51.12, comparable to MM-Zero after training under 4096 (51.08). This is a cross-limit comparison on a subset, so it supports the point about budgets but not a precise ranking.

\paragraph{No training contribution beyond the budget was detected in a paired test.} We reproduced the published scores (base 50.27 against the paper's 50.03, v5 54.25 against 54.18) and the training procedure locally (three rounds: v1 52.25, v2 53.68, v3 53.07). On 7,257 fixed items with an independent selector, only raising the base model's limit to 8192 gives $+1.92$ ($p<0.0001$), while the two earlier MM-Zero versions we could test gave $+0.69$ ($p=0.138$) and $+0.37$ ($p=0.434$), so no contribution of training beyond termination was detected. ``Not detected'' is not ``absent'', and these are not the v5 weights. Among base-model failures that did not finish, 4,006 tasks went from 0.30 to 29.18 accuracy after MM-Zero training. On ChartQA the gain is entirely in machine-templated questions (augmented $+6.12$, human $-0.47$).

\paragraph{Potential proxy-reward vulnerabilities.} The source audit identifies potential mismatches between proxy rewards and externally verified correctness. The training label is the solver's own majority vote, not a program's answer, and the generator's answer agrees with it only 5--12\% of the time. The self-consistency denominator drops truncated responses. No check requires that an image be necessary for the answer, and about 16\% of the released images are almost entirely black. The difficulty reward favors self-consistency near 0.5, where a majority-vote label has a small vote margin and need not be correct. This is a tension between difficulty selection and label reliability, rather than a logical contradiction. We have not shown that any of these causes the effect in Table~\ref{tab:mmz}; each is a pathway by which a gain at a short limit could reflect termination behaviour rather than visual reasoning. A longer list is in Appendix~\ref{app:mmz}.

\paragraph{Our own reward is not exempt.} Our reward includes a termination term. The first window of fresh random tasks at 4096 tokens shows its effect: the arm gains $+1.22$, almost entirely from its truncation rate falling from 17.2\% to 12.3\%, while numeric-answer items are flat across arms. Moreover \arsi retains about 80\% of its gain over the base model when the limit grows from 4096 to 8192, and truncation at 8192 is still 5.4\% (every-window random refresh), 9.6\% (ladder control), and 11.9\% (failure-prioritized sampling) at window 4 of the earlier experiments. We therefore treat any 4096-token result as suspect, use 8192 as the main setting, and report a stable-format view. Only the longest-trained earlier chain (fixed-pool continuation) retains its entire gain across limits.

\paragraph{What we do not claim.} MM-Zero's gain is not spurious: at 8192 the same weights are still $+2.45$ above the base model. We withdraw an earlier accusation of ours that it hurts items the base model already answers; once a mean-reversion baseline is added, the change is $+1.98$ and there is no harm. The comparison of fixed-pool continuation with MM-Zero v5 at 8192 is between two fixed sets of weights, each trained once, and fixed-pool continuation belongs to the earlier lineage in which a model snapshot writes tasks, not to \arsi. \arsi itself exceeds MM-Zero by $+0.79$ (58.16 against 57.37), which is within the estimated training noise and establishes neither superiority nor non-inferiority.

\paragraph{A practical recommendation.} Zero-data self-improvement results should report, at a minimum, the base model at the same relaxed limit, the truncation rate of every model, and the gain at two or more limits. A gain that falls sharply with the limit is partly a gain in finishing.

\FloatBarrier

\section{Supplementary Fresh-Pool Training Comparison}\label{app:fresh-pool-comparison}

\paragraph{Does the validated pipeline improve training?}
For Qwen, we compare pool replacement with continuation at the same number of updates. At the training progress corresponding to $S_{23}$, both arms share the trajectory through $S_{13}$. Random refresh then uses a fresh 400-task pool, whereas continuation retains the original 500-task pool. Table~\ref{tab:qwen-fresh-pool} evaluates both arms on all eleven benchmarks (22,042 questions) at 8192 tokens: 60.40 versus 58.35, a $+2.05$-point Macro11 difference. The completion-stratum decomposition uses these same outputs and benchmark weights.

\begin{table}[!htbp]
\caption{Fresh-pool training comparison on all eleven benchmarks, using accuracy scoring and an 8192-token limit at the same update budget. Macro11 and the difference are computed before rounding. Pool replacement also changes pool size from 500 to 400 tasks.}
\label{tab:qwen-fresh-pool}
\centering
\papertablefont
\begin{tabular}{@{}l r@{}}
\toprule
Qwen variant at matched $S_{23}$ training progress & Macro11 \\
\midrule
Original-pool continuation & 58.35 \\
Fresh random replacement (\arsi) & 60.40 \\
\midrule
Replacement minus continuation & $+2.05$ \\
\bottomrule
\end{tabular}
\end{table}

This comparison favors the implemented fresh-pool policy in the observed Qwen run. Its both-finished contribution is $-0.60$ points, with $+2.66$ from the other completion strata. Each contribution retains every item in its benchmark denominator and then averages the eleven benchmarks. The partition is descriptive because completion status can change with training. Each arm was trained once, and the gain cannot be assigned specifically to adaptive timing: the observed random-refresh decisions coincide with a three-window schedule. The current strategy changes which fixed-library instances are consumed, rather than learning new templates or new generation rules.

\paragraph{Downstream gains survive a format check.} On the same eleven-benchmark outputs, original answer matching gives a gain of $+3.33$, while format-corrected scoring preserves $+3.27$ (57.77 to 61.05). Checkpoints and responses are held fixed. ChartQA improves by $+2.12$, while truncation changes from 0.60\% to 0.56\%. This shows that fewer truncated answers do not explain every improvement; it does not determine the overall proportion attributable to reasoning. Per-benchmark checks, mechanism diagnostics, and cross-model observations are reported together in Appendices~\ref{sec:audit}, \ref{sec:why}, and~\ref{sec:models}.

\FloatBarrier

\section{Independent Training-Pool Checks and Controller Replay}\label{app:pipeline-audit-detail}

\paragraph{Independent answer checks expose remaining defects.}
Table~\ref{tab:pipeline-audit} summarizes the recorded independent recomputation results. The fully audited mv, ref, and tab/sc/ca groups have no observed disagreements; the plane-geometry result covers a sample only. The hr audit finds five cases with ties for ``third highest.'' They are retained as observed defects in the audit, so the pipeline cannot be described as guaranteeing zero label ambiguity. These checks establish agreement of reference answers on the audited records, rather than comprehensive visual or linguistic validity. Appendix~\ref{app:task-details} gives the generator definitions and admission procedure.

\begin{table}[!htbp]
\caption{Independent reference-answer checks. Denominators are the audited records for each group, not all 16,000 reservoir candidates. Agreement is distinct from proof that the question is visually necessary or unambiguous.}
\label{tab:pipeline-audit}
\centering
\papertablefont
\begin{tabular}{@{}p{.39\linewidth} r p{.27\linewidth}@{}}
\toprule
Generator group & Answer agreement & Audit scope \\
\midrule
Multi-hop geometry (mv) & 3,881 / 3,881 & All candidates \\
Referring charts (ref) & 1,568 / 1,568 & All candidates \\
Tables, counting, single-series charts & 4,580 / 4,580 & All candidates \\
Plane geometry (ga) & 150 / 150 & Sample only \\
Multi-series chart reading (hr) & 1,922 / 1,927 & Audited subset; five tie ambiguities \\
\bottomrule
\end{tabular}

\end{table}

\paragraph{Difficulty has separate construction and empirical meanings.}
The construction records quantities such as dependency depth, required reads, and reading mode. A hop count certifies program structure when the later computation uses an earlier result; it does not certify a particular model error rate. We therefore measure accuracy and truncation on held-out synthetic tasks. In the earlier multi-hop ladder, accuracy falls from 99.7\% at one hop to 21.2\% at five, but truncation reaches 75\% at five hops. This supports a model- and budget-dependent difficulty ladder, with a substantial length component. It does not establish that additional hops produce better transferable training data. The historical twelve-benchmark comparison of multi-hop and parallel-reading tasks gives $-0.37$ points with interval $[-1.50,+0.74]$, providing no detected transfer advantage (Appendix~\ref{sec:history}).

\paragraph{Checks on actually consumed replacement pools.}
The two Qwen random replacement pools contain 800 tasks. Their actual training parquet IDs, gold/solution values, and image hashes agree with their selected reservoir records, and the parquet hashes match the training receipts. There are no repeated content keys within or between the two pools and no overlap with diagnostic holdout IDs. Independent solvers for mv, ref, ca, tab, and sc agree on 508/508 tasks; 199 ga and 93 hr tasks are not independently solved in this check. These are checks of textual-state answers and data consistency, not a complete audit of rendered visual or linguistic validity.

\paragraph{The refresh controller has checkable invariants.}
Equation~\eqref{eq:trigger} requires enough accepted windows on the same pool and compares only that pool's logged signals. Hence a normal refresh cannot fire before $n_{\min}$ windows, and its lagged comparison cannot cross a pool boundary because $n_{\min}>\ell$. Upon a successful refresh, the replacement draw uses unused task identities under fixed family quotas; the next window inherits the last accepted weights. These are conditional consequences of the specified rule and its pool bookkeeping. In the recorded Qwen run, refreshes occur at $S_{13}$ and $S_{23}$, and the decline in $U_w$ supports both. The accuracy condition never fires alone. This verifies the realized decision path, without proving that the detector finds a benchmark plateau or the optimal refresh time. The detector reads no benchmark scores at runtime; the fixed family mix and vocabulary were nevertheless informed by evaluation-set metadata, as disclosed in Appendix~\ref{app:task-eligibility}.

\paragraph{Reward-component replay of the controller.}\label{app:controller-replay}
We recompute 38 controller receipts from 28 distinct training logs. Shared windows reuse their corresponding log. Reconstructed advantages and signal counts agree with the recorded values. On the main random-refresh arm, prompts with nonzero total-reward advantages decline from 196 to 163 between the first and third windows of the first replacement pool, triggering the retention rule. Counting only prompts with within-group correctness variation gives 166 to 155, a ratio of 0.934; the accuracy increase is 0.0946875. Neither reaches the corresponding trigger threshold. Holding the observed rollouts fixed, the second refresh therefore depends on auxiliary-reward variation being included in the signal. This is a deterministic replay of those observations, not a prediction of a policy retrained with only correctness feedback.

\paragraph{{Decisions across pool stages.}}
{
Table~\ref{tab:adaptive-pool-decisions} records the three pool stages for each model. The comparison uses the first and third accepted windows on the same pool, matching the controller's two-window lag. In the first two stages, the listed refreshes were executed. The third-stage rows replay the fixed rule after $S_{33}$ to obtain the next-window decision. Removing the accuracy condition leaves the recorded Qwen decisions unchanged. MiMo's first two refreshes satisfy both conditions, while neither condition holds in its final-stage replay. The full final-pool signal sequences are 181, 168, and 152 for Qwen and 152, 154, and 150 for MiMo.
\par}

\begin{table}[!htbp]
\caption{{Training feedback and controller actions across all three pool stages. Retention is $U_{\mathrm{last}}/U_{\mathrm{first}}$, and $\Delta a$ is in percentage points. Refresh requires retention at most 85\% or an accuracy increase of at least ten points, after the minimum observation period.}}
\label{tab:adaptive-pool-decisions}\label{tab:cross-model-refresh-replay}
\centering
\begingroup

\papertablefont
\begin{tabularx}{\linewidth}{@{}lXrrrrll@{}}
\toprule
Model & Pool & $U_{\mathrm{first}}$ & $U_{\mathrm{last}}$ & Retention & $\Delta a$ & Action & Record \\
\midrule
Qwen & Initial & 239 & 195 & 81.59\% & $+7.00$ & Refresh & Executed \\
 & Replacement 1 & 196 & 163 & 83.16\% & $+9.47$ & Refresh & Executed \\
 & Replacement 2 & 181 & 152 & 83.98\% & $+6.59$ & Refresh & Replay \\
\midrule
MiMo & Initial & 245 & 153 & 62.45\% & $+14.28$ & Refresh & Executed \\
 & Replacement 1 & 199 & 143 & 71.86\% & $+14.50$ & Refresh & Executed \\
 & Replacement 2 & 152 & 150 & 98.68\% & $+6.63$ & Keep & Replay \\
\bottomrule
\end{tabularx}
\endgroup
\end{table}

\FloatBarrier

\section{Full-Suite Output Decomposition}\label{app:full-output-decomposition}

\paragraph{MiMo shows net correction beyond truncation recovery.}
We compare MiMo at $S_0$ and $S_{32}$ on all 22,042 questions from the eleven main-result benchmarks, using accuracy scoring and the same 8192-token limit. For each benchmark, we count wrong-to-right and right-to-wrong transitions among questions whose responses finish normally at both checkpoints. Their difference, divided by the full size of that benchmark, is the contribution of the both-finished stratum; averaging these contributions gives its Macro11 contribution. We retain both correction directions and do not filter questions by their outcome.

\begin{table}[!htbp]
\caption{MiMo's full eleven-benchmark output comparison, $S_0\to S_{32}$. Gains and contributions are percentage points; transition counts are totals across benchmarks. The contribution averages per-benchmark net transitions divided by each benchmark's full test size. Both decodings use the same trained weights, so they are decoding repeats rather than independent training runs.}
\label{tab:mimo-nontruncation}
\centering
\papertablefont
\begin{tabular}{@{}l r r r r@{}}
\toprule
Decoding & \shortstack{Total gain\\(Macro11)} & \shortstack{Finished\\wrong$\to$right} & \shortstack{Finished\\right$\to$wrong} & \shortstack{Both-finished\\contribution} \\
\midrule
Primary & $+1.63$ & 1,460 & 1,235 & $+0.99$ \\
Repeat & $+2.21$ & 1,483 & 1,169 & $+1.69$ \\
\bottomrule
\end{tabular}

\end{table}

MiMo improves from 59.08 to 60.71 Macro11 on the primary decoding. Table~\ref{tab:mimo-nontruncation} shows that net correction occurs even when both responses finish, and its direction repeats under another decoding. This supports a model-specific observation beyond truncation recovery. Completion status is affected by training, however, so the stratum contribution is descriptive rather than a causal fraction of reasoning improvement. The corresponding full-suite decomposition for both models, including Qwen, is retained in Appendix~\ref{app:full-output-decomposition}.

We retain both backbones' full eleven-benchmark decompositions to make the scope of Table~\ref{tab:mimo-nontruncation} explicit. Every comparison uses the complete 22,042-question suite, an 8192-token limit, and format-corrected scoring. For each benchmark, we partition questions by whether the responses finish normally at both checkpoints; all other completion strata form the complementary group. Contributions use each benchmark's full test size as the denominator and are then averaged across eleven benchmarks.

\begin{table}[!htbp]
\caption{Full-suite Macro11 accounting through $S_{32}$, in percentage points. The Qwen $S_0$ comparison includes initialization; the $\St$ comparison starts after initialization. Contributions add before rounding. Completion strata are post-training outcomes, so this decomposition does not identify a causal fraction of reasoning improvement.}
\label{tab:full-output-decomposition}
\centering
\paperdensefont
\begin{tabular}{@{}l l l r r r@{}}
\toprule
Model & Start & Decoding & \shortstack{Total\\gain} & \shortstack{Both-finished\\contribution} & \shortstack{Other strata\\contribution} \\
\midrule
Qwen & $S_0$ & Primary & $+4.40$ & $-0.75$ & $+5.15$ \\
Qwen & $S_0$ & Repeat & $+4.85$ & $-0.31$ & $+5.16$ \\
Qwen & $\St$ & Primary & $+3.27$ & $-0.71$ & $+3.98$ \\
MiMo & $S_0$ & Primary & $+1.63$ & $+0.99$ & $+0.64$ \\
MiMo & $S_0$ & Repeat & $+2.21$ & $+1.69$ & $+0.52$ \\
\bottomrule
\end{tabular}

\end{table}

Qwen's aggregate both-finished contribution is negative in both decodings, while MiMo's is positive. We therefore use MiMo for the full-suite non-truncation observation and Qwen for the separate fresh-pool comparison at the same update budget, without extending either conclusion to all backbones. Individual non-truncated corrections also occur for Qwen: on complete ChartQA, the primary decoding has 127 wrong-to-right versus 51 right-to-wrong transitions among both-finished responses, contributing $+3.04$ points over all 2,500 questions; the repeat has 131 versus 56, contributing $+3.00$ points. These local observations do not override the full-suite result.

\FloatBarrier

\section{Generated-Task and Reward-Check Protocols}\label{app:generated-check-protocol}

\begin{table}[!htbp]
\centering\begin{minipage}{.72\linewidth}
\caption{Independent checks of the 1,000 generated tasks in $\mathcal{A}$. Membership is fixed before checking, without replacement of failed cases. Solvers use the recorded textual state, and image hashes check file identity.}
\label{tab:generated-audit-1000}
\centering
\papertablefont
\begin{tabularx}{\linewidth}{@{}Xrr@{}}
\toprule
Generator group & Tasks & \shortstack{Independent answer\\agreement} \\
\midrule
Multi-hop geometry (mv) & 200 & 200 / 200 \\
Referring charts (ref) & 200 & 200 / 200 \\
Single-series charts (ca) & 200 & 200 / 200 \\
Tables (tab) & 200 & 200 / 200 \\
Counting (sc) & 200 & 200 / 200 \\
\midrule
Total & 1,000 & 1,000 / 1,000 \\
\bottomrule
\end{tabularx}

\end{minipage}
\end{table}

\paragraph{Generated tasks.}
Collection $\mathcal{A}$ contains 200 tasks each from mv, ref, ca, tab, and sc, taken from actually consumed replacement pools in the Qwen comparison arms. Within each supported generator, tasks are sampled uniformly without replacement before independent answer recomputation, using one fixed random-number generator without seed search. The membership list is saved before the checks, and a failed check is not replaced by another task. All 1,000 independently recomputed answers agree; source labels and image hashes also agree on all 1,000. ga and hr are outside this finite audit, and textual-state agreement is not a complete rendered-image audit.

\paragraph{Reward stress tests.}
For each task in $\mathcal{A}$ we construct seven responses: correct committed, wrong committed, wrong boxed but truncated, correct boxed but truncated, bare correct, wrong unboxed and truncated, and wrong followed by a final correct box. We invoke the production reward functions on every variant, holding the task reference fixed. Every correct committed answer receives unit correctness reward, every wrong committed answer receives zero, and the former outranks the latter on all 1,000 tasks. With correctness and termination fixed, formatting alone adds 0.1; with correctness and formatting fixed, normal commitment alone adds 0.3. Each isolated auxiliary increase is observed on all 1,000 references. These are constructed scoring probes, not sampled model behavior or a reward-removal training experiment. The fixed coefficients imply a minimum correctness-ranking margin of $1-0.1-0.3=0.6$, as in Eq.~\eqref{eq:reward-dominance}.

\FloatBarrier

\section{Adversarial Checks and Machine-Checked Specification Properties}\label{app:mechanism-validation}

\paragraph{Fixed references and adversarial responses.}
We retain the generated-task audit's 1,000 references without replacing any failing task. The 18 variants include canonical correct and incorrect answers, both orders of conflicting boxes, a quoted answer followed by a different final answer, gold-prefixed disjunctions, unrelated suffixes, negated answers, empty outputs, and different finish reasons. Expected binary correctness is specified for 16 variants per task; the remaining two diagnose unboxed and truncated correct answers. The production checker passes 15,780/16,000 specified checks. The 220 failures consist of two variants on each of 110 choice tasks. The candidate passes 16,000/16,000 by requiring a complete terminal option in its choice branch; other answer kinds retain their production checks. This finite adversarial set is not exhaustive.

Re-scoring five accepted replacement-pool windows reproduces every logged choice correctness value on 3,760 choice responses out of 16,000 responses. The candidate changes one score: a response had an earlier answer box followed by unfinished reasoning. This is a difference in terminal-answer enforcement, not evidence that the model intentionally exploited the checker. A prospective matched training fork retains or removes the termination coefficient from a common accepted checkpoint, using the same pool, update count, and a fixed outcome-independent 1,000-question evaluation frame. It evaluates both arms at 8192 and 16384 tokens. Its training results are pending and are not included in the reported outcomes. Because the termination weight also affects dynamic group survival, that fork measures the combined reward-and-sampling intervention.

\paragraph{Specification properties.}
Z3 checks the negation of each of the following properties and returns unsatisfiable:
\begin{enumerate}[leftmargin=*,nosep]
\item Given $C\in\{0,1\}$, $F\in[0,1]$, and $T\in[-1,0]$, unit correctness outranks zero correctness by at least $0.6$ with the fixed coefficients.
\item If the current pool has $j\geq n_{\min}>\ell$ accepted windows, the lagged comparison remains within that pool.
\item For a valid difficulty pointer and probe rate in $[0,1]$, an update stays within the ladder and moves by at most one step.
\item If each successive pool is drawn outside the accumulated history and holdout, successive pools are disjoint from that history, the holdout, and each other.
\end{enumerate}
The saved SMT queries reproduce these results on reloading. These proofs depend on the stated data types, bounds, and admission construction; renderer correctness and all Python execution paths are outside the model.

\paragraph{Implementation conformance and negative controls.}
The production detector matches the specified rule in 321,605 threshold and initialization cases, including every pair of integer signal counts in $[0,400]$ at two accuracy boundaries. The difficulty controller passes 11,935 valid pointer/probe-count combinations, and all 38 accepted-window receipts reproduce. These are finite implementation tests, separate from the SMT properties. The stronger statement that firing implies at least 80\% accuracy is satisfiable in negation: a model with zero accuracy and zero signal fires. A signal collapse accompanied by declining accuracy also fires. The detector therefore identifies the specified refresh condition, not mastery.

\paragraph{Empirical difficulty calibration on the same tasks.}
The fixed collection contains 600 tasks on four ordered axes: geometry F (92), geometry S (108), referring charts (200), and counting (200). The other 400 chart/table tasks do not define an ordered axis in this check. Using the frozen eight-response probes at a 4096-token limit, task-level Spearman correlations between rung and correctness are $-0.591$, $-0.515$, $-0.124$, and $-0.239$, respectively. An exploratory permutation check with 9,999 draws and correction across four axes supports three trends; referring charts do not pass that check (adjusted $p=0.3364$). Geometry's high-rung error rates coincide with substantial truncation: on S5, accuracy is 1.39\% and truncation is 88.89\%. Thus the construction provides empirical difficulty variation in this collection, with a length component; these diagnostics do not establish a causal effect of hops or a training advantage from difficulty-aware selection.

\paragraph{Task-precondition and representation checks.}
We additionally check the same fixed 1,000 tasks for applicable construction constraints: agreement between encoded configurations and independent text parsers; exact table headers including units; nonnegative counts and stated differences; unique in-bounds referents; positive geometry targets; and unique choice labels and option values. All 1,000 pass the applicable checks. Coherent ambiguity injections are separate from simple file corruption: the checks reject 110 duplicated correct options, 106 tied table extrema or ranks, and 105 ambiguous base referents, as well as 1,000 configuration/target mismatches. These 1,321 cases test the independent audit, not every production admission path.

\paragraph{Checks against actual image content.}
Replaying the frozen generators on the selected keys reproduces the configuration, question and answer for all 200 geometry and 200 single-series chart tasks. With an explicit Fontconfig profile for the applicable Cairo renderers, the original PNGs are reproduced byte-for-byte on 200 geometry, 200 table and 200 counting tasks; all 200 chart images also reproduce in their recorded rendering branch. No original image is replaced. Independent SVG inspection verifies table values and layout, and counting-category counts, canvas bounds and absence of filled-shape overlap. Fixed OCR on the original table PNGs matches every row name and all numeric columns on 192/200 tasks; eight numeric OCR disagreements are retained with image crops, rather than counted as completed human annotations.

For the remaining 200 referring-chart tasks, source render receipts and configuration records match; independent coordinate read-back agrees with the configured values. All 200 original plot regions match the reconstructed geometry, and OCR of the original category labels and numerical axis ticks agrees on all 200. Their whole-image reproduction remains 0/200, with generic-font rendering differences unresolved. Thus the audit provides 800 exact whole-image reproductions plus 200 plot-region and key-text checks, not 1,000 exact whole-image reproductions. The recorded font profile reproduces the applicable groups but is not claimed to uniquely recover the historical machine configuration. These finite checks strengthen data binding without certifying every rendered property.

\paragraph{Quality checks on the fixed 1,000 tasks.}
All tasks pass independent reference recomputation, source consistency, image-file hashes, PNG decoding, minimum dimensions, nonblank-image checks, and diagnostic-holdout ID exclusion. Four defects are injected per task: a changed reference, modified image bytes, a changed source question, and a truncated PNG. The corresponding independent audit checks detect all 4,000 injected defects. This establishes detection of those specified defects on these tasks; it does not prove that all original admission paths enforce these checks or that every rendered question is semantically unambiguous.

\FloatBarrier

\section{Outcome-Independent 1,000-Question Output Audit}\label{app:random1000}

The random cohort below samples the full test frames to describe both improvements and regressions without conditioning question selection on correctness changes. It serves a quantitative purpose distinct from the qualitative correction examples in Section~\ref{sec:analysis}.

\paragraph{A fixed 1,000-question output audit.}
We sample without replacement within each of the eleven main-result benchmarks, using 91 questions from each of the first ten and 90 from SCEMQA-MC. Question selection does not use model outcomes. The same questions compare the public $S_0$ with $S_{32}$ for Qwen and MiMo, at an 8192-token limit with accuracy scoring. These are automated output statistics, not completed human annotations. The nearly equal benchmark quotas make this a benchmark-balanced sample; its pooled accuracy is not the micro accuracy of the full test suite.

Let $y_i^0,y_i^1$ be the two checkpoints' correctness indicators, and $f_i^0,f_i^1$ indicate normal completion. Partition the sample into four strata $\mathcal{I}_{ab}=\{i:f_i^0=a,f_i^1=b\}$. The score change has the exact accounting decomposition
\begin{equation}
 \Delta_{\mathrm{sample}}=\sum_{a,b\in\{0,1\}}D_{ab},\qquad
 D_{ab}=\frac{100}{1000}\sum_{i\in\mathcal{I}_{ab}}(y_i^1-y_i^0).
 \label{eq:sample-decomposition}
\end{equation}
Within $\mathcal{I}_{11}$, we count wrong-to-right and right-to-wrong transitions symmetrically. Their difference is the net correction count among responses that finish at both checkpoints. Each $D_{ab}$ uses all 1,000 questions as its denominator, so the contributions add to the sample gain.

\begin{table}[!htbp]
\caption{Automated analysis of the fixed random 1,000-question sample, $S_0\to S_{32}$, at 8192 tokens. Score changes are percentage points over all 1,000 questions. The last two columns include only questions for which both responses finish; Other strata sums $D_{00}+D_{01}+D_{10}$. No result-conditioned filtering is applied.}
\label{tab:sample1000}
\centering
\papertablefont
\begin{tabular}{@{}l r r r r r r@{}}
\toprule
Model & $\Delta_{\mathrm{sample}}$ & $D_{11}$ & \shortstack{Other\\strata} & $|\mathcal{I}_{11}|$ & \shortstack{Wrong\\$\to$right} & \shortstack{Right\\$\to$wrong} \\
\midrule
Qwen & $+4.70$ & $-1.20$ & $+5.90$ & 834 & 52 & 64 \\
MiMo & $+1.30$ & $+1.30$ & $0.00$ & 901 & 70 & 57 \\
\bottomrule
\end{tabular}

\end{table}

Table~\ref{tab:sample1000} distinguishes the two models: Qwen's net gain is concentrated outside the both-finished stratum, whereas MiMo has a positive net correction count within it. A second decoding of the same fixed checkpoints gives total gains of $+3.70$ and $+1.00$, with $D_{11}=-0.80$ and $+0.20$, respectively. These are decoding repeats, not independent training runs. Completion status is itself affected by training, so this partition is descriptive and does not identify a causal contribution of reasoning or of the stopping reward. In particular, the sample does not establish that Qwen's aggregate improvement comes from non-termination reasoning gains.

\paragraph{Formatting and correctness must be checked separately.}
On the same primary outputs, original answer matching gives sample gains of $+5.10$ for Qwen and $+13.90$ for MiMo, compared with $+4.70$ and $+1.30$ after format correction. The large MiMo discrepancy shows why changes in answer extraction and presentation must be separated from the more permissive correctness measure. It does not by itself demonstrate deliberate reward hacking. The sample diagnoses output changes; attributing them to a particular reward component requires a matched training ablation.

\FloatBarrier

\section{Audit: Is Our Own Gain Real?}\label{sec:audit}

This section examines \arsi from $\St$ to $S_{32}$ on the same eleven-benchmark suite and outputs as the primary results. Per-benchmark checks, visual-type subsets, and same-output rescoring use this common evaluation scope.

\paragraph{Every benchmark and every visual type improves.} Table~\ref{tab:twelve} reports all eleven benchmarks at $\St$ and $S_{32}$ with an 8192-token limit. All eleven improve, from $+1.19$ (SCEMQA-MC) to $+8.33$ (PhyX-MC). The retained visual-type subsets also improve: geometry from 56.19 to 59.83, chart from 79.63 to 81.66, table from 70.96 to 72.99, counting from 54.81 to 61.74. The type definitions and denominators are given together in Table~\ref{tab:vtype}.

\begin{table}[!htbp]
\caption{Per-benchmark accuracy on the eleven-benchmark suite at 8192 tokens, $\St$ against \arsi at $S_{32}$. All eleven benchmark point estimates improve.}
\label{tab:twelve}
\centering
\papertablefont
\begin{tabular}{@{}lrrr@{}}
\toprule
Benchmark & $\St$ & \arsi & Change \\
\midrule
ChartQA & 82.84 & 84.96 & $+2.12$ \\
MMMU & 62.23 & 64.36 & $+2.12$ \\
MMSI & 26.40 & 29.30 & $+2.90$ \\
MathVerse & 62.26 & 63.48 & $+1.22$ \\
MathVision & 47.04 & 51.45 & $+4.41$ \\
VisNumBench & 52.90 & 55.99 & $+3.08$ \\
MMMU-Pro (4 options) & 58.98 & 63.19 & $+4.21$ \\
MMMU-Pro (10 options) & 51.76 & 53.96 & $+2.20$ \\
MMMU-Pro (vision) & 49.54 & 53.76 & $+4.22$ \\
PhyX-MC & 65.83 & 74.17 & $+8.33$ \\
SCEMQA-MC & 75.71 & 76.90 & $+1.19$ \\
\midrule
\rowcolor{cBox}
Macro11 & 57.77 & \textbf{61.05} & $+3.27$ \\
\bottomrule
\end{tabular}
\end{table}

\paragraph{The gain persists after answer-format correction.} On the same eleven-benchmark outputs, original answer matching scores 57.24 at $\St$ and 60.57 at $S_{32}$, a gain of $+3.33$. Format-corrected scoring gives 57.77 and 61.05, preserving a gain of $+3.27$. The rescoring accepts unboxed final answers and normalized option values and applies the specified ChartQA numeric tolerance. Checkpoints and responses are held fixed; rescoring does not regenerate answers or retrain models.

\paragraph{The gain appears where answers rarely truncate.} ChartQA truncation changes from 0.60\% at $\St$ to 0.56\% at $S_{32}$, while accuracy rises from 82.84 to 84.96. This shows improvement on a benchmark with little truncation, but does not quantify how much of the aggregate gain comes from improved reasoning.

\paragraph{Scope of the audit.} The audit uses 8192 tokens; the eight-window run was not evaluated at 16k. Truncation at $S_{32}$ is 4.16\% pooled over the eleven benchmarks. The separate 4096-to-16k retention results in Section~\ref{sec:history} retain the historical twelve-benchmark protocol and regenerate only the originally truncated answers of earlier chains.
\section{Why Random Wins: Mechanisms}\label{sec:why}

We offer three mechanisms. They are supported by logs but not isolated by ablations.
\begin{enumerate}[leftmargin=*]
  \item \textbf{A random pool keeps the tasks the model cannot finish.} Choosing tasks moves the pool away from the full difficulty mix and removes tasks the model cannot finish within the limit. Those are exactly the tasks on which the termination term gives its strongest signal (in a random pool 21.9\% of training answers were truncated, and the correlation between length and advantage was $-0.31$; in selected pools truncation was 17\%). The consequence is visible at evaluation: on the eleven-benchmark suite at 8192 tokens, truncation at window 8 is 4.16\% for \arsi and 8.14\% for \armFront; at window 4 of the earlier selection experiment it is 6.56\% for every-window random refresh, 12.70\% for learnable-band sampling, and 14.30\% for failure-prioritized sampling.
  \item \textbf{Selection changes the answer types.} The pool built by failure-prioritized sampling has no multiple-choice tasks, whereas every-window random refresh has 222 and learnable-band sampling has 227, and failure-prioritized sampling scores 72.81\% on the 114 retained yes/no items (WATCH under the recomputed sentinel). This mixes format effects into the comparison.
  \item \textbf{Random pools do not saturate within 4--8 windows.} The number of tasks with training signal per window is 209, 180, 189, and 183 for every-window random refresh, flat after window 2, so the data do not support the idea that longer training would let a controller overtake random.
\end{enumerate}
Part of the ``difficulty'' of the frontier is also not difficulty. In the first 6,000 probed candidates, 39\% of geometry answers were truncated at 4096, so much of what looks hard in multi-hop geometry is the model running out of length. Some difficulty is created by format: the model often writes the option's numeric value in the answer box instead of the letter, which the reward scores 0. In the first 12,000 candidates, a lenient grader found 317 tasks (2.6\%) that a strict grader marks wrong at least three times more often, 311 of them multiple-choice multi-hop geometry.

\section{Evidence from Earlier Stages}\label{sec:history}

The frozen design follows three earlier versions. The first (v1) had a frozen model snapshot write the next tasks, a program compute answers, a vision-language model operate difficulty tools, and a controller move among tiers. The second (v2) used only programmatic tasks and refreshed at every window with the tier controller. In the third (this report), tiers are used for diagnosis and a log-based detector decides when to refresh. Table~\ref{tab:evol} summarizes the changes. The chains and extension below use the eleven-benchmark accuracy scores at 4096 tokens, one run per arm. Held-out synthetic tasks are reported separately; the length-budget audits report their separately identified evaluation scope.

\begin{table}[!htbp]
\caption{Evolution of the design.}
\label{tab:evol}
\centering
\papertablefont
\begin{tabular}{@{}>{\raggedright\arraybackslash}p{2.5cm} >{\raggedright\arraybackslash}p{4.15cm} >{\raggedright\arraybackslash}p{4.15cm} >{\columncolor{cBox}[\tabcolsep][0pt]\raggedright\arraybackslash}p{4.15cm}@{}}
\toprule
 & v1 & v2 & v3 (this report) \\
\midrule
Who writes tasks & a snapshot of the model; program computes answers & programs only & programs only \\
Pool & 300 rows, 225 shared, 75 new slots & 400 fresh tasks per window & 400 tasks, replaced when the detector fires \\
Replacement & controller picks tiers & frontier tiers & \textbf{random} (fixed family shares) \\
When & every 2 windows & every window & \textbf{log-based detector} \\
Difficulty tiers & state space and controller & control and diagnosis & \textbf{diagnosis and reporting only} \\
Health gate & $\geq 99/100$ & $\geq 99/100$ & $\geq 97/100$ \\
Replaced because & partial replacement lost to no replacement; model-based selection lost to random & detector with random used a quarter of the data; frontier lost to random a third time & final \\
\bottomrule
\end{tabular}

\end{table}

\paragraph{The self-writing loop ran.} A model snapshot acting as task writer produced no truncated outputs at the writing stage, 100\% of its specifications were valid, and its rewriting success matched the base model (84\% against 85.5\%). The first closed-loop prototype also exercised model-operated difficulty tools. Two defects were found afterwards: an unquoted command argument broke depth feedback, and a sign error inflated one tier's difficulty. Both were fixed and the controller's decisions stayed the same. The current benchmark comparisons use the verified checkpoints in Tables~\ref{tab:chains} and~\ref{tab:ext}.

\paragraph{The model is blind to its own state.} Models' predictions of which questions they would get wrong give AUC about 0.50 for four models, so state must be measured by running the model. This is why the detector uses the training log rather than a self-assessment.

\paragraph{Escalating difficulty hurts held-out synthetic performance.} In six-window chains from $\St$ with pools replaced every two windows, keeping the initial state is best on held-out synthetic tasks; re-selecting by state and climbing to H3 is worst. The fixed-writer arm also outperforms the state-following writer on these held-out tasks. On Macro11, saturation-based escalation is best at 56.69, while the six benchmark point estimates span 55.88--56.69. These single-run benchmark results do not establish a benefit from a stronger writer or isolate the effect of escalation. Table~\ref{tab:chains} reports both evaluation scopes.

\begin{table}[!htbp]
\caption{Six-window chains from $\St$. Benchmark scores use the eleven-benchmark protocol at 4096 tokens; held-out synthetic-task differences retain their separate original evaluation. The saturation-based arm has the highest benchmark point estimate.}
\label{tab:chains}
\centering
\papertablefont
\begin{tabular}{@{}l lrr@{}}
\toprule
Escalation rule & Late-stage tiers & \shortstack{Held-out\\vs.\ uniform} & \shortstack{Macro11\\4096} \\
\midrule
no escalation & F2 / H1 & $+1.27$ & 56.05 \\
up only on saturation & H1, H2 & $+0.06$ & \textbf{56.69} \\
uniform & all ten tiers & $+0.00$ & 56.47 \\
half by state, half uniform & H2 & $-2.12$ & 56.02 \\
writer fixed at $\St$ & up to H2 & $-0.99$ & 55.90 \\
re-select by state each time & up to H3 & $-2.42$ & 55.88 \\
\bottomrule
\end{tabular}
\end{table}

\paragraph{Most ``curriculum gains'' were termination effects.} We re-generated every answer truncated at 4096 with a 16k limit (Table~\ref{tab:gap}). The gain over $\St$ was unchanged (7-chain mean $+1.18\to+1.17$, retention 0.99, 95\% interval 0.44--1.43, flagged as unstable), whereas differences between selection rules mostly vanished. This is why our main comparisons use 8192 tokens.

\begin{table}[!htbp]
\caption{Re-generating the earlier chains' 4096-truncated answers at a 16k limit, under the original twelve-benchmark evaluation protocol. These are historical trajectories, not the eight-window \arsi run.}
\label{tab:gap}
\centering
\papertablefont
\begin{tabular}{@{}>{\raggedright\arraybackslash}p{5.5cm}r rl@{}}
\toprule
\multirow{2}{*}{Comparison} & \multicolumn{2}{c}{Gain} & \multirow{2}{*}{Reading} \\
\cmidrule(lr){2-3}
 & \textit{4096} & \textit{16k} & \\
\midrule
Chains $-$ $\St$ (mean of 7) & $+1.18$ & $+1.17$ & real capability \\
Uniform $-$ state-selected & $+0.74$ & $+0.08$ & only faster termination \\
Tool-assisted $-$ plain & $+1.26$ & $-0.13$ & only faster termination \\
Promotion on saturation $-$ no escalation & $+0.83$ & $+0.68$ & mostly kept, below noise \\
\bottomrule
\end{tabular}

\end{table}

\paragraph{Where the benchmark errors are.} At $\St$, error can arise from running out of output length, from tasks outside the model's current competence, or from answer extraction. The eleven-benchmark per-set audit reports these observable accuracy and truncation quantities directly. A useful selection rule must distinguish such causes rather than treating every failure as a learnable example.

\paragraph{Historical multi-hop audit.} The following diagnostic retains its original twelve-benchmark evaluation scope.  On one multi-hop ladder, accuracy over one to five hops is .997, .881, .640, .388, and .212. Given the same information as text, accuracy stays at .84--1.0, and truncation at three to five hops is .30, .56, and .75. For an 8B model, hard multi-hop tasks are tasks where reading the figure is uncertain, the model keeps deliberating, and the answer runs out of length. Training to complete long chains (hops 3--4 against 1--2) improved held-out in-domain tasks by $+6.58$ but did not transfer: MathVision $-0.30$, truncation $+1.97$, benchmark $-1.0$. The second-window comparison remains negative on the eleven-benchmark suite at 4096 tokens: $-1.07$ in primary accuracy and $-2.37$ under \lj. A further check found 54 training rows with the same construction as a benchmark item, bounding the effect on the average at 0.02 points.

\paragraph{Difficulty is learnable.} In the ladder-controller arm, the pass rate on held-out tasks at the pointer rises quickly: on one multi-hop ladder tier 2 goes from .58 to .93 in one window, and on the other from .52 to .87 in two. Tier 3 of the first ladder goes from .32 to .72, .76, and .81 over three windows with truncation falling from .21 to .13, and tier 4 reaches .59 by window 3. The other four ladders move little. The controller moved on only two ladders, so what was tested is mainly a frontier-weighted mix against random, not frequent promotion and demotion.

\paragraph{Fresh tasks help after the plateau, and partial replacement trails fixed-pool continuation.} On the eleven-benchmark format-corrected scoring readout at 8192 tokens, every-window random refresh at window 4 is $+1.82$ above original-pool continuation and $+2.52$ above $\St$ (stable-format difference versus continuation: $+1.66$; \lj $+1.87$). The policies also differ in pool size and exposure: continuation reuses 500 rows about 3.2 times, while random refresh sees 1,600 distinct tasks. In the earlier extension at 4096 tokens, replacing 225 of 300 rows every two windows trails the fixed pool at every shared window from 7 to 11. Partial refresh leads only on a held-out chart set, so that in-domain observation does not establish broader transfer. The full extension results are in Table~\ref{tab:ext}.

\begin{table}[!htbp]
\caption{Extension run under the eleven-benchmark protocol at 4096 tokens, from the window-6 checkpoint (56.47). Partial-pool refresh has no accepted window-12 result because its health gate stopped training (98 of 100 normal updates, twice). Fixed-pool continuation has the higher point estimate at every shared window; its scores fluctuate rather than rising monotonically.}
\label{tab:ext}
\centering
\papertablefont
\begin{tabular}{@{}lrrrrrr@{}}
\toprule
\multirow{2}{*}{Arm} & \multicolumn{6}{c}{Macro11 at window (4096 tokens)} \\
\cmidrule(l){2-7}
 & 7 & 8 & 9 & 10 & 11 & 12 \\
\midrule
Partial refresh (225 of 300) & 55.93 & 56.86 & 56.42 & 56.65 & 56.48 & -- \\
Fixed pool (300 tasks) & 57.26 & 57.59 & 58.38 & 57.76 & 58.74 & 58.61 \\
\midrule
Partial refresh $-$ fixed pool & $-1.33$ & $-0.72$ & $-1.96$ & $-1.10$ & $-2.26$ & -- \\
\bottomrule
\end{tabular}
\end{table}

\begin{table}[!htbp]
\caption{Supplementary length-budget comparisons on the eleven-benchmark suite, including earlier continuation and curriculum experiments. Earlier window indices belong to their original trajectories and do not denote the main $S_{ij}$ states. Each row evaluates fixed weights at both limits.}
\label{tab:budget-supplement}
\centering
\papertablefont
\begin{tabular}{@{}>{\raggedright\arraybackslash}p{5.5cm}r r rr r@{}}
\toprule
\multirow{2}{*}{Checkpoint} & \multicolumn{3}{c}{Accuracy} & \multicolumn{2}{c}{Truncation (\%)} \\
\cmidrule(lr){2-4}\cmidrule(l){5-6}
 & \textit{4096} & \textit{8192} & \textit{Gain} & \textit{4096} & \textit{8192} \\
\midrule
Base model & 52.91 & 54.91 & $+2.00$ & 20.10 & 14.42 \\
MM-Zero v5 (released weights) & 57.09 & 57.37 & $+0.28$ & 2.84 & 0.87 \\
\midrule
Fixed-pool continuation, window 10 & 56.23 & 58.35 & $+2.12$ & 12.60 & 6.83 \\
Fixed-pool continuation, window 11 & 57.24 & 59.16 & $+1.92$ & 10.16 & 4.55 \\
Fixed-pool continuation, window 12 & 56.86 & 58.83 & $+1.97$ & 10.27 & 5.21 \\
Every-window random refresh, window 4 & 55.51 & 57.35 & $+1.85$ & 10.55 & 5.44 \\
Ladder control, window 4 & 54.62 & 56.93 & $+2.31$ & 15.69 & 9.62 \\
\rowcolor{cBox}
\arsi, $S_{32}$ & 56.99 & 58.16 & $+1.17$ & 6.82 & 3.44 \\
\bottomrule
\end{tabular}

\end{table}

\section{{Supplementary MiMo Results}}\label{sec:models}

These MiMo-VL-7B-SFT policy comparisons use the same eleven-benchmark suite and an 8192-token output limit as the main-lineage results. We use the same frozen detector and shared random replacement pools. The \armCont and \armPeriodic comparison pools were selected at Qwen's difficulty frontier, not at MiMo's own; the \arsi pools use model-independent random sampling. One run per arm.

\paragraph{The detector replicates on MiMo.} On the old pool it fires at window 3, as on Qwen: $U$ falls 245, 194, 153 (ratio .62), with accuracy increasing from .490 to .633. On MiMo it fires again on the first new pool at window 6 ($U$ 199 to 143, ratio .72).

\paragraph{MiMo-VL: refreshed-pool point estimates.} The shared old-pool checkpoint scores 60.35 Macro11 after three windows. At window 8, \arsi reaches 60.71 versus 60.01 for original-pool continuation, a $+0.70$-point difference (Table~\ref{tab:mimo}). Every-window and periodic refresh score 60.69 and 60.53. All three refreshing arms have higher endpoint point estimates than continued reuse in this run. \arsi has the highest point estimate, with small margins of $+0.02$ over every-window refresh and $+0.18$ over periodic refresh.

\begin{table}[!htbp]
\caption{MiMo-VL-7B-SFT on the eleven-benchmark suite, Macro11 at 8192 tokens, one run per arm. The shared old-pool checkpoint at window 3 scores 60.35. Differences are computed before rounding from the available aggregate precision.}
\label{tab:mimo}
\centering
\papertablefont
\begin{tabular}{@{}lr r rr r@{}}
\toprule
\multirow{2}{*}{Arm} & \multicolumn{3}{c}{Macro11 at window} & \multicolumn{2}{c}{Window 8 minus} \\
\cmidrule(lr){2-4}\cmidrule(l){5-6}
 & \textit{4} & \textit{6} & \textit{8} & \textit{$\St$} & \textit{\armNR} \\
\midrule
\armNR & 60.97 & 60.19 & 60.01 & $-0.34$ & $+0.00$ \\
\armPeriodic & 60.87 & 60.44 & 60.53 & $+0.18$ & $+0.52$ \\
\armCont & 61.04 & 61.07 & 60.69 & $+0.34$ & $+0.68$ \\
\rowcolor{cBox}
\arsi & 60.40 & 61.04 & 60.71 & $+0.36$ & $+0.70$ \\
\bottomrule
\end{tabular}
\end{table}

\paragraph{What this changes about our claims.} The detector transfers to MiMo, and random fresh-pool replacement has a positive $+0.70$-point endpoint difference over continued reuse in the observed run. Its small endpoint margins over every-window and periodic refresh, together with one training run per policy, do not establish an advantage of detector-triggered timing.

\FloatBarrier
\Needspace{.92\textheight}
\section{What the Evidence Does and Does Not Support}\label{sec:claims}

\begin{table}[H]
\caption{Claims from the current eleven-benchmark policy comparisons and explicitly scoped historical auxiliary experiments. Each numerical comparison uses its stated protocol.}
\label{tab:claims}
\centering
\papertablefont
\begin{tabular}{>{\raggedright\arraybackslash}p{7.2cm} >{\raggedright\arraybackslash}p{7.2cm}}
\toprule
{\color{cUp}\ding{51}}\ Supported & {\color{cDn}\ding{55}}\ Not supported \\
\midrule
A training log alone, with no test set, can tell that a pool is learned out. It lags the benchmark by about one window, fires on the old pool and on two replacement pools, and the half-life of a 400-task pool is three windows. &
That the detector finds the best refresh time. \\\addlinespace[3pt]
After the detector fires, random fresh tasks reach 61.05 with 800 new tasks, against 60.64 for replacing every window with 3,200 (4.09 against 0.90 points per thousand tasks). &
That the $+0.40$ point estimate establishes a detector-timing advantage. \\\addlinespace[3pt]
Choosing the replacement by difficulty frontier or by the model's rollouts is not better than random (three experiments: $-2.63$, $-0.82$, $-1.45$). &
That difficulty control or state-based selection improves the benchmark. \\\addlinespace[3pt]
The old pool is learned out and the plateau is real: \armNR scores 58.48 and 58.35 before running out of signal; at window 6 every refreshing arm is 1.39 to 2.05 above \armNR. &
That the pre-registered primary endpoint (\armFront $-$ \armNR) holds: \armNR halted, so there is no reading. \\\addlinespace[3pt]
On Macro11, every-window random refresh $-$ original-pool continuation is $+1.82$ (stable $+1.66$, \lj $+1.87$). &
Any ``significant'' improvement; every arm was trained once. \\\addlinespace[3pt]
Difficulty is quantifiable as monotone tiers and learnable in domain (tier 2 of one ladder .58 to .93). &
Multi-hop training transfers specifically (MathVision $-0.30$; against a matched control $-0.37$). \\\addlinespace[3pt]
State must be measured: self-assessment AUC $\approx 0.50$. &
A stronger model writes better tasks (a fixed writer did no worse). \\\addlinespace[3pt]
The mean gain over $\St$ in seven earlier chains is retained ($+1.18\to+1.17$) when their 4096-truncated answers are regenerated at 16k; differences between selection rules largely shrink. &
The MLLM-operated difficulty tool adds real capability ($-0.13$ at 16k). \\\addlinespace[3pt]
Part of MM-Zero v5's headline gain is termination: its gain over base falls from $+4.18$ to $+2.45$ with the limit, and the untrained base gains $+2.00$. &
That MM-Zero's gain is spurious or that \arsi significantly exceeds it ($+0.79$, within noise). \\\addlinespace[3pt]
On the same eleven-benchmark outputs, format-corrected scoring preserves a gain of $+3.27$ versus $+3.33$ with original answer matching; ChartQA improves $+2.12$ with truncation below 1\%; all eleven benchmarks and all four retained visual-type subsets improve. &
Open-ended recursive self-improvement or unbounded gain (trained to window 8; fixed-pool continuation to window 12). \\\addlinespace[3pt]
On MiMo-VL the detector fires at window 3, as on Qwen, and random refresh exceeds original-pool continuation by $+0.70$ at window 8. &
That the small MiMo endpoint margins ($+0.02$ over \armCont and $+0.18$ over \armPeriodic) establish a timing advantage. \\
\bottomrule
\end{tabular}

\end{table}

\FloatBarrier
\Needspace{.55\textheight}
\section{Scores by Model State}\label{app:perwin}

\begin{table}[!htbp]
\caption{Macro11 at matched training progress on the eleven-benchmark suite; best values are bold. Two 4096-token evaluations of the same $\St$ weights score 54.55 / 54.72; they share the science-set outputs. $\St$ scores 57.77 at 8192 tokens. Shared prefixes reuse the same checkpoints.}
\label{tab:perwin}
\centering
\footnotesize
\begin{tabular}{@{}lrrrrrrrrrrrr@{}}
\toprule
\multirow{2}{*}{Arm} & \multicolumn{8}{c}{4096 tokens} & \multicolumn{4}{c}{8192 tokens} \\
\cmidrule(lr){2-9}\cmidrule(l){10-13}
 & $S_{11}$ & $S_{12}$ & $S_{13}$ & $S_{21}$ & $S_{22}$ & $S_{23}$ & $S_{31}$ & $S_{32}$ & $S_{12}$ & $S_{21}$ & $S_{23}$ & $S_{32}$ \\
\midrule
\armNR & \textbf{55.06} & 54.93 & \textbf{55.82} & 55.72 & 56.79 & 55.74 & -- & -- & \textbf{58.25} & 58.48 & 58.35 & -- \\
\armFront & \textbf{55.06} & 54.93 & \textbf{55.82} & 56.74 & 56.74 & 56.94 & 57.41 & 56.83 & \textbf{58.25} & 59.76 & 59.83 & 59.59 \\
\armPeriodic & \textbf{55.06} & 54.93 & 55.47 & 56.00 & 57.11 & 56.12 & 56.76 & 57.11 & \textbf{58.25} & 59.22 & 59.91 & 59.89 \\
\armCont & 54.74 & \textbf{54.93} & 54.86 & 55.85 & 55.77 & 55.92 & 56.90 & 56.82 & 58.14 & 59.47 & 59.74 & 60.64 \\
\rowcolor{cBox}
\arsi & \textbf{55.06} & 54.93 & \textbf{55.82} & \textbf{57.33} & \textbf{57.63} & \textbf{58.18} & \textbf{58.45} & \textbf{59.50} & \textbf{58.25} & \textbf{59.84} & \textbf{60.40} & \textbf{61.05} \\
\bottomrule
\end{tabular}
\end{table}
\FloatBarrier

\paragraph{Differences between arms.} Table~\ref{tab:diffs} gives the pairwise differences used in the text (stable format in parentheses).

\begin{table}[!htbp]
\caption{Eleven-benchmark pairwise accuracy differences at 8192 tokens, with the corresponding stable-format differences in parentheses. All differences use unrounded scores. \armNR has no $S_{32}$ result because its training halted at $S_{31}$. These are point estimates from one training run per arm.}
\label{tab:diffs}
\centering
\papertablefont
\begin{tabular}{@{}lrrr@{}}
\toprule
Difference & $S_{32}$ & $S_{23}$ & $S_{21}$ \\
\midrule
\armFront $-$ \armNR & -- & $+1.48$ ($+1.37$) & $+1.28$ ($+1.16$) \\
\armFront $-$ \armCont & $-1.05$ ($-0.89$) & $+0.09$ ($+0.16$) & $+0.29$ ($+0.40$) \\
\armFront $-$ \arsi & $-1.45$ ($-1.40$) & $-0.57$ ($-0.56$) & $-0.07$ ($-0.03$) \\
\armFront $-$ \armPeriodic & $-0.30$ ($-0.26$) & $-0.08$ ($-0.04$) & $+0.54$ ($+0.53$) \\
\arsi $-$ \armNR & -- & $+2.05$ ($+1.93$) & $+1.36$ ($+1.18$) \\
\armPeriodic $-$ \armNR & -- & $+1.57$ ($+1.41$) & $+0.74$ ($+0.63$) \\
\armCont $-$ \armNR & -- & $+1.39$ ($+1.21$) & $+0.99$ ($+0.75$) \\
\arsi $-$ \armCont & $+0.40$ ($+0.50$) & $+0.66$ ($+0.72$) & $+0.36$ ($+0.43$) \\
\bottomrule
\end{tabular}
\end{table}

\FloatBarrier
\Needspace{.40\textheight}
\section{Per-Benchmark Results at \texorpdfstring{$S_{32}$}{S32}}\label{app:perset}

\begin{table}[!htbp]
\caption{Per-benchmark endpoint results on the eleven-benchmark suite at $S_{32}$ and 8192 tokens. The best accuracy in each row is bold. Truncation is a percentage within each benchmark; the final row uses equal-benchmark accuracy and pooled-item truncation.}
\label{tab:perset}
\centering
\paperdensefont
\begin{tabular}{@{}lrrrrrr>{\columncolor{cBox}}r>{\columncolor{cBox}[\tabcolsep][0pt]}r@{}}
\toprule
\multirow{2}{*}{Benchmark} & \multicolumn{2}{c}{\armFront} & \multicolumn{2}{c}{\armPeriodic} & \multicolumn{2}{c}{\armCont} & \multicolumn{2}{c}{\arsi} \\
\cmidrule(lr){2-3}\cmidrule(lr){4-5}\cmidrule(lr){6-7}\cmidrule(l){8-9}
 & \textit{Acc.} & \textit{Trunc.} & \textit{Acc.} & \textit{Trunc.} & \textit{Acc.} & \textit{Trunc.} & \textit{Acc.} & \textit{Trunc.} \\
\midrule
ChartQA & 84.40 & 0.20 & \textbf{85.96} & 0.28 & 84.00 & 0.32 & 84.96 & 0.56 \\
MMMU & 63.24 & 7.93 & 64.25 & 10.17 & \textbf{64.80} & 8.83 & 64.36 & 4.58 \\
MMSI & 28.10 & 3.30 & \textbf{29.30} & 4.10 & 27.00 & 2.10 & \textbf{29.30} & 2.50 \\
MathVerse & 63.68 & 3.78 & 63.81 & 5.00 & \textbf{65.94} & 2.18 & 63.48 & 1.93 \\
MathVision & 48.29 & 18.06 & 48.55 & 21.61 & 50.30 & 18.75 & \textbf{51.45} & 8.36 \\
VisNumBench & 54.68 & 3.19 & 53.63 & 5.59 & \textbf{57.55} & 3.08 & 55.99 & 1.52 \\
MMMU-Pro (4 options) & 61.49 & 9.48 & 61.87 & 12.44 & 61.81 & 9.67 & \textbf{63.19} & 4.59 \\
MMMU-Pro (10 options) & 53.45 & 9.67 & 53.39 & 11.49 & 53.71 & 10.68 & \textbf{53.96} & 5.46 \\
MMMU-Pro (vision) & 51.45 & 9.19 & 51.10 & 11.56 & 52.77 & 8.27 & \textbf{53.76} & 4.97 \\
PhyX-MC & 70.20 & 14.67 & 69.10 & 18.63 & 69.80 & 18.50 & \textbf{74.17} & 7.30 \\
SCEMQA-MC & 76.55 & 2.62 & 77.86 & 4.40 & \textbf{79.40} & 1.79 & 76.90 & 1.43 \\
\midrule
Macro11 / pooled trunc. & 59.59 & 8.14 & 59.89 & 10.33 & 60.64 & 8.44 & \textbf{61.05} & 4.16 \\
\bottomrule
\end{tabular}
\end{table}

\FloatBarrier
\section{Evaluation Protocol and Noise Analysis}\label{app:eval}

\paragraph{Benchmark suites.} The main suite contains nine benchmarks shared with earlier diagnostic evaluations: ChartQA \citep{masry2022chartqa}, MMMU \citep{yue2024mmmu}, MathVerse \citep{zhang2024mathverse}, MathVision \citep{wang2024mathvision}, VisNumBench \citep{weng2025visnumbench}, MMSI \citep{yang2026mmsibench}, and the 4-option, 10-option, and vision versions of MMMU-Pro \citep{yue2025mmmupro}. The primary eleven-benchmark suite adds PhyX-MC and SCEMQA-MC, for 22,042 questions. The historical twelve-benchmark diagnostic suite instead adds MathVista \citep{lu2024mathvista}, HallusionBench \citep{guan2024hallusionbench}, and MM-Vet \citep{yu2024mmvet}, for 20,371 questions. Both use complete test splits. Macro11 and \macro are equal-weight means over their respective benchmarks, rather than pooled item accuracies.

\paragraph{Evaluation schedule and metrics.} Main trajectories, refresh-policy and selection ablations, endpoint audits, and the earlier-chain and extension benchmark comparisons use the eleven-benchmark suite and format-corrected scoring. The primary output limit is 8192 tokens; the per-window and earlier-chain tables additionally identify 4096-token results. Data efficiency divides the gain over $\St$ under the same suite, scoring rule, and limit by thousands of new training tasks. The fresh-pool completion decomposition uses all eleven benchmarks with the same equal-benchmark weights. The fixed 1,000-question sample and qualitative cases retain their stated item scopes. The length-budget comparisons use the same eleven-benchmark suite at matched output limits; earlier 16k and multi-hop diagnostics retain their explicitly stated historical scopes.

\paragraph{Auxiliary views.} Stable format keeps letter-, number-, and expression-gold items, applies format-corrected scoring to those same outputs, and equally averages the eleven benchmark subset accuracies (21,415 items). The sentinel is recomputed on the 114 yes/no items retained in the eleven-benchmark suite: COLLAPSE means accuracy below 70\%; WATCH means accuracy below 74\% or a numeric-box share at least 10 points above the two-run $\St$ reference; otherwise it reports OK. Visual-type statistics use the original fixed labels restricted to retained benchmarks and exclude the unannotated science sets.

\paragraph{Scoring and generation.} Each evaluation samples one response per item at temperature 0.6 and top-$p$ 0.95. The prompt and original answer checker follow the public MM-Zero evaluation code \citep{li2026mmzero}. The original matching procedure combines answer matching, unambiguous option-value matching, and Math-Verify equivalence for non-letter gold answers. The format-corrected accuracy reported in the main and ablation tables retains those accepted answers and additionally accepts non-truncated unboxed final answers and normalized option values. ChartQA numeric answers receive a 5\% tolerance, except for year answers or questions asking for a year. Truncation is recorded separately; a length-limited response retains credit when the checker accepts its answer. Format checks reuse the same fixed responses, and each evaluation uses complete test splits at its stated output limit.

\paragraph{LLM-judged accuracy.} \lj uses Qwen2.5-14B-Instruct with the MM-Zero judging prompt, answer extraction, and verdict parser. Questions of at most 400 characters retain their existing judge verdicts; longer questions are re-evaluated with the complete question. The judge receives the extracted final answer, falling back to the first 600 characters of the response when extraction is empty. Judging uses temperature 0 and a 16-token output limit. Reported \lj scores equally average the eleven full benchmark sets.

\paragraph{Measured variability on the eleven-benchmark protocol.} Each policy arm was trained once. The four ladder-control runs with identical first-window data and seed score 54.74, 55.27, 54.45, 54.03 at 4096 tokens (range 1.25, sample sd 0.52). At window 2 and 8192 tokens, the ladder and truncation-aware runs, whose first two pool files hash identically, score 58.14 and 59.60 (difference 1.46). The two evaluations of the same $\St$ weights at 4096 score 54.55 and 54.72, differing by 0.17; their science outputs are shared, so this pair measures decoding variation only on the nine common benchmarks. These examples show variability but do not provide replicated uncertainty for each endpoint policy difference.

\paragraph{Interpreting differences.} We report descriptive single-run point estimates and stable-format checks. The available repeats do not estimate variation across seeds for each policy. Shared-prefix runs are not independent replications, and endpoint differences alone do not establish significance, equivalence, or superiority.

\section{Task Construction and Training Details}\label{app:task-details}

\subsection{Template synthesis and task instantiation}\label{app:template-pipeline}

\paragraph{From seed types to templates.} Given seed task types in geometry, charts, tables, and counting, the base model generates reusable problem structures and their executable implementations. For each structure, it determines the conditions and question target, chooses the variable parameters and their admissible ranges, and writes the procedures for sampling an instance, drawing it, composing its question, and computing its answer. These components form a template $\tau$. The resulting library $\mathcal{T}$ is fixed before GRPO training.

\paragraph{From templates to tasks.} The seven generators in Table~\ref{tab:gen} select structures from the library and sample their parameters. Geometry parameters determine the points, relations, and given quantities. Chart and table parameters determine categories, labels, series, and numerical values. Counting parameters determine the objects and queried attributes. Each generator uses the sampled instance both to render the image and to construct the question and answer. For multiple-choice tasks, the program also constructs distractors and assigns the correct option its answer label.

\paragraph{Execution and checks.} Each candidate undergoes the checks implemented by its generator. Geometry checks compare the derived answer with the constructed relations or coordinate measurements. Chart checks compare the rendered marks with the underlying values and axis specification. Further checks reject ambiguous or invalid questions, and freshness and contamination screens decide whether an accepted candidate can enter the reservoir. Retained records include the image, question, answer, generator identity, and seed. Algorithm~\ref{alg:task-construction} summarizes the pipeline.

\begin{algorithm}[!htbp]
\caption{Template synthesis and replacement-task construction}
\label{alg:task-construction}
\KwIn{Base model, seed task types, generator allocation from Table~\ref{tab:gen}}
Generate template library $\mathcal{T}$ with the base model\;
Fix $\mathcal{T}$ and initialize an empty candidate reservoir\;
\While{a generator has not filled its candidate quota}{
  Select a template $\tau$ and sample parameters $z\sim p_\tau$\;
  Execute $G_\tau(z)$ to obtain image $I$, question $q$, and answer $a$\;
  \If{construction, freshness, and contamination checks pass}{
    Retain $(I,q,a)$ with its generator identity and seed\;
  }
}
Exclude diagnostic holdouts and unsupported training cells to obtain $\mathcal{C}$\;
Draw disjoint 400-task pools from $\mathcal{C}$ under the fixed family quotas\;
Use the training-log detector to decide when to activate each pool\;
\end{algorithm}

\paragraph{A concrete geometry example.} An initial-pool geometry template draws parallel lines $AB$ and $CD$, with a transversal meeting them at $G$ and $H$. The segment $GM$ bisects $\angle BGH$ and meets $CD$ at $M$. The program samples the given angle $\beta=\angle BGH$ from $40^\circ,42^\circ,\ldots,150^\circ$ and constructs the point coordinates. When the question asks for the acute angle $\angle GMH$, the answer is $\beta/2$. Sampling $64^\circ$ therefore produces a question with answer $32^\circ$. The program recomputes the angle from the coordinates and rejects the instance if the two answers disagree.

\subsection{Candidate reservoir and pool construction}\label{app:task-eligibility}

\paragraph{Candidate reservoir.} We run the generators once before the continuation experiments to build the 16,000-task reservoir in Table~\ref{tab:gen}. After the construction checks in Appendix~\ref{app:template-pipeline}, candidates pass a freshness check against earlier pools and task banks, an isomorphism-based contamination screen, and a legacy contamination guard. We exclude 1,488 held-out diagnostic items and 30 plane-geometry items outside the configured difficulty ladders, leaving 14,482 eligible candidates. The two random replacement pools are prepared in advance and share no tasks.

\paragraph{Family mix and vocabulary.} Each replacement pool contains 194 geometry, 131 chart, 54 table, and 21 counting tasks. The shares follow the visual-type counts of the 12 evaluation benchmarks: 4,912 geometry, 3,308 chart, 1,369 table, and 537 counting items (48.5\%, 32.7\%, 13.5\%, 5.3\%). Chart and table vocabulary also uses 19 thematic terms extracted from benchmark question text. These choices fix the task mix and vocabulary before training. The refresh detector does not read benchmark answers or scores. Function plots, numerical estimation, and spatial tasks are outside the current generator coverage.

\begin{table}[!htbp]
\caption{The seven program-based generators and the candidate library. Gold answers were independently recomputed: mv 3,881/3,881, ref 1,568/1,568, tab/sc/ca 4,580/4,580, ga 150/150 on a sample, and hr 1,922/1,927 (the five differences are ties for ``third highest'', 0.03\% of candidates).}
\label{tab:gen}
\centering
\papertablefont
\begin{tabularx}{\linewidth}{@{}l l >{\raggedright\arraybackslash}X r >{\raggedright\arraybackslash}p{3.7cm}@{}}
\toprule
Family & Generator & Task type & Candidates & Difficulty tiers (diagnosis only) \\
\midrule
\multirow{2}{*}{Geometry} & mv & MathVision-style multi-hop geometry; each hop uses the previous result and a new fact in the figure & 3,881 & two ladders of 5 tiers (1--5 hops) \\
 & ga & plane geometry & 3,880 & 5 tiers \\
\midrule
\multirow{3}{*}{Chart} & hr & multi-series chart reading & 2,091 & ten-cell reading ladder \\
 & ca & single-series chart (bar, line) & 1,568 & none \\
 & ref & referring multi-hop chart & 1,568 & 3 tiers \\
\midrule
Table & tab & table & 2,163 & none \\
\midrule
Counting & sc & counting scenes & 849 & 3 tiers \\
\midrule
\multicolumn{3}{@{}l}{Total (all fresh and screened)} & 16,000 & \\
\bottomrule
\end{tabularx}

\end{table}

\subsection{Initial pool and optimization}\label{app:initial-pool}

\paragraph{Initial pool.} The initial 500 tasks come from three earlier generation batches, two for charts and one for geometry. The retained pool contains 204 geometry questions and 296 chart questions. Programs supply the answers, while the earlier pipeline uses an API-served Qwen3-VL-8B for question processing and checks. Of the retained questions, 342 use model-rewritten wording, 117 keep the original program wording, and 41 revert to it after rewriting. Pool selection uses eight-response probes and filters candidates by correct-answer count and truncation, with constraints on the question-type mix and reuse of each image. This pool is separate from the later 16,000-task reservoir. Two questions in the initial pool were later identified as configuration isomorphs of MathVerse items.

\paragraph{Initialization and optimization.} The starting checkpoint $\St$ follows 100 GRPO updates of Qwen3-VL-8B-Instruct on the old pool. In the preceding plateau check, another 300 steps on the same pool changed the benchmark average by $+0.45$, $-0.27$, $-0.38$, and $-0.63$ across four training seeds. The continuation experiments start from $\St$ and retain that pool until their respective replacement rules act.

The five-arm comparison reuses accepted checkpoints where the runs coincide. All arms except \armCont share the first two windows on the old pool, and \arsi and \armFront also share the third. The first four windows of \armCont come from an earlier run of the same difficulty-control strategy, with fresh pools used from window 1.

Training uses FSDP2 and colocated vLLM, with 100 optimizer updates per window and a constant learning rate of $10^{-6}$. Each window starts with a fresh optimizer and scheduler, retaining the preceding model parameters. Each prompt has eight rollouts, sampled at temperature 1.0 and top-$p$ 1.0 with a 4,096-token response limit. A generation batch contains eight prompt groups and serves two optimizer steps. The detector counts each batch once, giving 50 distinct batches and 3,200 responses per window. The GRPO objective uses group-normalized advantages, a clipping threshold of 0.2, and no KL penalty. Dynamic resampling checks variance in correctness plus termination reward and excludes the format term, with at most six resampling rounds. The final optimization reward and logged advantages include all three terms. The detector reads those logged advantages after resampling.

The reward coefficients in Eq.~\eqref{eq:reward} are $\lambda_{\mathrm{fmt}}=0.1$ and $\lambda_{\mathrm{stop}}=0.3$, with unit weight on answer correctness. These coefficients stay fixed across windows.

The stopping term in Eq.~\eqref{eq:reward} uses the decoder's recorded finish reason and the final boxed answer. A response that reaches the token limit receives $-1$ even if it contains a boxed expression earlier in the text. Once a response ends normally with a final boxed answer, this term stays at 0 regardless of the response length. Its effect on truncation and benchmark scores is examined in Sections~\ref{sec:why} and~\ref{sec:budget}.

\paragraph{Refresh-controller settings.}\label{app:controller-settings}
The controller in Eq.~\eqref{eq:trigger} uses lag $\ell=2$, minimum within-pool observation period $n_{\min}=3$, signal-retention threshold $\rho=0.85$, and accuracy-increase threshold $\eta=0.10$. The numerical tolerance in Eq.~\eqref{eq:pool-signals} is $\varepsilon=10^{-9}$. Thus, after at least three accepted windows on the same pool, a decline of at least 15\% in the distinct-prompt advantage signal or a rise of at least ten percentage points in logged accuracy, relative to two windows earlier, triggers replacement. The initial pool contains 500 tasks, each replacement contains $B=400$, and each window has $H=100$ updates with $m=8$ responses per prompt. Replacement-family quotas are $\mathbf{k}=(194,131,54,21)$, as detailed in Appendix~\ref{app:task-eligibility}. All controller settings are fixed before training.

\subsection{Diagnostic difficulty tiers}\label{sec:tiers}

For diagnosis we organize tasks into six controlled ladders: two multi-hop geometry ladders of five tiers, plane geometry of five, chart reading of ten cells, referring multi-hop of three, and counting of three. Each ladder orders related tasks from easy to hard. Accuracy is measured on held-out synthetic tasks that are excluded from training. Models' own predictions of which questions they will get wrong give AUC about 0.50, a program-based predictor 0.58--0.61, and measured accuracy of other models 0.72--0.82. At $\St$, the pass rate falls monotonically with tier, for example .94, .57, .32, .14, and .07 from tier 1 to 5 on one multi-hop ladder. The comparison arms build frontier pools from these measurements (Section~\ref{sec:pick}).

For \armFront and \armPeriodic, the frontier is the easiest tier in each ladder with a held-out pass rate of at most 70\%. If none qualifies, we select the hardest tier. The selected tier supplies 50\% of the ladder's quota, and each adjacent tier supplies 25\%. \armCont re-measures pass rates after every window and moves the selected tier up one step at a pass rate of at least 80\%, down one step at a pass rate of at most 20\%, and otherwise keeps it unchanged. It uses the same allocation across the selected and adjacent tiers. These measurements do not affect random replacement or its refresh trigger.

\section{Training Health Gate}\label{app:health}

In the main Qwen experiment, a window is accepted when its checkpoint export is complete, at least 50 generation batches are present, the learning rate remains $10^{-6}$ for all 100 updates, finish reasons are recorded, the training source has not changed, and the weights have moved. The weight check uses the median relative change of three sampled tensors, with a threshold of $3.69\times10^{-5}$. At least 97 of the 100 updates must have finite, nonzero gradient norms. A failed window is retried once from the same preceding checkpoint with the same pool and seed. A successful retry enters the detector normally. A second failure halts the run.

The trainer can also stop an attempt before all 100 updates complete: if six consecutive generation batches each contain at least one zero-advantage group after resampling, the attempt is aborted. This incomplete attempt fails the window gate and enters the same-pool retry procedure. The condition concerns the presence of a zero-advantage group in each batch, even when other groups carry signal; it does not establish that every task in the pool has become uninformative.

The detector, controller, probes, and pool builder use access guards that block benchmark scores and evaluation outputs. The legacy contamination guard runs separately and returns a flag. An independent checker logged file accesses during the guarded processes and recorded zero test-set accesses.

The earlier experiments required at least 99 of 100 updates to have normal gradients. Seven attempts failed at 98/100 across five windows. Fixed-pool continuation failed once at each of windows 10 and 11, and original-pool continuation failed once at window 3. Their retries passed, and the 8192-token comparison with MM-Zero uses the retrained fixed-pool continuation weights. Partial-pool refresh failed twice at window 12, and one ladder-control run failed twice at window 3. Both runs stopped. Each failure arose from a single generation batch in which all eight groups had zero variance. The new experiments pre-registered a threshold of 97/100. In them, \armNR failed twice at window 7 with 94/100 normal steps, 35 zero-signal warnings, and training accuracy .728.

\section{Engineering Notes}

\begin{itemize}[leftmargin=*]
  \item \textbf{File timestamps on the shared disk} have a one-second resolution and can lag. A pre-registered ordering check based on modification times blocked arm \armNR at window 5 (two files written in the same second). The order is now proven by a content hash chain, with modification times as a coarse check with five seconds of tolerance, and the check is rehearsed with same-second mutants. The first version of the experiment was frozen and rebuilt as sagpd2 under the rule that each experiment is frozen only once.
  \item \textbf{Quota exhaustion} halted construction for about 11 hours once and left five machines idle; long dependency chains are now queued ahead of time.
  \item \textbf{A pre-registered health gate does block runs.} Readouts are written as ``no reading'' and the gate is not relaxed afterwards.
  \item \textbf{Command arguments must be quoted} (an unquoted argument disabled depth feedback in one experiment), and base port plus stride times shard count must stay under 65,535.
  \item \textbf{Zero-data discipline.} Using the 70,000 public ThinkLite tasks as a candidate pool was considered and dropped before any download, since it would violate the zero-data rule.
\end{itemize}

\section{Further Documented Issues in the MM-Zero Reproduction}\label{app:mmz}

The following are observations in our reproduction setting and are not claims about the correctness of the released system.
\begin{itemize}[leftmargin=*]
  \item \textbf{Reproduction.} Published base and v5 scores reproduce (50.27 against 50.03, and 54.25 against 54.18). Local training reproduces three rounds (v1 $+2.67$, v2 $+4.05$, v3 $+3.30$ over base).
  \item \textbf{Gain is mostly finishing, by four routes.} Paired per-item analysis (4,006 base failures that did not finish, from 0.30 to 29.18); strict reproduction (7,257 finishing items, v1 $+0.69$, v2 $+0.37$, p-values above); an untrained base at 16384 matching MM-Zero at 4096 (51.12 against 51.08); and the pre-registered 8192 comparison (59\% retained).
  \item \textbf{Source and paper review.} The two equations for the difficulty reward and the training label contradict each other; no image-necessity check (the figure-1 example can be answered without the image); text and table disagree (56.6\% against 54.5\%; the 4B block average differs from its cells by up to 0.89; 10 against 60 training steps; 5 against 8 rollouts; the same 8B base evaluated twice differs by 0.4 with no error bars); the central claim that three roles beat two has no experiment (the coder was never removed); one equation branch can never fire and another affects 0.03\% of batches yet is credited with 2.9 points; labels have no program-based ground truth; the denominator drops truncated responses; about 16\% of released images are almost entirely black.
  \item \textbf{A retracted accusation.} We had earlier reported that MM-Zero lowers accuracy on items the base model already answers ($-1.28$). With a zero-effect baseline (the same model compared with itself gives $-3.26$) the change is $+2.0$ and there is no harm. That claim is withdrawn.
  \item \textbf{Not done.} A 16k fair-budget comparison (about four node-hours), and mechanistic ablations of MM-Zero were not run.
\end{itemize}

\section{Limitations}\label{sec:limits}

\begin{itemize}[leftmargin=*,itemsep=2pt,parsep=0pt,topsep=4pt]
  \item \textbf{Replication and evaluation horizon.} Each arm was trained once. The main comparison spans eight continuation windows and has no 16k evaluation. Small score differences lack significance testing. \armNR halted at $S_{31}$, leaving the pre-registered $S_{32}$ endpoint unobserved.
  \item \textbf{Refresh timing.} The first refresh point was set using existing logs. Without a random-refresh control on a fixed three-window schedule, the contribution of history-dependent timing remains entangled with the use of fresh tasks.
  \item \textbf{Cross-model transfer.} MiMo has a positive fresh-versus-retained endpoint difference in one run, with small margins over \armCont and \armPeriodic. Their pools were selected at Qwen's difficulty frontier, limiting conclusions about transfer across model families.
  \item \textbf{Generator scope.} The fixed programmatic library covers geometry, charts, tables, and counting, but omits function plots, numerical estimation, and spatial tasks. Task-family shares and 19 thematic terms were informed by the evaluation benchmarks; diversity beyond these templates remains unmeasured. The initial task pool also contains model-rewritten questions with program-computed answers.
\end{itemize}

\section{Additional Qualitative Case Studies}\label{app:additional-cases}

We select ten additional favorable examples from eight of the main-result benchmarks, comprising six Qwen3-VL-8B-Instruct and four MiMo-VL-7B-SFT cases. They are distinct from the four cases in Section~\ref{sec:cases}. Every example compares that model's public starting weights $S_0$ with its own $S_{32}$: the second accepted window on Pool 3, corresponding to continuation window 8. All cases use this common comparison stage rather than different training stages, at the same 8192-token evaluation limit. Each card identifies the model and both stages explicitly. In each of two independent decodings, both responses end normally, the base is incorrect under format-corrected scoring, and $S_{32}$ is correct under both answer checks. The primary decoding at $S_{33}$ also remains correct. The cards reproduce short verbatim excerpts from the primary decoding; the surrounding explanations summarize our inspection of the original images and responses. These selected successes illustrate local corrections, not their population frequency, an overall capability gain, or a causal effect of a particular training component. Several corrected responses are longer than the base responses, so shorter output alone cannot explain these cases.

\begingroup
\newcommand{\appendixcasetokens}[2]{%
  \tcblower
  \begingroup
  \pgfmathsetlengthmacro{\casebarwidth}{.93\linewidth}
  \pgfmathsetmacro{\casebarmax}{max(#1,#2)}
  \begin{tikzpicture}[x=\casebarwidth,y=.34cm,font=\scriptsize]
    \node[anchor=west,inner sep=0pt] at (0,2.35) {Generated tokens (primary decoding)};
    \fill[caseBefore!24,rounded corners=1pt] (0,1) rectangle ({#1/\casebarmax},1.55);
    \node[anchor=west,inner sep=2pt] at (0,1.28) {Before $S_0$: #1};
    \fill[caseAfter!24,rounded corners=1pt] (0,0) rectangle ({#2/\casebarmax},.55);
    \node[anchor=west,inner sep=2pt] at (0,.28) {After $S_{32}$: #2};
  \end{tikzpicture}
  \endgroup\par
}

\begin{figure}[H]
\centering
\begin{minipage}[t]{.48\linewidth}\vspace{0pt}
\begin{casecard}{caseBlue}{C1. Include both relevant groups!}
\textbf{ChartQA \#627}\par
\casemodelstage{Qwen3-VL-8B-Instruct}
\caseimage[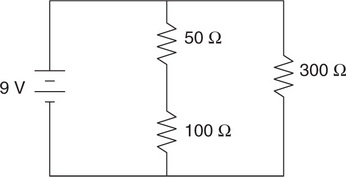]{}
\textbf{Task.} What percentage would spend time in Poland? Reference: 73.

\medskip{\sffamily\bfseries\color{caseBefore}BEFORE: completed but incorrect}\par
The base response counts only the 66 percent who choose Poland exclusively.
{\small\itshape ``The category "Only in Poland" directly answers the question''}\par

\medskip{\sffamily\bfseries\color{caseAfter}AFTER: completed and correct}\par
The later response also includes the 7 percent who choose both Poland and abroad.
{\small\itshape ``So, I add the two relevant percentages: 66\% + 7\% = 73\%.''}\par
\[
66+7=\boxed{73}.
\]
\appendixcasetokens{118}{149}
\smallskip\textbf{What changed?} The labels are read correctly in both responses; the correction is which categories belong in the requested total.
\end{casecard}
\end{minipage}\hfill
\begin{minipage}[t]{.48\linewidth}\vspace{0pt}
\begin{casecard}{caseBlue}{C2. A branch is not one resistor!}
\textbf{SCEMQA-MC \#652}\par
\casemodelstage{Qwen3-VL-8B-Instruct}
\caseimage[assets/cases/supplementary/ChartQA_627_0.png]{assets/cases/supplementary/sci_scemqa_mc_652_0.png}
\textbf{Task.} Find the voltage across the $50\,\Omega$ resistor. Reference: C, 3.0 V.

\medskip{\sffamily\bfseries\color{caseBefore}BEFORE: completed but incorrect}\par
The base identifies the series branch but incorrectly assigns its full 9 V to the 50-ohm component.
{\small\itshape ``both are equal to the battery voltage.''}\par

\medskip{\sffamily\bfseries\color{caseAfter}AFTER: completed and correct}\par
The later response distinguishes branch voltage from component voltage and applies the divider rule.
{\small\itshape ``this voltage is divided between the two resistors according to their resistance values.''}\par
\[
V_{50}=9\,\frac{50}{50+100}=\boxed{3\,\mathrm{V}}.
\]
\appendixcasetokens{247}{349}
\smallskip\textbf{What changed?} The circuit interpretation is retained while the component-level calculation is repaired.
\end{casecard}
\end{minipage}
\caption{\textbf{Additional completed-answer corrections.} Category aggregation and component-level calculation. C1 includes both relevant chart categories; C2 divides the branch voltage across its series resistors. Both decoding repeats show the wrong-to-right transition for each item; examples are selected qualitatively.}
\label{fig:additional-cases-1}
\end{figure}
\FloatBarrier

\clearpage

\begin{figure}[H]
\centering
\begin{minipage}[t]{.48\linewidth}\vspace{0pt}
\begin{casecard}{caseBlue}{C3. Read the hits, then solve!}
\textbf{MATH-Vision \#585}\par
\casemodelstage{Qwen3-VL-8B-Instruct}
\caseimage[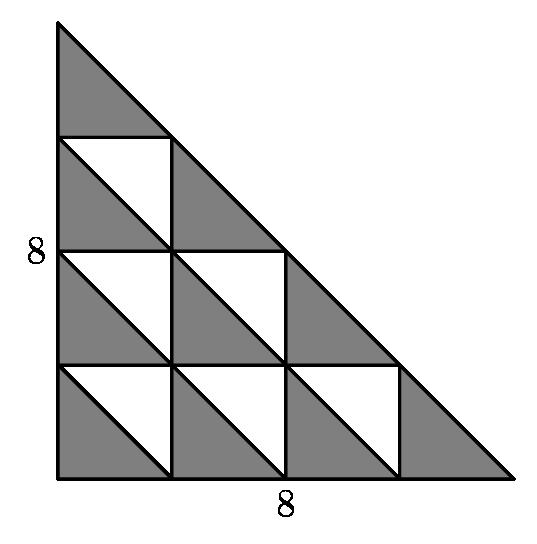]{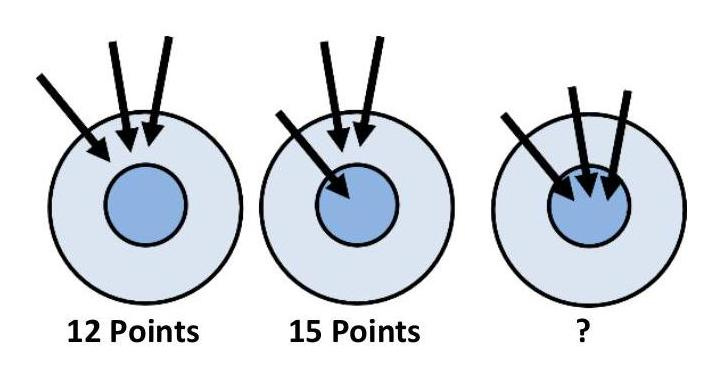}
\textbf{Task.} Three darts score 12, then 15 points. What is the third score? Reference: 21.

\medskip{\sffamily\bfseries\color{caseBefore}BEFORE: completed but incorrect}\par
The base reverses the first target's hit locations, assigns 4 points to the inner field, and answers 12.
{\small\itshape ``In the first shot, 3 darts hit the inner blue circle''}\par

\medskip{\sffamily\bfseries\color{caseAfter}AFTER: completed and correct}\par
The later response correctly distinguishes the outer and inner hits and solves for their separate scores.
{\small\itshape ``In the first scenario, all three darts hit the outer ring''}\par
\[
3x=12,\quad 2x+y=15\;\Longrightarrow\;3y=\boxed{21}.
\]
\appendixcasetokens{1350}{234}
\smallskip\textbf{What changed?} Correct visual grounding makes the same simple algebra produce the right answer.
\end{casecard}
\end{minipage}\hfill
\begin{minipage}[t]{.48\linewidth}\vspace{0pt}
\begin{casecard}{caseBlue}{C4. Count the shaded pieces!}
\textbf{MATH-Vision \#2555}\par
\casemodelstage{Qwen3-VL-8B-Instruct}
\caseimage[assets/cases/supplementary/mathvision_585_0.png]{assets/cases/supplementary/mathvision_2555_0.png}
\textbf{Task.} Find the shaded area of a right triangle with legs 8 and 16 congruent pieces. Reference: 20.

\medskip{\sffamily\bfseries\color{caseBefore}BEFORE: completed but incorrect}\par
The base eventually assumes four white pieces and twelve shaded pieces, giving 24.
{\small\itshape ``So the shaded area is 24.''}\par

\medskip{\sffamily\bfseries\color{caseAfter}AFTER: completed and correct}\par
The later response counts ten shaded pieces and uses the area of one piece.
{\small\itshape ``By examining the image, we can see that there are 10 shaded small triangles.''}\par
\[
A_{\mathrm{shaded}}=\frac{\tfrac12(8)(8)}{16}\,(10)=\boxed{20}.
\]
\appendixcasetokens{4555}{241}
\smallskip\textbf{What changed?} The correction concerns the visual count; the total-area formula was already available to the base.
\end{casecard}
\end{minipage}
\caption{\textbf{Additional completed-answer corrections.} Visual grounding followed by arithmetic. C3 distinguishes the two target fields before solving their scores; C4 counts the shaded pieces before multiplying by their area. Both decoding repeats show the wrong-to-right transition for each item; examples are selected qualitatively.}
\label{fig:additional-cases-2}
\end{figure}
\FloatBarrier

\clearpage

\begin{figure}[H]
\centering
\begin{minipage}[t]{.48\linewidth}\vspace{0pt}
\begin{casecard}{caseBlue}{C5. Use the supplementary angle!}
\textbf{MathVerse \#226}\par
\casemodelstage{Qwen3-VL-8B-Instruct}
\caseimage[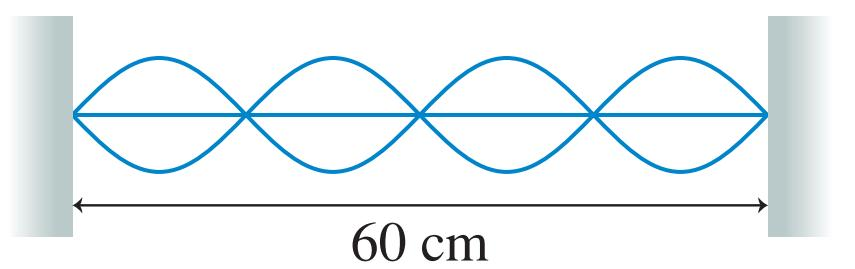]{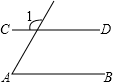}
\textbf{Task.} With $AB\parallel CD$ and angle $A=70^\circ$, find angle 1. Reference: D, $110^\circ$.

\medskip{\sffamily\bfseries\color{caseBefore}BEFORE: completed but incorrect}\par
The base treats the marked angle as corresponding to the given angle and selects 70 degrees.
{\small\itshape ``so it's a corresponding angle.''}\par

\medskip{\sffamily\bfseries\color{caseAfter}AFTER: completed and correct}\par
The later response identifies the adjacent linear pair at the lower parallel line.
{\small\itshape ``Angle 1 and this 70\(^{\circ}\) angle are on a straight line, so they are supplementary.''}\par
\[
\angle 1=180^\circ-70^\circ=\boxed{110^\circ}.
\]
\appendixcasetokens{234}{353}
\smallskip\textbf{What changed?} The marked region is interpreted correctly before applying the parallel-line relation.
\end{casecard}
\end{minipage}\hfill
\begin{minipage}[t]{.48\linewidth}\vspace{0pt}
\begin{casecard}{caseBlue}{C6. A loop is half a wavelength!}
\textbf{PhyX \#1353}\par
\casemodelstage{MiMo-VL-7B-SFT}
\caseimage[assets/cases/supplementary/mathverse_226_0.png]{assets/cases/supplementary/sci_phyx_mc_1353_0.png}
\textbf{Task.} A standing wave spans 60 cm and oscillates at 100 Hz. Find its speed. Reference: D, 30 m/s.

\medskip{\sffamily\bfseries\color{caseBefore}BEFORE: completed but incorrect}\par
The base counts the four standing-wave loops as four full wavelengths and obtains 15 m/s.
{\small\itshape ``there are 4 full wavelengths in this distance.''}\par

\medskip{\sffamily\bfseries\color{caseAfter}AFTER: completed and correct}\par
After revising its count, the later response ends with the diagram-consistent two-wavelength explanation.
{\small\itshape ``there are 2 full wavelengths.''}\par
\[
\lambda=\frac{0.60}{2}=0.30\,\mathrm{m},\qquad v=100\lambda=\boxed{30\,\mathrm{m/s}}.
\]
\appendixcasetokens{927}{1185}
\smallskip\textbf{What changed?} The final explanation repairs the wavelength count. Its intermediate reasoning still contains revisions.
\end{casecard}
\end{minipage}
\caption{\textbf{Additional completed-answer corrections.} Interpreting the marked geometry. C5 uses a supplementary rather than equal angle; C6 distinguishes standing-wave loops from full wavelengths in its final explanation. Both decoding repeats show the wrong-to-right transition for each item; examples are selected qualitatively.}
\label{fig:additional-cases-3}
\end{figure}
\FloatBarrier

\clearpage

\begin{figure}[H]
\centering
\begin{minipage}[t]{.48\linewidth}\vspace{0pt}
\begin{casecard}{caseBlue}{C7. Compare the full object length!}
\textbf{VisNumBench \#1547}\par
\casemodelstage{MiMo-VL-7B-SFT}
\caseimage[]{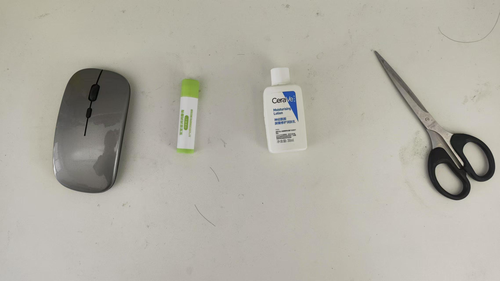}
\textbf{Task.} Which object, A--D from left to right, has the greatest length? Reference: d, the scissors.

\medskip{\sffamily\bfseries\color{caseBefore}BEFORE: completed but incorrect}\par
The base selects the mouse even though the scissors extend farther along their long axis.
{\small\itshape ``The mouse (A) seems to be the longest among them.''}\par

\medskip{\sffamily\bfseries\color{caseAfter}AFTER: completed and correct}\par
The later response compares the full object extents and selects the rightmost scissors.
{\small\itshape ``Scissors are longer than the others.''}\par
\appendixcasetokens{127}{207}
\smallskip\textbf{What changed?} The visual comparison changes from A to D; both outputs already submit an explicit answer.
\end{casecard}
\end{minipage}\hfill
\begin{minipage}[t]{.48\linewidth}\vspace{0pt}
\begin{casecard}{caseBlue}{C8. Locate the quietest marking!}
\textbf{MMMU-Pro (4 options) \#1008}\par
\casemodelstage{MiMo-VL-7B-SFT}
\caseimage[assets/cases/supplementary/visnumbench_1547_0.png]{assets/cases/supplementary/mmmu_pro_4_options_1008_0.png}
\textbf{Task.} Is bar 8 the quietest bar in the displayed music? Reference: A, True.

\medskip{\sffamily\bfseries\color{caseBefore}BEFORE: completed but incorrect}\par
The base recognizes pianissimo but places the marking outside bar 8 and answers False.
{\small\itshape ``So bar 8 is not the quietest.''}\par

\medskip{\sffamily\bfseries\color{caseAfter}AFTER: completed and correct}\par
The later response associates the displayed pianissimo marking with bar 8 and answers True.
{\small\itshape ``Bar 8 features the pp marking.''}\par
\appendixcasetokens{590}{299}
\smallskip\textbf{What changed?} The correction links a known musical symbol to its location. Here $pp$ is the softest marking shown, not a universal minimum dynamic.
\end{casecard}
\end{minipage}
\caption{\textbf{Additional completed-answer corrections.} Visual comparison and symbol localization. C7 compares full object lengths; C8 locates the pianissimo marking in the requested bar. Both decoding repeats show the wrong-to-right transition for each item; examples are selected qualitatively.}
\label{fig:additional-cases-4}
\end{figure}
\FloatBarrier

\clearpage

\begin{figure}[H]
\centering
\begin{minipage}[t]{.48\linewidth}\vspace{0pt}
\begin{casecard}{caseBlue}{C9. Account for all three effects!}
\textbf{MMMU \#442}\par
\casemodelstage{MiMo-VL-7B-SFT}
\caseimage[]{}
\textbf{Task.} What is the total capacity of the forward-feed evaporator? Reference: A, $E_1+E_2+E_3$.

\medskip{\sffamily\bfseries\color{caseBefore}BEFORE: completed but incorrect}\par
The base focuses on the first feed stage and chooses its evaporation term alone.
{\small\itshape ``Option C is E1. So the answer should be C.''}\par

\medskip{\sffamily\bfseries\color{caseAfter}AFTER: completed and correct}\par
The later final explanation treats capacity as total evaporation across the three-effect system.
{\small\itshape ``each effect contributes to removing water from the feed.''}\par
\[
E_{\mathrm{total}}=\boxed{E_1+E_2+E_3}.
\]
\appendixcasetokens{252}{825}
\smallskip\textbf{What changed?} The system boundary changes from one effect to the full apparatus. The base's raw text selects C, so this is not solely a formatting repair.
\end{casecard}
\end{minipage}\hfill
\begin{minipage}[t]{.48\linewidth}\vspace{0pt}
\begin{casecard}{caseBlue}{C10. Read the value, not the footer!}
\textbf{ChartQA \#714}\par
\casemodelstage{Qwen3-VL-8B-Instruct}
\caseimage[assets/cases/supplementary/MMMU_442_0.png]{assets/cases/supplementary/ChartQA_714_0.png}
\textbf{Task.} How many female murder offenders were there in 2019? Reference: 1,408.

\medskip{\sffamily\bfseries\color{caseBefore}BEFORE: completed but incorrect}\par
The base mistakes the 2021 copyright footer for the data year and withholds an answer despite mentioning 1,408.
{\small\itshape ``it is not possible to determine the number of female murder offenders in 2019 from this image.''}\par

\medskip{\sffamily\bfseries\color{caseAfter}AFTER: completed and correct}\par
The later response answers from the Female label. The image itself does not separately print the data year.
{\small\itshape ``the value associated with the "Female" category is 1,408.''}\par
\appendixcasetokens{108}{132}
\smallskip\textbf{What changed?} This is a correction of answer acceptance and date interpretation, rather than newly learned arithmetic.
\end{casecard}
\end{minipage}
\caption{\textbf{Additional completed-answer corrections.} System scope and answer acceptance. C9 accounts for evaporation in every effect; C10 reads the chart label instead of treating a copyright date as a reason to abstain. Both decoding repeats show the wrong-to-right transition for each item; examples are selected qualitatively.}
\label{fig:additional-cases-5}
\end{figure}
\FloatBarrier

\endgroup

\FloatBarrier
\end{document}